\documentclass[5p,times,authoryear]{elsarticle}
\makeatletter
\let\c@author\relax
\let\c@bibfont\relax
\makeatother

\usepackage{standalone}
\usepackage[T1]{fontenc}
\usepackage{newtxtext}  
\usepackage{array}
\usepackage{multicol}
\usepackage{dcolumn}
\usepackage{booktabs}
\usepackage{microtype}
\usepackage{url}
\usepackage{makecell} 

\usepackage{amsmath}
\usepackage{amssymb}
\usepackage{mathtools}
\usepackage{amsthm}

\usepackage{siunitx}
\robustify\bfseries  

\usepackage[inline]{enumitem}

\usepackage{csquotes}

\usepackage{xspace}
\usepackage{relsize}  
\newcommand{\class}[1]{{\smaller{\texttt{#1}}}\xspace}%
\newcommand{\meadow}{\class{pasture meadow grassland grass}}
\newcommand{\wheat}{\class{wheat}}
\newcommand{\barley}{\class{barley}}
\newcommand{\region}[1]{{\smaller{\textsf{#1}}}\xspace}%
\newcommand{\portugal}{\region{PT}}
\newcommand{\latvia}{\region{LV}}
\newcommand{\estonia}{\region{EE}}
\newcommand{\regioncase}[1]{{\smaller{\textsf{#1}}}}%
\newcommand{\lvee}{\textbf{\regioncase{LV}$\rightarrow$\region{EE}}\xspace}
\newcommand{\lvptee}{\textbf{\regioncase{LV}+\regioncase{PT}$\rightarrow$\region{EE}}\xspace}

\newcommand{\dataset}[1]{\textsc{#1}\xspace}
\newcommand{\EuroCrops}{\dataset{EuroCrops}}
\newcommand{\EuroCropsML}{\dataset{EuroCropsML}}
\newcommand{\ZueriCrop}{\dataset{ZueriCrop}}
\newcommand{\Breizhcrops}{\dataset{Breizhcrops}}
\newcommand{\CropHarvest}{\dataset{CropHarvest}}

\newcommand{\CroplandDataLayer}{\dataset{Cropland Data Layer}}

\newcommand{\SenMSCRTS}{\dataset{Sen12MS-CR-TS}}
\newcommand{\PixelSenMSCRTS}{\dataset{Pixel-Sen12MS-CR-TS}}
\newcommand{\task}[1]{{\textsc{#1}}\xspace}%
\newcommand{\taskalldata}{\textbf{\task{\EuroCropsML S1+S2+ERA5}}}
\newcommand{\taskStwo}{\textbf{\task{\EuroCropsML S2}}}

\newcommand{\channel}[1]{{\smaller{\texttt{#1}}}\xspace}%
\newcommand{\channelb}[1]{{\smaller{\texttt{#1}}}}%
\newcommand{\VVb}{\channelb{VV}}
\newcommand{\VHb}{\channelb{VH}}
\newcommand{\VV}{\channel{VV}}
\newcommand{\VH}{\channel{VH}}
\newcommand{\Bone}{\channel{B01}}
\newcommand{\Btwo}{\channel{B02}}

\newcommand{\Bfour}{\channel{B04}}
\newcommand{\Bfive}{\channel{B05}}

\newcommand{\Bseven}{\channel{B07}}
\newcommand{\Beight}{\channel{B08}}
\newcommand{\BeightA}{\channel{B8A}}
\newcommand{\Bnine}{\channel{B09}}
\newcommand{\Bten}{\channel{B10}}
\newcommand{\Beleven}{\channel{B11}}
\newcommand{\Btwelve}{\channel{B12}}

\usepackage[dvipsnames]{xcolor,colortbl}
\newcolumntype{X}{>{\scriptsize}l}
\definecolor{purple}{RGB}{129, 114, 179}
\definecolor{indianred}{RGB}{196,78,82}
\definecolor{lavender}{RGB}{234,234,242}
\definecolor{mediumseagreen}{RGB}{85,168,104}
\definecolor{peru}{RGB}{221,132,82}
\definecolor{steelblue}{RGB}{76,114,176}
\definecolor{LightOrange}{RGB}{210,110,50}
\definecolor{LightCyan}{RGB}{0,180,180}

\usepackage[
  acronym,
  automake,
]{glossaries-extra}
\makeglossaries

\setabbreviationstyle[acronym]{long-short-user}
\renewcommand*{\glsxtruserparen}[2]{
  \glsxtrfullsep{#2}%
  \glsxtrparen
   {#1\ifglshasfield{\glsxtruserfield}{#2}{;\xspace%
     \expandafter\citealp\expandafter{\glscurrentfieldvalue}%
   }{}%
   }%
}

\glsdefpostdesc{acronym}{
 \ifglshasfield{\glsxtruserfield}{\glscurrententrylabel}%
 {~\expandafter\citep\expandafter{\glscurrentfieldvalue}}%
 {}%
}

\glsdisablehyper
\defglsentryfmt[acronym]{\ifglsused{\glslabel}{\glsgenentryfmt}{\emph{\glsgenentryfmt}}}

\newacronym{gl:ai}{AI}{artificial intelligence}
\newacronym{gl:dl}{DL}{deep learning}
\newacronym{gl:ml}{ML}{machine learning}

\newacronym{gl:presto}{PRESTO}{\textbf{P}retrained \textbf{Re}mote \textbf{S}ensing \textbf{T}ransf\textbf{o}rmer}
\newacronym{gl:ssl}{SSL}{self-supervised learning}
\newacronym[user1={finn_2017_maml}]{gl:maml}{MAML}{model-agnostic meta-learning}
\newacronym[user1={finn_2017_maml}]{gl:fomaml}{FOMAML}{first-order model-agnostic meta-learning}
\newacronym[user1={Raghu2019RapidLO}]{gl:anil}{ANIL}{almost no inner loop}
\newacronym[user1={tseng_2022_timl}]{gl:timl}{TIML}{task-informed meta-learning}
\newacronym[user1={Fuller23:croma}]{gl:croma}{CROMA}{contrastive radar-optical masked autoencoders}

\newacronym{gl:vit}{ViT}{Vision Transformer}
\newacronym{gl:cv}{CV}{computer vision}
\newacronym{gl:mlp}{MLP}{multi-layer perceptron}
\newacronym[user1={He22:MAE}]{gl:mae}{MAE}{masked autoencoder}
\newacronym[user1={fu2024_crossMAE}]{gl:crossmae}{CrossMAE}{cross-attention masked autoencoder}
\newacronym{gl:prestoxts}{PRESTO-XTS}{PRESTO-E\textbf{X}tended \textbf{T}emporal and \textbf{S}pectral}
\newacronym{gl:crossprestoxts}{CrossPRESTO-XTS}{cross-attention PRESTO-XTS}
\newacronym[user1={He16:DRL}]{gl:resnet}{ResNet}{residual network}

\newacronym{gl:eu}{EU}{European Union}
\newacronym{gl:esa}{ESA}{European Space Agency}
\newacronym{gl:gee}{GEE}{Google Earth Engine}

\newacronym{gl:mse}{MSE}{mean squared error}
\newacronym{gl:tpe}{TPE}{bayesian tree-parzen estimator}
\newacronym{gl:sgd}{SGD}{stochastic gradient descent}
\newacronym{gl:cap}{CAP}{common agriculture policy}

\newacronym[user1={schneider_eurocrops_2021,schneider_eurocrops_2023}]{gl:hcat}{HCAT}{hierarchical crop and agriculture taxonomy}
\newacronym[user1={Rouse74:NDVI}]{gl:ndvi}{NDVI}{normalized difference vegetation index}
\newacronym[user1={Hersbach20:ERA5}]{gl:era}{ERA5}{\textbf{E}uropean \textbf{R}e\textbf{A}nalysis Version \textbf{5}}
\newacronym{gl:sar}{SAR}{synthetic aperture radar}
\newacronym{gl:grd}{GRD}{ground range detected}
\newacronym{gl:iw}{IW}{interferometric wide}
\newacronym[user1={Rabus23:shuttleradartopo}]{gl:srtm}{SRTM}{shuttle radar topography mission}
\newacronym{gl:dem}{DEM}{digital elevation model}

\newacronym{gl:vv}{\VVb}{vertical transmit \& vertical receive}
\newacronym{gl:vh}{\VHb}{vertical transmit \& horizontal receive}
\newacronym{gl:eo}{EO}{Earth observation}

\usepackage[colorlinks=true, linkcolor=blue, breaklinks=true, urlcolor=blue]{hyperref}
\usepackage[capitalize,noabbrev,nameinlink]{cleveref} 
\usepackage{xurl} 
\crefformat{appendix}{#2#1#3}  

\crefformat{footnote}{#2\footnotemark[#1]#3}
\crefformat{enumi}{#2\textup{#1}#3}

\DeclareSIUnit[product-units=single]\pixel{px}

\usepackage[capitalize,noabbrev]{cleveref}

\theoremstyle{plain}

\theoremstyle{definition}

\theoremstyle{remark}

\usepackage[textsize=tiny]{todonotes}
\usepackage{bm}
\usepackage{nicefrac}
\usepackage{tumabbrev}
\usepackage{mathabbrev}
\usepackage{upgreek}

\usepackage{pgfplots,pgfplotstable}
\usepgfplotslibrary{groupplots}
\pgfplotsset{compat=newest}
\usepgfplotslibrary{fillbetween}
\pgfplotsset{compat=newest}
\usepackage{tikz}
\usetikzlibrary{calc, positioning}
\usetikzlibrary{shadings}
\usetikzlibrary{shapes, positioning, calc, arrows.meta, shapes.geometric}

\usepackage{multicol}
\usepackage{multirow}
\usepackage{float} 

\usepackage[round, authoryear]{natbib}

\usepackage{subcaption}
\setcitestyle{authoryear, maxnames=2}

\begin{document}

\title{Benchmarking for Practice: Few-Shot Time-Series Crop-Type Classification on the EuroCropsML Dataset}
\date{June 2025}
\author[tum]{Joana Reuss\corref{cor1}}
\author[dida]{Jan Macdonald}
\author[dida,eth]{Simon Becker}
\author[tum]{Ekaterina Gikalo}
\author[dida]{Konrad Schultka}
\author[dida,zib]{Lorenz Richter}
\author[tum,mdsi,ellis]{Marco K\"orner}
\affiliation[tum]{
    organization={Technical University of Munich (TUM), TUM School of Engineering and Design, Department of Aerospace and Geodesy, Chair of Remote Sensing Technology}, 
    city={Munich},
    postcode={80333}, 
    country={Germany}
}
\affiliation[mdsi]{
    organization={Technical University of Munich (TUM), Munich Data Science Institute (MDSI)}, 
    city={Garching},
    postcode={85748}, 
    country={Germany}
}
\affiliation[ellis]{
    organization={ELLIS Unit Jena, University of Jena, Jena, Germany}, 
    city={Jena},
    postcode={07743}, 
    country={Germany}
}
\affiliation[dida]{
    organization={dida}, 
    city={Berlin},
    postcode={10827}, 
    country={Germany}
}         
\affiliation[eth]{
    organization={ETH Zurich, Department of Mathematics}, 
    city={Zurich},
    postcode={8004}, 
    country={Switzerland}
}
\affiliation[zib]{
    organization={Zuse Institute Berlin}, 
    city={Berlin},
    postcode={14195}, 
    country={Germany}
}

\cortext[cor1]{\hspace*{0.1em}Corresponding author. \\
\hspace*{1.8em}\textit{E-Mail address:} \href{mailto:joana.reuss@tum.de}{joana.reuss@tum.de} (J{.} Reuss)}

\begin{abstract}
Accurate crop-type classification from satellite time series is essential for agricultural monitoring. Consequently, various \glsxtrlong{gl:ml} algorithms, aiming on enhancing classification performance on data-scarce tasks, have been developed.
While previous evaluations demonstrated the effectiveness of these algorithms in some situations, these studies frequently lacked real-world scenarios.
Hence, the performance of the algorithms in challenging practical applications has not yet been profoundly evaluated.
To facilitate future research in this domain, we present the first comprehensive benchmark for evaluating supervised and \glsxtrshort{gl:ssl} methods for crop-type classification under real-world conditions.
This benchmark study relies on the \EuroCropsML time-series dataset, which combines farmer-reported crop data with Sentinel-2 satellite observations from Estonia, Latvia, and Portugal.
Our findings indicate that \glsxtrshort{gl:maml}-based meta-learning algorithms achieve slightly higher accuracy compared to supervised transfer learning and \glsxtrshort{gl:ssl} methods.
However, compared to simpler transfer learning, the improvement of meta-learning comes at the cost of increased computational demands and training time.
Moreover, supervised methods benefit most when pre-trained and fine-tuned on geographically close regions.
In addition, while \glsxtrshort{gl:ssl} generally lags behind meta-learning, it demonstrates advantages over training from scratch---particularly in capturing fine-grained features essential for real-world crop-type classification---and also surpasses standard transfer learning.
This highlights its practical value when labeled pre-training crop data is scarce.
Our insights underscore the trade-offs between accuracy and computational demand in selecting supervised machine learning methods for real-world crop-type classification tasks and highlight the difficulties of knowledge transfer across diverse geographic regions.
Furthermore, they demonstrate the practical value of \glsxtrshort{gl:ssl} approaches when labeled pre-training crop data is scarce.
\end{abstract}

\begin{keyword}
transfer and meta-learning; few-shot learning; self-supervised learning; crop-type classification; EuroCropsML

\end{keyword}

\maketitle

\section{Introduction}
\label{sec:introduction}
Satellites continuously generate spatio-temporal remote sensing data of the Earth's surface, which forms a cornerstone of modern \gls{gl:eo} techniques.
They play an important role in a very wide range of applications, from hurricane forecasting \citep{Boussioux_HurricaneForecastingANovelMultimodalMachineLearningFramework} or methane detection \citep{Kumar_2020_methanedetection} to vegetation analysis \citep{odenweller_1984_cropidentification_landsat,reed_1994_phenological_variability}.
In particular, satellite imagery has been used in agriculture for fertilization analysis \citep{santaga_2021_S2_nitrogenmanagement}, harvest detection \citep{markovic_2023_S1_soybeanharvest_detection}, as well as yield forecast and crop assessment \citep{bekmukhamedov_2024_riceyieldforecast}.

A common key challenge in these tasks is the efficient extraction of relevant and valuable information from the abundance of collected remote sensing data.
Recently, \gls{gl:ai}, particularly \gls{gl:ml} and \gls{gl:dl} methods, have shown a significant impact on \gls{gl:eo} and remote sensing, and can address these challenges across entire data processing chains \citep{tuia2023artificial}.

Crop data plays a vital role in supporting decision-making processes in various economic sectors related to food supply, agriculture, the environment, and forestry, as well as socioeconomic development \citep{Gmez2016OpticalRS}.
Analyzing and understanding the distribution of crop types in different regions is essential for making informed decisions about food security, infrastructure development, agricultural management, and precision farming \citep{bekmukhamedov_2024_riceyieldforecast,Gmez2016OpticalRS}. 
Furthermore, variations in crop data are used directly in research on climate change \citep{Song2021MassiveSE}.
While \gls{gl:ml} methods have recently achieved strong results in this domain \citep{QI23:deeplcropclassification,Saini18:cropclassificationRFSVM}, the availability of comprehensive crop data is crucial to fully leverage the potential benefits of data-driven \gls{gl:ml} methods.
However, availability varies significantly across different geographical regions, reflecting a widespread Eurocentric and Amerocentric bias in spatio-temporal datasets \citep{Shankar_2017_geodiversityissues}, which hinders the adoption of ML-based data processing in data-scarce regions until now.

\emph{Transfer} and \emph{meta-learning} approaches offer promising directions for addressing the challenges posed by such data imbalances.
Transfer learning refers to the approach of generalizing insights obtained from one machine learning task (typically data-rich) to another related task (typically data-scarce) in order to improve the performance on the second task.
More broadly, meta-learning refers to methods that attempt to improve the generalization capabilities of machine learning models by considering multiple different but related tasks (each of which individually could be data-scarce). 
A general comparison of the two approaches was recently made by \citet{Dumoulin2021ComparingTA}, see also \cref{sec:related_work} for more details. 

However, labeled crop data remain costly to obtain \citep{Shankar_2017_geodiversityissues}.
This makes learning frameworks that can effectively exploit unlabeled satellite imagery particularly valuable.
\Gls{gl:ssl} addresses this need by enabling models to learn useful data representations without requiring expensive annotations.
In remote sensing, where vast amounts of unlabeled time-series data are readily available, \gls{gl:ssl} appears to be particularly well-suited to uncover complex spatial, spectral, and temporal patterns---making it a compelling alternative in regions with limited or no labeled data.

Despite the growing interest in these approaches, there is currently a notable gap in the thorough analysis and comparison of transfer, meta-, and \glsxtrlong{gl:ssl} techniques for real-world multi-class crop-type classification using spatio-temporal satellite data.
Moreover, the advantages and disadvantages of these techniques have not been thoroughly examined.
In this work, we provide such an analysis.
More precisely, we evaluate the potential of various existing supervised and \glsxtrlong{gl:ssl} approaches for few-shot time-series crop-type classification by benchmarking their performance on the \EuroCropsML dataset \citep{Reuss25:EML}.
The dataset extends the \EuroCrops reference data \citep{schneider_eurocrops_2023} with Sentinel-2 L1C reflectance data and is publicly available \citep{reuss_macdonald_eurocropsml_2024}.
We investigate how both supervised and \gls{gl:ssl}-based pre-training influence classification performance in real-world settings with limited annotated data, and assess the challenges of transferring learned representations across diverse geographical, climatic, and agricultural conditions.
Our study aims to have a realistic trial case to compare and benchmark existing and future crop-type classification algorithms and their performance in few-shot regimes. 
Moreover, it demonstrates how to make use of the \EuroCropsML dataset as a real-world benchmark.

\begin{figure}[t]
    \centering
    \includegraphics[width=0.9\linewidth]{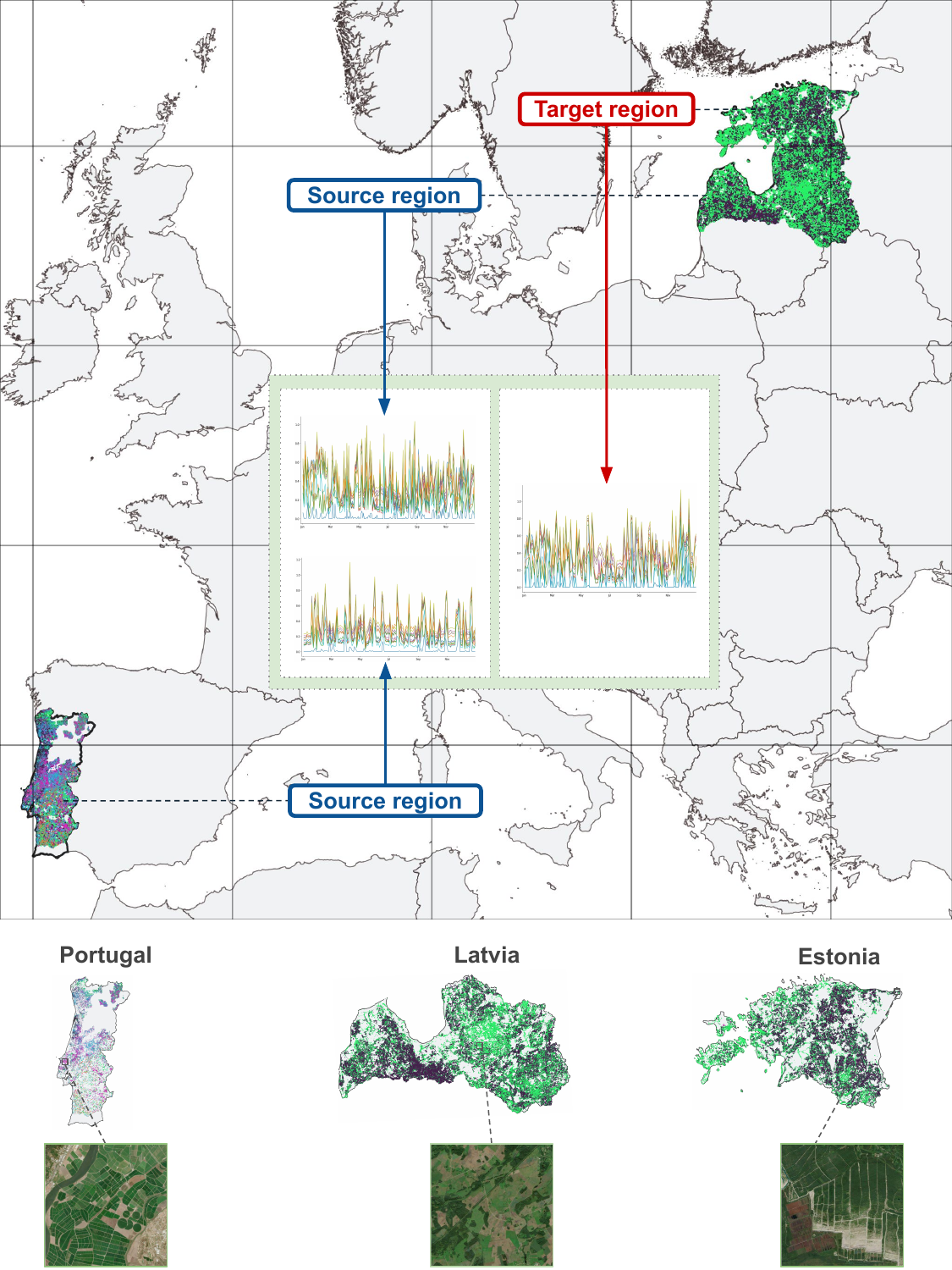}
    \caption{Visualization of crop fields (using \EuroCrops \gls{gl:hcat} level \num{3}) in the source and target region(s).
    The initial supervised pre-training is conducted on the Sentinel-2 L1C agricultural time series of the source region(s), followed by a process of fine-tuning and evaluation of the models on a distinct target region.}
    \label{fig:learningprocess_map}
\end{figure}

Using a state-of-the-art \emph{Transformer} encoder architecture \citep{vaswani_vanillatransformer}, each supervised learning approach is initially trained on a designated source region and subsequently evaluated on a geographically distinct target region, as visualized in \cref{fig:learningprocess_map}. 
This experimental setup enables the assessment of model generalization under domain shifts.

We examine the following approaches, described in more detail in \cref{sec:methodology}: 
\begin{description}
    \item[Transfer Learning] Transfer learning consists of a two-step procedure of pre-training an \gls{gl:ml} model that gets subsequently fine-tuned. 
    \item[\glsxtrshort{gl:maml}] The \emph{Model-Agnostic Meta-Learning} approach introduced by \citet{finn_2017_maml} and further investigated by \citet{Raghu2019RapidLO,Ye_2021_HowTT_maml} is a very flexible and widely used optimization-based meta-learning algorithm (\cf \cref{sec:maml}).
    \item[\glsxtrshort{gl:fomaml}] The \emph{First-Order Model-Agnostic Meta-Learning} algorithm is a computationally more efficient variant of the original \glsxtrshort{gl:maml} framework, proposed by \citet{finn_2017_maml} (\cf \cref{sec:fomaml}).
    \item[\glsxtrshort{gl:anil}] The \emph{Almost No Inner Loop} algorithm introduced by \citet{Raghu2019RapidLO} is another computationally more efficient variant of \glsxtrshort{gl:maml} (\cf \cref{sec:anil}).
    \item[\glsxtrshort{gl:timl}] The \emph{Tasked-Informed Meta-Learning} algorithm was introduced by \citet{tseng_2022_timl} and enriches \glsxtrshort{gl:maml} by including additional task-specific information (\eg geographical coordinates), \cf \cref{sec:timl}.
\end{description}

Furthermore, we compare three \glsxtrlong{gl:ssl} pre-trained \glspl{gl:mae}:
\begin{sloppypar}
\begin{description}
    \item[\glsxtrshort{gl:presto}] The \emph{\textbf{P}retrained \textbf{Re}mote \textbf{S}ening \textbf{T}ransf\textbf{o}rmer} model \citet{Tseng2023LightweightPT} is a pixel-wise \gls{gl:mae} and serves as our \gls{gl:ssl} baseline.
    \item[\glsxtrshort{gl:prestoxts}] Our variant \emph{\glsxtrshort{gl:presto}-E\textbf{X}tended \textbf{T}emporal and \textbf{S}pectral} adapts the original \glsxtrshort{gl:presto} model by expanding its temporal and spectral resolution, \cf \cref{sec:modifications}.
    \item[\glsxtrshort{gl:crossprestoxts}] \emph{Cross-attention \glsxtrshort{gl:prestoxts}} is a computationally more efficient version of \glsxtrshort{gl:prestoxts}, \cf \cref{sec:modifications}. 
\end{description}
\end{sloppypar}

Both of our \gls{gl:ssl} variants are pre-trained from scratch on combined European Sentinel-1 and Sentinel-2 imagery from the \SenMSCRTS dataset, along with \gls{gl:era} reanalysis temperature and precipitation data.
Beyond the original benchmark, for \gls{gl:ssl}, we also analyze the effect of incorporating additional data during fine-tuning, matching the input modalities during pre-training.
In particular, we complement the \EuroCropsML Sentinel-2 Estonia (target region) data with Sentinel-1 radar data and \gls{gl:era} reanalysis temperature and precipitation measurements, as outlined in \cref{sec:exp_finetuning}.

To establish an overall baseline, in addition, we train a randomly initialized Transformer-based attention model directly on the target region data, allowing us to quantify the benefits of knowledge transfer through the above pre-training approaches.

By providing a thorough practical benchmark for few-shot time-series crop-type classification, we aim to support future research and facilitate practical applications in agricultural remote sensing.
In summary, our main findings comparing the aforementioned algorithms on the \EuroCropsML benchmark dataset are: 
\begin{enumerate}
    \item Meta-learning algorithms of the \glsxtrshort{gl:maml} family tend to slightly outperform other algorithms, such as the baseline (no pre-training), transfer learning (standard pre-training), algorithms of the \glsxtrshort{gl:timl} family, and \gls{gl:ssl} methods in terms of prediction accuracy and Cohen's kappa score.
    \item Despite having additional location information available, the algorithms of the \glsxtrshort{gl:timl} family show no improved performance. 
    \item The meta-learning algorithms tend to require longer overall training times and more computational resources compared to the baselines and \glsxtrlong{gl:ssl}.
    \item Meta-learning algorithms require more extensive tuning of hyperparameters, which has a detrimental effect on their marginally superior prediction performance.
    \item Transferring pre-gained knowledge geographically poses a significant challenge for all investigated supervised algorithms.
    \item Pre-training on geographically and spectrally distant data introduces stronger domain shifts, potentially degrading overall accuracy but also regularizing against dominance by specific classes.
    \item \Glsxtrlong{gl:ssl} offers clear advantages over training from scratch---particularly for minority classes---underscoring the potential of \gls{gl:ssl} when labeled pre-training data is unavailable or scarce and highlighting the need for more targeted representation learning approaches in agricultural applications.
\end{enumerate}

\section{Related work}
\label{sec:related_work}
Recent studies have explored both supervised and self-supervised methods for crop-type classification, aiming to reduce reliance on labeled data and enhance model performance across diverse settings.
This paper ties in with a better understanding of the performance of transfer, meta-, and \glsxtrlong{gl:ssl} algorithms on different data diversities specific to remote sensing and crop classification.
It presents a benchmark on a real-world crop-type dataset with different diversity levels, which allows an in-depth analysis of \gls{gl:ml} algorithms for Earth observation.
We start by reviewing different \gls{gl:ml} approaches commonly used for this task before discussing available crop datasets. 

\subsection{Supervised learning}
\label{sec:methods:supervisedlearning}
Supervised pre-training methods leverage large labeled datasets in order to learn generalizable features that can subsequently be transferred to unseen, related tasks, enabling more efficient learning from limited labeled data.

\subsubsection{Transfer learning}\label{sec:transfer}
The main idea of transfer learning is the transfer of knowledge from one task to another related task.
This involves two phases.
In the first phase, the \gls{gl:ml} model is trained on a (large) dataset associated with the first task (pre-training).
In the second phase, the pre-trained model is taken as an initialization for training on a (smaller) dataset associated with the second task (fine-tuning).

Transfer learning methods have been widely used to overcome the imbalance of agricultural data available in different parts of the world. 
For example, models to perform crop yield estimation demonstrated that extrapolating between different countries in South America showed promising results \citep{Wang_2018_cropyieldprediction}. 
\Citet{kerner_2020_cropmaps} generated a cropland map for Togo in less than ten days by developing rapid transfer learning techniques for regions with little to no available ground truth data.
\citet{hao_2020_transferlearning_cropland_layer} trained a random forest-based crop-type classification model on spatio-temporal Sentinel-2 data from areas in the United States and subsequently tested on regions in China and Canada. 
In general, transfer learning models showed promising capabilities in identifying crops, sometimes even without local training samples. 
Furthermore, it was noted that transfer learning techniques typically required longer time-series data compared to instances where sufficient local data was utilized for crop-type classification.
Similarly, \citet{antonijevic_2023_transferlearning_crops} trained models to classify nine different types of crops in Brittany or Serbia while fine-tuning and testing them on data from the other respective area. 
They discovered that the transfer of knowledge from the original domain to the new was effective, both in terms of accuracy and computational efficiency.

\subsubsection{Meta-learning}
\label{sec:meta-learning}
While transfer learning focuses on transferring knowledge and insights gained from one task to another related task, the core concept of meta-learning can be described as \emph{learning-to-learn}. 

In their seminal work, \citet{Thrun1998LearningTL} introduce the concept of meta-learning as the task of learning properties of classes of functions from a few examples, that is, learning entire function spaces. 
On a less abstract level, the objective of meta-learning is to swiftly grasp a new task, a paradigm commonly referred to as \emph{few-shot learning}.
This typically involves adapting a machine learning model across several related tasks while monitoring the respective learning processes.
The goal is then to improve the learning process itself to make a meta-learned model that is well suited for efficient adaptations to future tasks \citep{Rusu_2018_MetaLearning_LatentEmbedds,Yoon_2018_bayesian_maml,Lee_2019_MetaLearningWD,Nichol_2018_fomaml}.
For a contemporary overview, see also the survey of \citet{Hospedales2022MetaLearningSurvey}.

Adapted to our crop classification problem, meta-learning aims to learn about the process of crop classification itself so that it can generalize to new tasks, such as new regions or unseen crops, with minimal effort.
On the other hand, transfer learning uses a pre-trained model and tries to adapt it to classify crops from an unknown region.

Meta-learning approaches can be divided into three groups: 
\begin{description}
    \item[Model-based] 
    These methods are based on fully trained cyclic or recurrent neural network models with internal or external memory that learn to adapt to their state by reading a short sequence of task-specific training data. 
    Examples include \emph{Memory-Augmented Neural Networks} \citep{santoro_2016_memory_augmented_metalearning} and \emph{Neural Attentive Meta-Learners} \citep{mishra_2018_metalearner}.
    \item[Metric-based] These approaches learn effective and task-adapted distance metrics that are combined with non-parametric techniques during inference. 
    Examples of this include \emph{Prototypical Networks} \citep{snell_2017_prototypicalnetworks}, \emph{Matching Networks} \citep{vinyals_2016_matchingnetworks}, and \emph{Relation Networks} \citep{Sung_2018_relation_networks}.
    \item[Optimization-based] These algorithms represent a major research direction within the area of meta-learning, focusing on the optimization process of training \gls{gl:ml} models across different tasks.
    By gaining insights into these learning processes, they enable for a joint optimization of, \eg model hyperparameters and parameter initializations across tasks, allowing for a rapid adaptation to new tasks.
    The most prominent example is \gls{gl:maml}.
\end{description}

In this work, we will focus on the third category, as it is most widely applicable and fairly independent of the chosen \gls{gl:ml} architecture.

\subsubsection{Meta-learning in remote sensing}
Remote sensing encompasses a wide range of problems, such as semantic segmentation (\eg spatially resolved classifications of land cover types), change detection, and object detection.
These often share underlying characteristic structures, making them suitable for meta-learning approaches \citep{Schmitt2021ThereIN}.
In fact, meta-learning has been successfully applied to geospatial data \citep{Wang_2020_metalearning, russwurm_2020_metalearning,tseng_cropharvest_2021,tseng_2022_timl}.

A distinctive feature of remote sensing and Earth observation applications is the association of most data with specific geographic locations. 
These, in turn, are inherently tied to the prevalence of predictive quantities, such as crop type categories. 
Hence, meta-learning methods can learn to take advantage of these prevalences by observing tasks from different locations. 

More specifically, for crop-type classification, there are three main reasons why the problem is predestined for meta-learning algorithms. 

\begin{description}
\item[Seasonal variations] Crop-type classification often involves monitoring crop growth and phenology over time. 
Meta-learning approaches can learn to adapt to temporal variations in crop appearance and spectral characteristics \citep{Wang_2020_metalearning}, allowing a more robust classification in different growing seasons and agricultural cycles.

\item[Source regions] Remote sensing data is collected from various geographical regions. 
The dimension of the source region captures the variability in environmental conditions, terrain types, climate patterns, and land use dynamics. 
Classification models may need to adapt to differences in data distribution between regions, such as rice fields on steep mountain terraces versus flat wheat fields, or mixed cropping systems versus large-scale monoculture farming.

\item[Sensor modalities] Satellite data can be captured using different sensors, each with its own spectral and temporal resolution, as well as other characteristics \citep{Garnot_2020_CVPR}. 
Classification models may need to generalize across different sensor modalities while accounting for differences in data characteristics and noise patterns.
\end{description}

Explicitly addressing the second point, \citet{tseng_2022_timl} introduced the \gls{gl:timl} algorithm to study crop-type classification and yield estimation. 
It is a modification of the \gls{gl:maml} algorithm \citep{finn_2017_maml} that incorporates prior location-dependent task information into the learned \gls{gl:ml} model. 
\Citet{tseng_2022_timl} found that the \gls{gl:timl} algorithm performed best on average compared to four baseline methods on crop-type classification tasks in three different countries (Brazil, Kenya, and Togo) based on the \CropHarvest dataset \citep{tseng_cropharvest_2021}. 
In particular, \gls{gl:timl} outperformed \gls{gl:maml} in all the regions considered.
The baselines considered by \citet{tseng_2022_timl} were a randomly initialized classifier, a classifier pre-trained on the binary \CropHarvest crop-\vs-non-crop identification task (\cf \cref{sec:datasets}), a Random Forest model, and the \gls{gl:maml} algorithm.

In another study, \citet{russwurm_2020_metalearning} observed that meta-learning algorithms can outperform regular pre-training of models on crop type data in a transfer learning setting when the dataset includes a distinct regional diversity.
In the same work, they also found that meta-learning outperformed standard pre-training and randomly initialized baseline models without any pre-training when predicting crop types from heterogeneous sparse labels. 

\subsection{Self-supervised learning}
\label{sec:methods:ssl}
To overcome the reliance on labeled pre-training data for crop-type classification, \gls{gl:ssl} has been explored as an alternative to the aforementioned supervised methods.

One notable approach is \emph{Cropformer} \citep{wang2023cropformer}, a two-step classification model that first undergoes self-supervised pre-training to capture crop growth patterns, followed by supervised fine-tuning.
It demonstrated improvements in accuracy across various scenarios, including full-season and in-season crop classification, as well as regional transferability.

\Citet{wang2022ccssl} introduced \emph{CC-SSL}, an end-to-end \gls{gl:ssl} framework for crop classification based on the novel \emph{Sim-SCAN} contrastive learning algorithm.
By incorporating a domain-specific input tensor transformation module capturing unique spatio-temporal crop growth patterns and sample balancing strategies, they achieved superior results compared to conventional supervised learning.

\citet{isprs-archives-XLVIII-M-1-2023-309-2023} focused on domain adaptation in crop classification, showing \gls{gl:ssl}'s robustness to environmental and management variability across time and space in Germany.

\glspl{gl:mae} for \gls{gl:ssl} were studied by \citet{YUAN2022102651,Cong22:satmae,Fuller23:croma} in the context of remote sensing.
The \gls{gl:croma} algorithm processes both Sentinel-1 and Sentinel-2 data but only considers single observation data without temporal aspects.
First \gls{gl:ssl} time-series models based on \gls{gl:vit} architectures turned out to be computationally expensive \citep{9252123}.
This was addressed by \citet{Tseng2023LightweightPT}, resulting in \gls{gl:presto}, which serves as a comparison baseline for our work.
Its architecture is described in \cref{sec:Presto}.

These methods highlight the effectiveness of \gls{gl:ssl} in leveraging unlabeled data for crop-type classification, particularly under limited label availability and across environmental and agricultural conditions.
However, they also underscore key limitations---most notably, high computational demands.

Building on these insights, we explore \gls{gl:ssl} as a scalable alternative for scenarios where transfer or meta-learning approaches are constrained by a limited availability of labeled pre-training data.
\gls{gl:ssl} can leverage large amounts of unlabeled data to extract generalizable features, potentially improving performance for tasks where labeled data are scarce or entirely absent.

\subsection{Crop datasets}
\label{sec:methods:datasets}
Significant obstacles persist in the development of \gls{gl:ml} platforms that work with agricultural \glsxtrlong{gl:eo} data. 
This is primarily due to different labeling and classification standards, biases in regional coverage, and the scarcity of publicly available labeled data.

The novel dataset \EuroCropsML \citep{reuss_macdonald_eurocropsml_2024,Reuss25:EML}, based on \EuroCrops \citep{schneider_eurocrops_2023,schneider_eurocrops_zenodo}, overcomes this major hurdle by including international data that extends beyond political borders.
This contrasts with national or regional datasets, such as the \Breizhcrops dataset for Brittany (France) \citep{russwurm_breizhcrops} and the \ZueriCrop dataset for the cantons Thurgau and Zurich (Switzerland) \citep{turkoglu_cropmapping}, which provide detailed information only for very local areas.

Although there also exists the \CroplandDataLayer, a crop-type classification dataset published by the United States Department of Agriculture \citep{NASS_cropland_layer_2024}, the labels provided here only have an accuracy slightly above \qty{80}{\percent}, directly affecting the usability for \gls{gl:ml} methods. 
This is not the case for the \EuroCropsML dataset, as it is based on the \EuroCrops labeling, which contains accurate data directly self-reported by the farmers themselves.   

Finally, the \CropHarvest collection \citep{tseng_cropharvest_2021} is a global dataset consisting of \num{90480} data points that unify 20 different smaller datasets containing crop-type classification labels. 
The data is composed of Sentinel-2 top-of-atmosphere (L1C) observations, Sentinel-1 observations, ERA5 climatology data, and topography data from a Digital Elevation Model \citep{tseng_cropharvest_2021}.
However, most of those data points merely distinguish between crop \vs non-crop labels. 
Only \qty{34.2}{\percent} of the data points contain additional agricultural class labels such as, \eg crop types. 
Furthermore, the types of crops in \CropHarvest are generally grouped into only nine different classes \citep[\cf][Figure 2b]{tseng_cropharvest_2021}, namely 
\begin{itemize*}[label={}, itemjoin={{,}}, itemjoin*={{, and}}, after={.}]
  \item \class{cereals}
  \item \class{vegetables and melons}
  \item \class{fruits and nuts}
  \item \class{oilseed crops}
  \item \class{root/tuber crops}
  \item \class{beverage and spice crops}
  \item \class{legu\-mi\-nous crops}
  \item \class{sugar crops}
  \item \class{other crops}
\end{itemize*}
This is in stark contrast to the granular resolution of the hierarchical and harmonized labeling scheme offered by \EuroCrops (\cf \cref{table:EUlist}). 
Moreover, each \CropHarvest data point contains a yearly time series at monthly intervals, which results in a much lower temporal resolution compared to the \EuroCropsML dataset with up to \num{216} time steps (\cf \cref{sec:introduction}).

The diverse geographical coverage, the high number of data points and time steps, as well as the harmonized crop class scheme, were the determining factors in our selection of \EuroCropsML as the dataset for our benchmarking experiments of real-world crop-type classification.

\section{Datasets}
\label{sec:datasets}
We rely on two distinct datasets for our analysis: \EuroCropsML serves as the labeled reference dataset for supervised pre-training, fine-tuning, and evaluating classification performance, while unlabeled \PixelSenMSCRTS data is utilized during the \gls{gl:ssl} pre-training phase to learn foundational data representations prior to task-specific fine-tuning.
Our analysis focuses primarily on Sentinel-2 satellite data.
However, in order to demonstrate the adaptability of the \EuroCropsML dataset and its potential for integration with diverse data sources, we complement it with additional Sentinel-1 and \gls{gl:era} climate variables such as the daily average ground temperature (at \qty{2}{\meter} height) and total daily precipitation.
Together, Sentinel-1 and Sentinel-2 provide a rich, multi-modal and time-resolved view of the Earth’s surface, combining structural and spectral information to improve monitoring and classification tasks in agriculture and beyond.
\gls{gl:era} climate variables complement the satellite imagery by providing information which directly influences crop growth and phenology.

\subsection{The \EuroCrops dataset}
\label{sec:eurocropsml}
\begin{figure}
    \centering
    \includegraphics[width=0.9\linewidth]{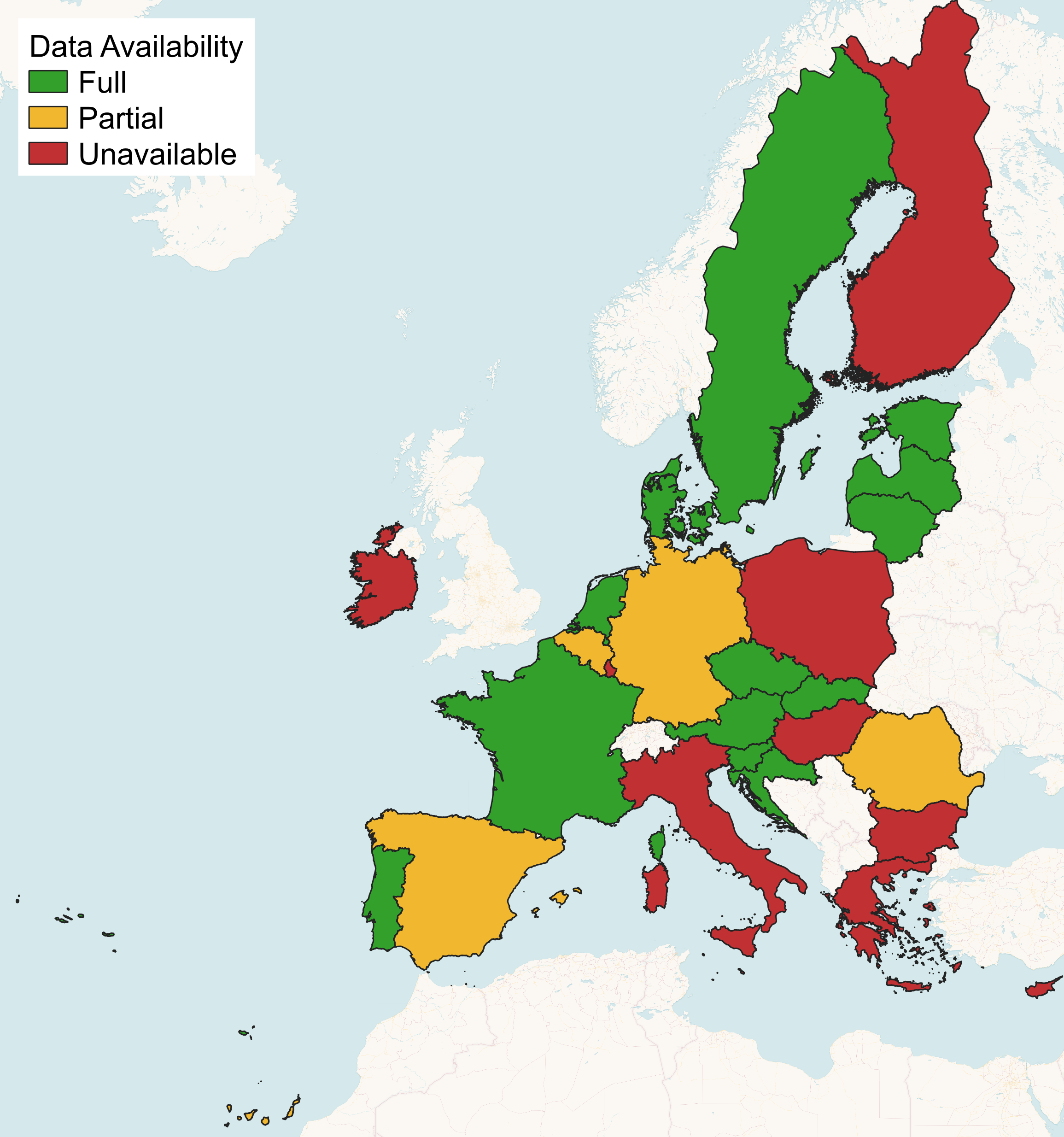}
    \caption{Currently 17 of the 27 member states of the European Union are harmonized within the \EuroCrops dataset, with four states only offering partial data.
    The data availability refers to version 10 of the \EuroCrops dataset \citep{schneider_eurocrops_zenodo}.}
    \label{fig:available_countries}
\end{figure}

\EuroCrops is a reference dataset, introduced by \citet{schneider_eurocrops_2023} containing pan-European agricultural parcel information data and crop types. 
Data have been collected directly from farmers’ self-declarations in accordance with the \gls{gl:cap} of the \gls{gl:eu}. 
Hence, it provides a large collection of reliable, diverse, and unbiased reference data to develop and benchmark new \gls{gl:ml} methods and allows for in-depth investigation of agricultural activities.

\Cref{fig:available_countries} shows an overview of the countries that comprise the full \EuroCrops dataset. \Cref{table:EUlist} provides a more detailed account of these data, showing the availability of data together with the number of crop classes, identified parcels, and available years of coverage.

To overcome country-specific class names and taxonomies for crop types, the dataset introduced the so-called \gls{gl:hcat} to harmonize pan-European standards.

\begin{table}
    \caption{%
        Member states of the EU sorted by the availability of their agricultural reference data. 
        This refers to version 10 of the \EuroCrops dataset \citep{schneider_eurocrops_zenodo}. The \emph{\# crop classes} column refers to the number of unique crop types after \gls{gl:hcat} harmonization.  
        The \emph{\# parcels} column refers to unique parcel geometries after removing potential duplicates. \\
        \textsuperscript{\textdagger}The data for Belgium, Germany, Romania, and Spain is incomplete, meaning only data for selected regions are available. 
        No data is yet available for
        \href{https://github.com/maja601/EuroCrops/wiki/Bulgaria}{Bulgaria},  \href{https://github.com/maja601/EuroCrops/wiki/Cyprus}{Cyprus},
        \href{https://github.com/maja601/EuroCrops/wiki/Finland}{Finland}, \href{https://github.com/maja601/EuroCrops/wiki/Greece}{Greece}, \href{https://github.com/maja601/EuroCrops/wiki/Hungary}{Hungary}, \href{https://github.com/maja601/EuroCrops/wiki/Ireland}{Ireland}, \href{https://github.com/maja601/EuroCrops/wiki/Italy}{Italy}, \href{https://github.com/maja601/EuroCrops/wiki/Luxembourg}{Luxembourg}, 
        \href{https://github.com/maja601/EuroCrops/wiki/Malta}{Malta}, and \href{https://github.com/maja601/EuroCrops/wiki/Poland}{Poland}.
    }
    \label{table:EUlist}

    \centering\scriptsize
    \begin{tabular}[t]{@{}Xrrr@{}}
      \toprule
      \multicolumn{1}{@{}X}{countries} & \multicolumn{1}{X}{\# crop classes}  & \multicolumn{1}{X}{\# parcels}  &  \multicolumn{1}{X@{}}{years}   \\
      \cmidrule(r){1-1} \cmidrule(lr){2-2} \cmidrule(lr){3-3} \cmidrule(l){4-4}
      \href{https://github.com/maja601/EuroCrops/wiki/Austria}{Austria} & \num{98} & \num{2610470} & 2021 \\  
         \href{https://github.com/maja601/EuroCrops/wiki/Croatia}{Croatia} & \num{11} & \num{1377155} & 2020 \\
         \href{https://github.com/maja601/EuroCrops/wiki/Czechia}{Czechia} & \num{142} & \num{665679} & 2023 \\
         \href{https://github.com/maja601/EuroCrops/wiki/Denmark}{Denmark} & \num{127}  & \num{587453} & 2019 \\
         \href{https://github.com/maja601/EuroCrops/wiki/Estonia}{Estonia} & \num{127}  & \num{176055} & 2021\\ 
         \href{https://github.com/maja601/EuroCrops/wiki/France}{France} & \num{149}  & \num{9517878} & 2018 \\
         \href{https://github.com/maja601/EuroCrops/wiki/Latvia}{Latvia} & \num{103} & \num{431174} & 2021 \\ 
         \href{https://github.com/maja601/EuroCrops/wiki/Lithuania}{Lithuania} & \num{22}  & \num{1102441} &2021 \\   
         \href{https://github.com/maja601/EuroCrops/wiki/Netherlands}{Netherlands} & \num{141} & \num{766645} & 2020\\ 
         \href{https://github.com/maja601/EuroCrops/wiki/Portugal}{Portugal} & \num{79}  & \num{100000} & 2021 \\
        \href{https://github.com/maja601/EuroCrops/wiki/Slovakia}{Slovakia} & \num{118}  & \num{255568} & 2021\\ 
         \href{https://github.com/maja601/EuroCrops/wiki/Slovenia}{Slovenia} & \num{99}  & \num{824449}  & 2021\\
         \href{https://github.com/maja601/EuroCrops/wiki/Sweden}{Sweden} & \num{48} & \num{1210533} & 2021 \\ 
      \cmidrule(r){1-1} \cmidrule(lr){2-2} \cmidrule(lr){3-3} \cmidrule(l){4-4}
         \href{https://github.com/maja601/EuroCrops/wiki/Belgium}{Belgium\textsuperscript{\textdagger}} & \num{129} & \num{591083} & 2021 \\
            \smaller\hspace{1em} Flanders & \smaller \num{129} & \smaller \num{591083} & \smaller 2021 \\
       \href{https://github.com/maja601/EuroCrops/wiki/Germany}{Germany\textsuperscript{\textdagger}} & \num{205}  & \num{1921513}  & 2021, 2023 \\
        \smaller\hspace{1em} Brandenburg & \smaller \num{116} & \smaller \num{286246} &  \smaller 2023 \\
        \smaller\hspace{1em} Lower Saxony & \smaller \num{150} & \smaller \num{902441} & \smaller 2021 \\
        \smaller\hspace{1em} NRW & \smaller \num{175} & \smaller \num{732826} & \smaller 2021 \\
       \href{https://github.com/maja601/EuroCrops/wiki/Romania}{Romania\textsuperscript{\textdagger}} & \num{8}  & \num{329706} & N/A \\
       \smaller\hspace{1em} Border region & \smaller \num{8} & \smaller \num{329706} &  \\
  \href{https://github.com/maja601/EuroCrops/wiki/Spain}{Spain\textsuperscript{\textdagger}} & \num{10}  & \num{996679} & 2021  \\
  \smaller\hspace{1em} Navarra & \smaller \num{10} & \smaller \num{996679} & \smaller 2021 \\
\bottomrule
    \end{tabular}
\end{table}

\subsection{The \EuroCropsML benchmarking dataset}
\label{sec:EuroCropsML}
\EuroCropsML \citep{Reuss25:EML} is a time-series dataset that combines the parcel information reference data and multi-class \gls{gl:hcat} labels from \EuroCrops with Sentinel-2 L1C optical satellite observations captured during the year 2021.
Each parcel data point in \EuroCropsML contains an annual time series of cloud-free multi-spectral Sentinel-2 observation data, where each individual time step represents the median pixel value in the geometry of the parcel for each of the \num{13} Sentinel-2
spectral bands.
In addition, the dataset contains further meta-data, including the spatial coordinates of the parcel's centroid and Eurostat's \emph{GISCO} \emph{Nomenclature of Territorial Units for Statistics (NUTS)} regions \citep{Reuss25:EML}.

Tailored specifically for evaluating (few-shot) crop-type classification methodologies, this dataset encompasses agricultural data from three carefully selected European countries, providing a diverse and comprehensive resource for benchmarking ML algorithms within the agricultural domain.

\subsubsection{Study area}
\label{sec:studyarea}
The \EuroCropsML dataset provides data for three European countries as \emph{regions of interest (ROI)}: \textbf{Estonia}, \textbf{Latvia}, and \textbf{Portugal}, as shown in \cref{fig:learningprocess_map}.

\begin{figure}[t]
    \centering
    \resizebox{.43\linewidth}{!}{\definecolor{mediumseagreen}{RGB}{85,168,104}
\definecolor{peru}{RGB}{221,132,82}
\definecolor{steelblue}{RGB}{76,114,176}
\begin{tikzpicture}
\scriptsize
\coordinate (LV) at (0,0);
\coordinate (EE) at (1,0);
\coordinate (cap) at ($(LV)!0.5!(EE)$);

\begin{scope}[blend group=soft light]
    \fill[steelblue, opacity=0.45] (LV) circle[radius=1.5];
    \fill[peru, opacity=0.45] (EE) circle[radius=1.5];
\end{scope}

\draw (LV) circle[radius=1.5];
\draw (EE) circle[radius=1.5];

\node[font=\sffamily] at ($(LV)+(-0.5,-1.5)$) [below,align=center]{Latvia};
\node[font=\sffamily] at ($(EE)+(0.5,-1.5)$) [below,align=center]{Estonia};

\node[font=\sffamily] at (LV) [left=0.75cm,align=center]{22};
\node[font=\sffamily] at (EE) [right=0.75cm,align=center]{46};
\node[font=\sffamily] at (cap) [align=center]{81};
\end{tikzpicture}}
    \hspace{.03\linewidth}
    \resizebox{.43\linewidth}{!}{\definecolor{mediumseagreen}{RGB}{85,168,104}
\definecolor{peru}{RGB}{221,132,82}
\definecolor{steelblue}{RGB}{76,114,176}
\begin{tikzpicture}
\scriptsize
\coordinate (LV) at (210:0.75);
\coordinate (PT) at (330:0.75);
\coordinate (EE) at (90:0.75);
\coordinate (capEELVPT) at (0,0);
\coordinate (capEEPT) at (30:1.15);
\coordinate (capEELV) at (150:1.15);
\coordinate (capLVPT) at (270:1.15);

\begin{scope}[blend group=soft light]
    \fill[steelblue, opacity=0.45] (LV) circle[radius=1.5];
    \fill[mediumseagreen, opacity=0.45] (PT) circle[radius=1.5];
    \fill[peru, opacity=0.45] (EE) circle[radius=1.5];
\end{scope}

\draw (LV) circle[radius=1.5];
\draw (PT) circle[radius=1.5];
\draw (EE) circle[radius=1.5];

\node[font=\sffamily] at ($(LV)+(-0.5,-1.5)$) [below,align=center]{Latvia};
\node[font=\sffamily] at ($(EE)+(0,+1.5)$) [above,align=center]{Estonia};
\node[font=\sffamily] at ($(PT)+(0.5,-1.5)$) [below,align=center]{Portugal};

\node[font=\sffamily] at (210:1.65) [align=center]{17};
\node[font=\sffamily] at (330:1.65) [align=center]{27};
\node[font=\sffamily] at (90:1.65) [align=center]{34};

\node[font=\sffamily] at (capEELV) [align=center]{46};
\node[font=\sffamily] at (capEEPT) [align=center]{12};
\node[font=\sffamily] at (capLVPT) [align=center]{5};

\node[font=\sffamily] at (capEELVPT) [align=center]{35};

\end{tikzpicture}}
    \caption{The number of annotated crop classes that are shared and distinct between the three ROIs: Estonia, Latvia, and Portugal \citep{Reuss25:EML}.}
    \label{fig:overlap_diagrams}
\end{figure}
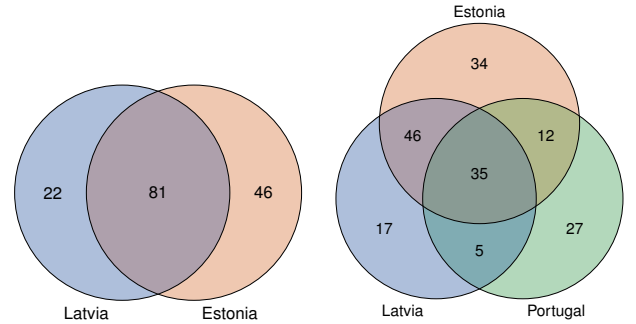

In the context of supervised learning, Estonia and Latvia offer a suitable setting for investigating how similarities between neighboring countries affect knowledge transfer.
Moreover, incorporating data from Portugal enables an examination of how differences in climate conditions, cultivation practices, and crop class distributions influence the transfer of knowledge from one region to a geographically distant one.
This becomes particularly insightful when the geographic location of parcels is explicitly incorporated in \gls{gl:ml} models and algorithms---such as \gls{gl:timl}---since the Baltic states and Portugal, despite being geographically distance, share several crop class occurrences, as illustrated in \cref{fig:overlap_diagrams}.

\begin{table}[t]
    \caption{The number of data points and distinct crop classes for the three countries that comprise the \EuroCropsML dataset. 
    The number of data points refers to the number of unique parcels after pre-processing.}
    \label{tab:LV_PT_datapoints}
    \centering\scriptsize
    \begin{tabular}[t]{@{}Xrr@{}}
      \toprule
      \multicolumn{1}{@{}X}{country} & \multicolumn{1}{X}{\# data points} & \multicolumn{1}{X@{}}{\# crop classes}\\
      \cmidrule(r){1-1} \cmidrule(lr){2-2} \cmidrule(l){3-3}
      Estonia & \num{175906} & \num{127} \\
      Latvia & \num{431143} & \num{103} \\
      Portugal & \num{99634} & \num{79} \\
      \bottomrule
    \end{tabular}
\end{table}

The details of the dataset are shown in \cref{tab:LV_PT_datapoints}.
It consists of a total of \num{706683} multi-class labeled data points with a total of \num{176} distinct classes, \num{35} of which are common to the three ROI countries, \cf \cref{fig:overlap_diagrams}.

\begin{figure}[t]
    \includegraphics[height=3.5cm,]{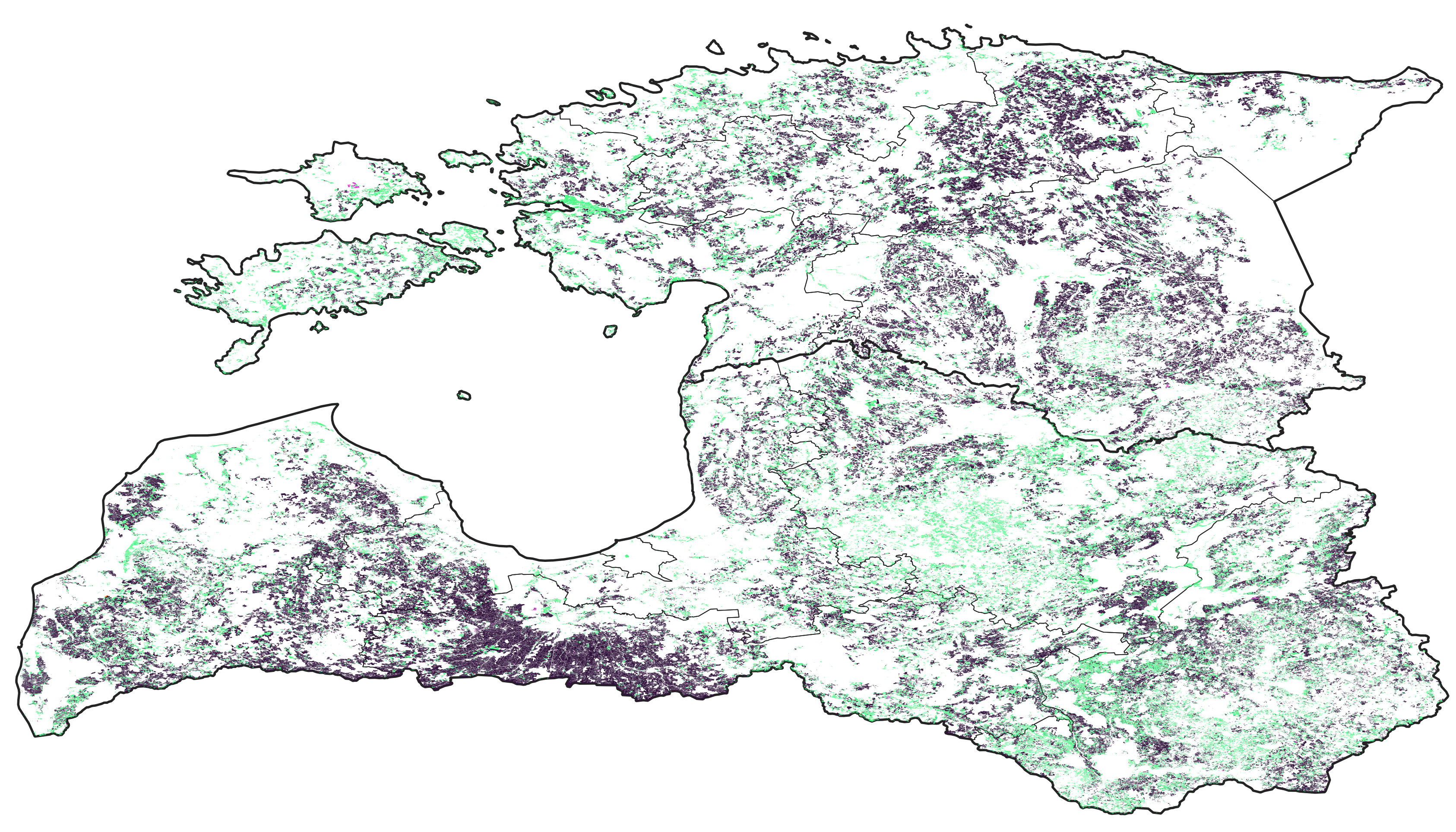}
    \includegraphics[height=3.5cm]{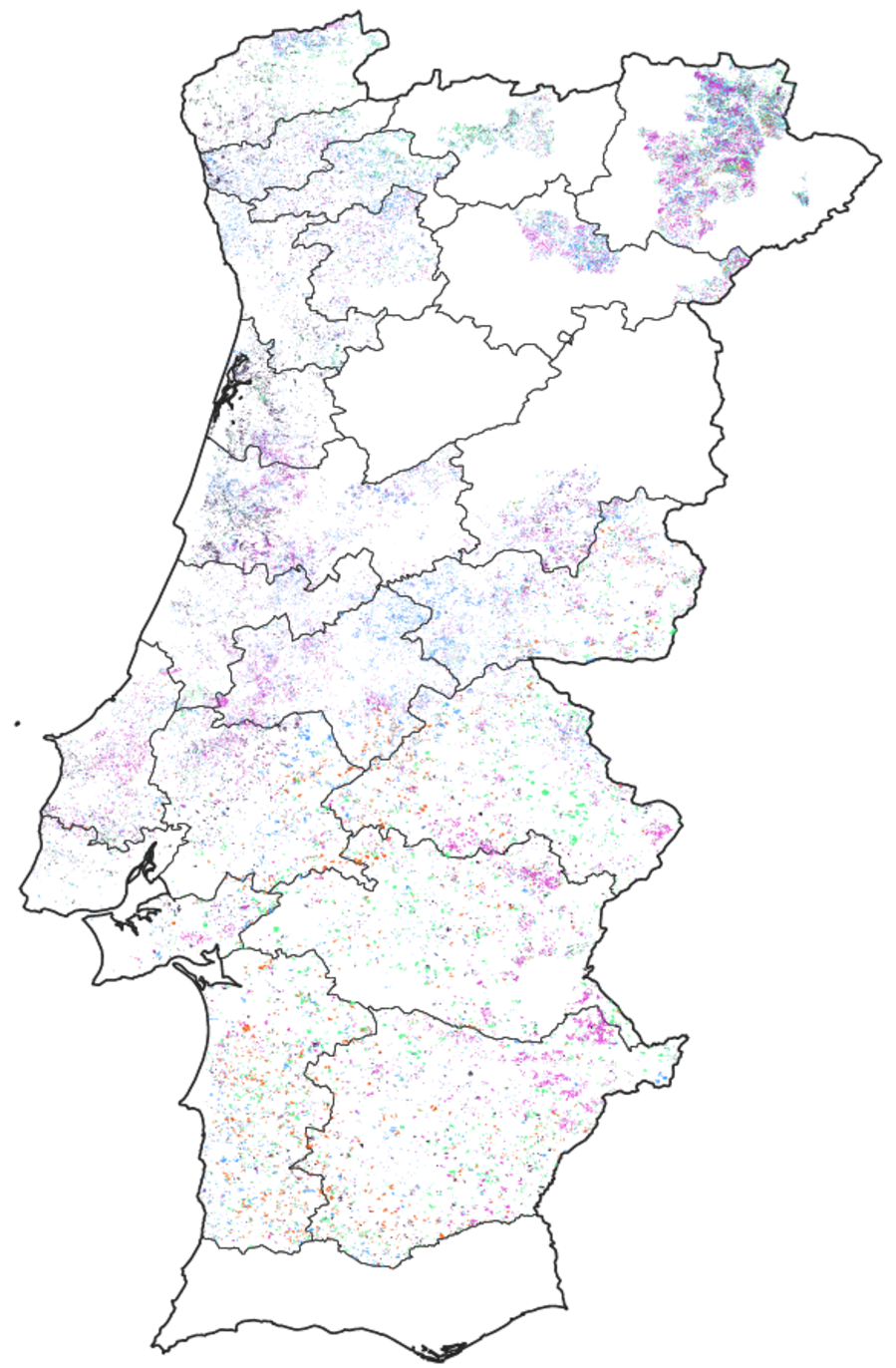}
    \caption{Illustration of the distribution of crop type labels for Estonia and Latvia (left) and Portugal (right), using \EuroCrops \gls{gl:hcat}3 level \num{3}. 
    The differences in the density of available data are clearly visible.}
    \label{fig:map}
\end{figure}

The map shown in \cref{fig:map} visualizes the spatial distribution of crop classes for Estonia, Latvia, and Portugal, clearly unveiling significant differences.
For further illustrations of the different cultivation practices within the \EuroCropsML countries, see \citet[Figures~1-3]{Reuss25:EML}.

Within our benchmark study, we fine-tune each model on the \EuroCropsML Estonia data.
While the \gls{gl:ssl} methods are pre-trained on unlabeled remote sensing data (\cf \cref{sec:pixelsen12mscrts}), we consider the following two scenarios for transfer and meta-learning \citep{Reuss25:EML}:
\begin{description}
    \item[Latvia$\rightarrow$Estonia (\lvee)] The models are pre-trained on Latvian data and subsequently fine-tuned and evaluated on Estonian data. 
    The pre-training dataset contains \num{103} distinct classes. 
    During fine-tuning, the models see \num{127} classes in total, out of which \num{46} were not previously seen during pre-training.
    \item[Latvia+Portugal$\rightarrow$Estonia (\lvptee)] The pre-training is conducted on data from Latvia and Portugal with a total of \num{142} classes, followed by fine-tuning and evaluating on data from Estonia. 
    The fine-tuning dataset contains \num{127} distinct classes, of which \num{34} were not previously seen during pre-training.
\end{description}

\subsubsection{Crop classes}
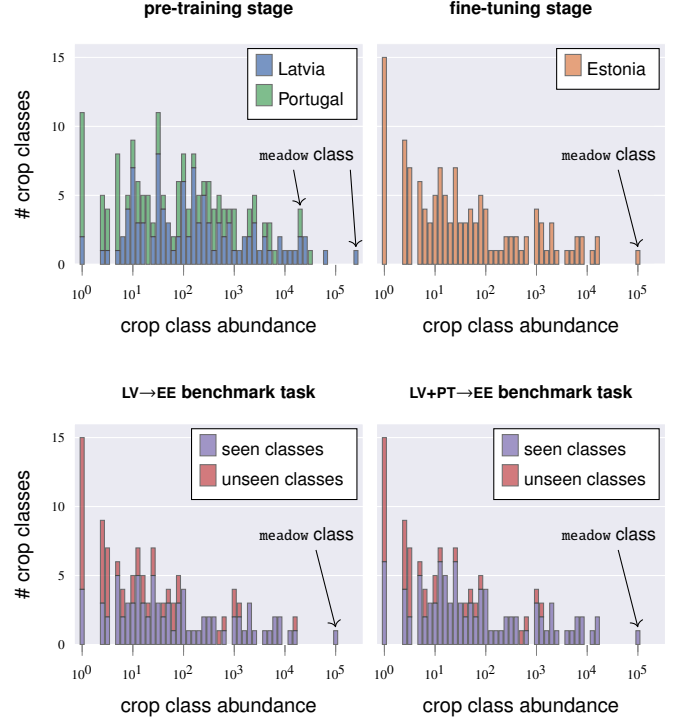
\begin{figure}[t]
    \begin{tikzpicture}

\definecolor{purple}{RGB}{129, 114, 179}
\definecolor{indianred}{RGB}{196,78,82}
\definecolor{lavender}{RGB}{234,234,242}
\definecolor{mediumseagreen}{RGB}{85,168,104}
\definecolor{peru}{RGB}{221,132,82}
\definecolor{steelblue}{RGB}{76,114,176}

\pgfplotstableread[col sep = comma]{images/combined_class_counts_binned.csv}\table
\pgfplotstableread[col sep = comma]{images/combined_overlap_counts_binned.csv}\tabletwo

\pgfplotsset{
every tick label/.append style={font=\sffamily\tiny},
BarPlot/.style={
    axis background/.style={fill=lavender},
    axis line style={white},
    font=\sffamily\footnotesize,
    ybar stacked,
    bar width=0.006\linewidth,
    width=0.61\linewidth,
    height=4.5cm,
    log origin=infty,
    ymin=0.0,
    ymax=16,
    ytick={0,5,10,15},
    ymajorgrids,
    ymajorticks=true,
    ytick style={color=white},
    y grid style={white},
    xtick=data,
    xmin=0.7,
    xmax=350000,
    minor xtick={1,10,100,1000,10000,100000},
    xmajorticks=false,
    xtick pos=left,
    xtick align=outside,
    xmajorticks=false,
    xtick=false,
    legend cell align={left},
    legend style={
        font=\sffamily\scriptsize,
    },
}
}

\begin{groupplot}[group style={group size=2 by 2, vertical sep=6em, horizontal sep=0.5em}]
\nextgroupplot[
BarPlot,
xmode=log,
log base x=10,
xlabel=crop class abundance,
xmajorticks=true,
ylabel=\# crop classes,
title=\textbf{\scriptsize pre-training stage},
]
\addplot[draw=darkgray,fill=steelblue, opacity=0.75] table [x=bins, y=LV] {\table};
\addplot[draw=darkgray,fill=mediumseagreen, opacity=0.75] table [x=bins, y=PT] {\table};
\node[anchor=east] (meadowLVPTtext) at (axis description cs:1.0,0.5) {\scriptsize \class{meadow} class};
\coordinate (meadowLV) at (axis cs:251188.643150958,1.25);
\coordinate (meadowPT) at (axis cs:19952.6231496888,4.25);
\legend{Latvia, Portugal};

\nextgroupplot[
BarPlot,
xmode=log,
log base x=10,
xlabel=crop class abundance,
xmajorticks=true,
yticklabel=\empty,
title=\textbf{\scriptsize fine-tuning stage},
]
\addplot[draw=darkgray,fill=peru, opacity=0.75] table [x=bins, y=EE] {\table};
\node[anchor=east] (meadowEEtext) at (axis description cs:1.0,0.5) {\scriptsize \class{meadow} class};
\coordinate (meadowEE) at (axis cs:100000,1.25);
\legend{Estonia};

\nextgroupplot[
BarPlot,
xmode=log,
log base x=10,
xlabel=crop class abundance,
xmajorticks=true,
ylabel=\# crop classes,
title=\textbf{\scriptsize\lvee benchmark task},
]
\addplot[draw=darkgray,fill=purple, opacity=0.75] table [x=bins, y={LVEEoverlap}] {\tabletwo};
\addplot[draw=darkgray,fill=indianred, opacity=0.75] table [x=bins, y={LVEEnonoverlap}] {\tabletwo};
\node[anchor=east] (meadowEEtext2) at (axis description cs:1.0,0.5) {\scriptsize \class{meadow} class};
\coordinate (meadowEE2) at (axis cs:100000,1.25);
\legend{seen classes, unseen classes};

\nextgroupplot[
BarPlot,
xmode=log,
log base x=10,
xlabel=crop class abundance,
xmajorticks=true,
yticklabel=\empty,
title=\textbf{\scriptsize\lvptee benchmark task},
]
\addplot[draw=darkgray,fill=purple, opacity=0.75] table [x=bins, y={LVPTEEoverlap}] {\tabletwo};
\addplot[draw=darkgray,fill=indianred, opacity=0.75] table [x=bins, y={LVPTEEnonoverlap}] {\tabletwo};
\node[anchor=east] (meadowEEtext3) at (axis description cs:1.0,0.5) {\scriptsize \class{meadow} class};
\coordinate (meadowEE3) at (axis cs:100000,1.25);
\legend{seen classes, unseen classes};

\end{groupplot}

\draw[->] (meadowLVPTtext.340) -- (meadowLV);
\draw[->] (meadowLVPTtext.230) -- (meadowPT);
\draw[->] (meadowEEtext.300) -- (meadowEE);
\draw[->] (meadowEEtext2.300) -- (meadowEE2);
\draw[->] (meadowEEtext3.300) -- (meadowEE3);

\end{tikzpicture}
    \caption{\textbf{Top:} Histograms showing the (stacked) binned distribution of the number of crop classes of a certain abundance (\# of parcels of that class) (in a log scale) in both the pre-training and fine-tuning datasets.\\
    \textbf{Bottom:} Histograms showing the stacked binned distribution of the number of crop classes in the fine-tuning data from Estonia of a certain abundance (\# of samples of that class) (in a log scale) that were previously seen or unseen during the supervised pre-training on either the data from Latvia only or data from Latvia and Portugal.}
    \label{fig:prefine}
\end{figure}

The original composition of the dataset, the abundance of different crop classes, and the distribution of crop classes seen and unseen during the fine-tuning stage are illustrated in \cref{fig:prefine}.
The \meadow class is the predominant class in Latvia (\num{215026} of \num{431143} parcels) and Estonia (\num{84104} of \num{175906} parcels) and the third most frequent class in Portugal (\num{16475} of \num{99634} parcels). 
Hence, for the supervised pre-training stage, we resample it to the median frequency of all other classes.
\Cref{fig:prefine} (\textbf{bottom}) shows that the amount of fine-tuning classes that were previously not seen during pre-training decreases when data from Portugal is added.

\subsubsection{Splits}
For the supervised pre-training stage, the pre-built data splits \citep{Reuss25:EML} allocate \qty{80}{\percent} of the data to training and reserve the remaining \qty{20}{\percent} for validation and hyperparameter tuning.
The first use case (\lvee) yields \num{172993} samples in the training subset and \num{43249} samples in the validation subset.
For the second use case (\lvptee), the training subset consists of \num{239538} data points, while \num{59885} are used for validation.

For the fine-tuning stage, the predefined splits reserve \qty{60}{\percent} of Estonia's data for training (\num{105543} samples) and \num{20}\% for validation and testing, respectively.
From the complete validation set, only \num{1000} fixed final data points are sampled; whereas, for testing, all \num{35182} samples are kept.

\subsubsection{Dataset extension}
To complement the optical Sentinel-2 data, we collect Sentinel-1 C-band \gls{gl:sar} \gls{gl:grd} data for 2021 via the \gls{gl:gee}.
The imagery is selected based on the orbit type (either ascending or descending)  that is available for each location and date, favoring descending when both are available.
We use the \gls{gl:iw} mode with a dual-polarization configuration: \gls{gl:vv} and \gls{gl:vh}.
Following the \EuroCropsML processing of Sentinel-2 data \citep{Reuss25:EML}, we compute the median pixel value of the sigma nought ($\sigma^0$) Sentinel-1 backscatter in decibels for every parcel and temporal observation for both \VV and \VH bands independently.
Additionally, \gls{gl:era} temperature and precipitation data are collected for each parcel and date with available satellite input.

\subsection{Pixel-SEN12MS-CR-TS}
\label{sec:pixelsen12mscrts}
\begin{figure*}[t]
    \centering
    \begin{subfigure}[b]{0.375\linewidth}
        \includegraphics[width=\linewidth]{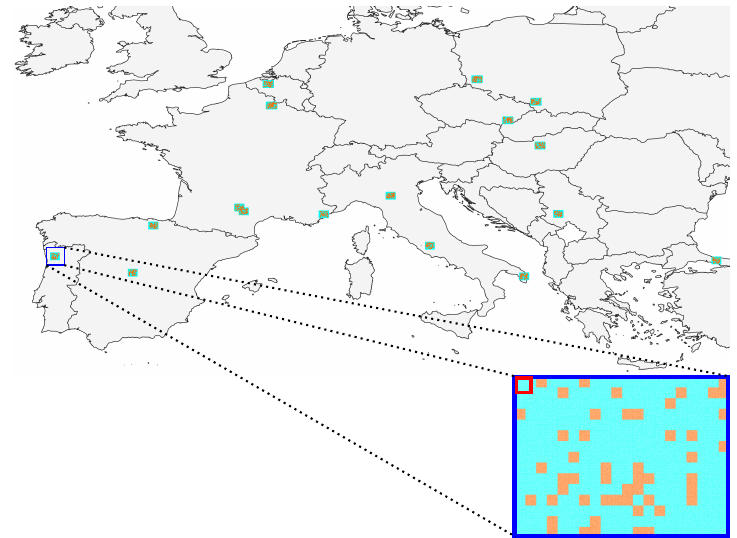}
        \caption{\label{sub@fig:pixelsen12mscrts_full} \PixelSenMSCRTS\ {\color{LightCyan} train} and {\color{LightOrange} validation} patches}
    \end{subfigure}
    \hspace{.012\linewidth}
    \begin{subfigure}[b]{0.27\linewidth}
        \includegraphics[width=\linewidth]{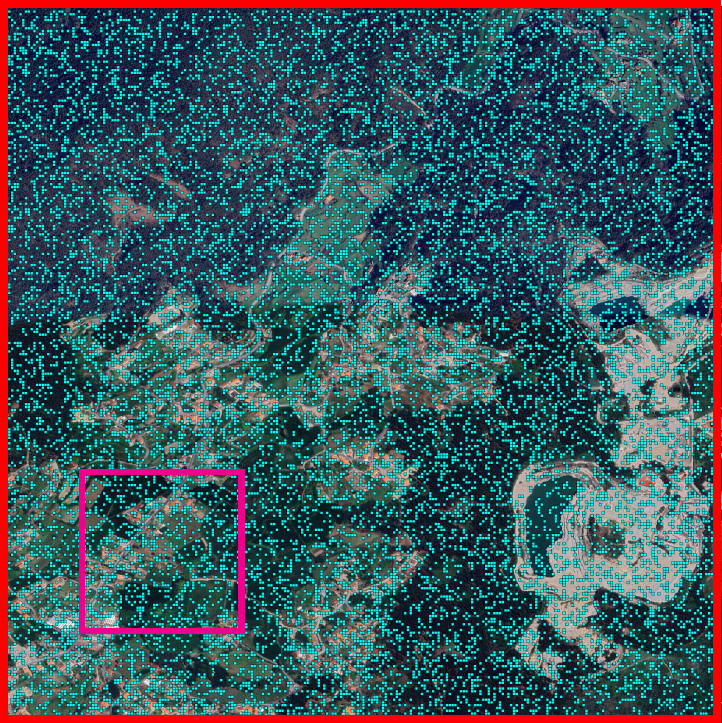}
        \caption{\label{sub@fig:pixelsen12mscrts_trainpatch} Random \PixelSenMSCRTS training patch from Portugal}
    \end{subfigure}
    \hspace{.01\linewidth}
    \begin{subfigure}[b]{0.272\linewidth}
        \includegraphics[width=\linewidth]{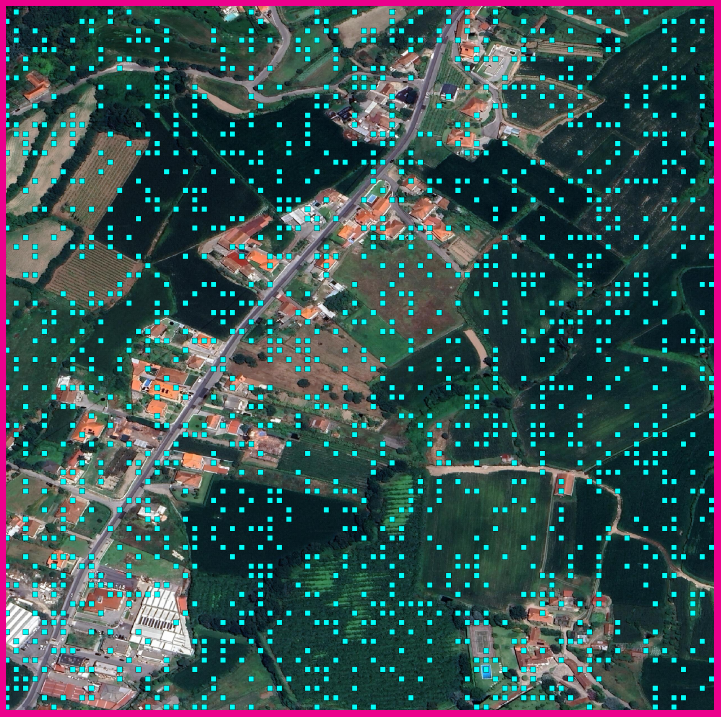}
        \caption{\label{sub@fig:pixelsen12mscrts_trainpatch_zoomed} Zoomed in area of the training patch from Portugal}
    \end{subfigure}
    \caption{\PixelSenMSCRTS dataset.
    \textbf{\color{blue}(a)} The map shows the distribution of {\color{LightCyan} training} and {\color{LightOrange} validation} patches in Europe.
    The {\color{blue}blue} square shows the spatial distribution of the Portugal patches.
    \textbf{\color{blue}(b)} A random training patch from Portugal (\cf {\color{red}red} square in \textbf{\color{blue}(a)}), where each point represents a single sampled pixel.
    \textbf{\color{blue}(c)} The local pixel distribution of a randomly zoomed-in area of the training patch, shown as a {\color{magenta}magenta} square in \textbf{\color{blue}(b)}.
    Again, each sampled pixel is represented by a single point.
    \label{fig:pixel1sen12mscrts_patches}
    }
\end{figure*}
\PixelSenMSCRTS is a comprehensive dataset tailored for pixel-wise \gls{gl:ssl} in remote sensing.
It is derived from the European patches of the public \SenMSCRTS dataset \citep{Ebel22:SEN12MS-CR-TS}, containing multi-temporal Sentinel-1 and Sentinel-2 imagery from 2018.
All Sentinel-2 bands are upsampled to \qty{10}{\meter} resolution.
From the original \num{5520} \qtyproduct{256 x 256}{\pixel} patches, we retain \num{5460} non-overlapping ones, from which we randomly allocate \num{80}\% to training and \num{20}\% for validation.
For pixel-level sampling, \num{25}\% of the pixels per patch are randomly selected, yielding \num{89456640} samples in total, each with up to \num{60} time steps (\num{30} from Sentinel-1 and Sentinel-2 each).
\Cref{fig:pixel1sen12mscrts_patches} shows a visual representation of the dataset's patch and pixel distributions.
\begin{table}[t]
    \caption{Data structure of the \PixelSenMSCRTS dataset.%
    }
    \label{tab:pixelsen12mscrts_variables}
    \centering
    \scriptsize
    \vspace{7pt}

\robustify\bfseries
\sisetup{detect-all=true,uncertainty-mode=separate,table-align-uncertainty=true,round-mode=uncertainty,round-precision=2}

\begin{tabular}{@{}llp{.6\linewidth}@{}}
\toprule
variable type & data source & description \\
\cmidrule(r){1-1} \cmidrule(lr){2-2} \cmidrule(l){3-3}

dynamic  & Sentinel-1 & dual \gls{gl:sar} \VV and \VH polarization channels \\
        & Sentinel-2 & \Btwo -- \Bfour (visible); \Bfive -- \Bseven (RE); \Beight -- \BeightA (NIR), \Bone, \Bnine, \Bten (atmospheric); \Beleven -- \Btwelve (SWIR)\\
        & NDVI & computed from Sentinel-2 \Bfour (red) and \Beight (NIR) bands\\
        & \gls{gl:era} & ERA5 reanalysis data with daily temperature and precipitation \\
        \addlinespace
        static & location  & 3D coordinates $(x,y,z)$ from geographic latitude and longitude in Cartesian space\\   
\bottomrule
\end{tabular}

\end{table}
We supplement the satellite data with \gls{gl:era} temperature and precipitation data, which we bilinearly interpolate to match the spatial resolution of the Sentinel data.
In addition, we calculate the \gls{gl:ndvi} from the Sentinel-2 bands \Bfour and \Beight, which is widely used as a general indicator for vegetation cover and health.
A summary of the data structure similar to the one introduced by \citet{Tseng2023LightweightPT}, \cf \cref{sec:Presto}, is shown in \cref{tab:pixelsen12mscrts_variables}. 
This dataset is used for pre-training our own \gls{gl:presto} variants, \cf \cref{sec:experiments}.

\section{Methodology}
\label{sec:methodology}
In this section, we describe the \gls{gl:ml} models and the different learning algorithms that are examined in our benchmark study using the \EuroCropsML dataset.

\subsection{Transformer model}
For all experimental settings, \cf \cref{sec:experiments}, we rely on a state-of-the-art Transformer encoder architecture with sinusoidal positional encoding \citep{vaswani_vanillatransformer,Schneider21:SIT}. 
We set the maximum sequence length of input data to \num{366} days (\ie a full year, including one leap day). 
For supervised learning, this encoder model is combined with a single linear layer that maps Transformer encodings to the crop-type class logits. 
For all \gls{gl:ssl} methods, we combine the encoder with a transformer decoder.
Further details of this architecture are outlined in \cref{sec:appendix_networkarchitecture}.

We refer to the encoder part as the \emph{backbone} of the model and denote it as
\begin{align*}
  M_\text{backbone}\left[\boldsymbol{\theta}_\text{backbone}\right] \colon \mathbb{R}^{n_b\times n_t}\to\mathbb{R}^{n_e},
\end{align*}
where $n_b \in \mathbb{N}$ and $n_t \in \mathbb{N}$ denote the number of bands (channels) and time steps in the multispectral temporal input data, respectively, and $n_e \in \mathbb{N}$ denotes the Transformer embedding dimension.
All trainable model parameters of the backbone are contained in the parameter vector $\boldsymbol{\theta}_\text{backbone}$.
Similarly, we refer to the classification layer during supervised learning and the \gls{gl:ssl} decoder as the \emph{head} of the model.
We denote it as
\begin{align*}
  M_\text{head}\left[\boldsymbol{\theta}_\text{head}\right] \colon \mathbb{R}^{n_e} \to \mathbb{R}^{n_c},
\end{align*}
where, for supervised learning, $n_c \in \mathbb{N}$ denotes the number of distinct classes, and for \gls{gl:ssl} the Transformer decoder embedding dimension.
As before, $\boldsymbol{\theta}_\text{head}$ collects all trainable model parameters of the head.
Note that while the model backbone architecture remains unchanged for all our experiments, the number of classes and, thus, the linear classification head changes between different pre-training, fine-tuning, and meta-learning tasks.
Moreover, when fine-tuning a Transformer encoder-decoder architecture, we replace the decoder head with a linear classification layer (\cf \cref{sec:methods:ssl}).

The complete end-to-end model for a specific task is given by the composition
\begin{align*}
  M[\boldsymbol{\theta}] &= M_\text{head}\left[\boldsymbol{\theta}_\text{head}\right] \circ M_\text{backbone}\left[\boldsymbol{\theta}_\text{backbone}\right],
\end{align*}
where $\boldsymbol{\theta}=\left[\boldsymbol{\theta}_\text{backbone},\boldsymbol{\theta}_\text{head}\right]$ represents all trainable parameters from both the model backbone and head.

\subsection{Supervised Meta-learning}
\label{sec:methods:meta-learning}

Meta-learning, as introduced by \citet{Thrun1998LearningTL}, see also the recent overview of \citet{Huisman2020ASO}, aims to implement the concept of \emph{learning-to-learn}.
In other words, meta-learning uses a variety of smaller related tasks to extract general information about a learning process.
The goal is to obtain a meta-learned model that is able to quickly adapt to new and unknown scenarios and, thus, can learn an unseen task fast and efficiently.

In the following, we focus on optimization-based meta-learning methods, which can be formulated as bi-level optimization problems consisting of nested \emph{inner} and \emph{outer} optimization loops.

To illustrate the main ideas, consider a distribution $p(\mathcal{T})$ over a set of related tasks $\mathcal{T} = \{\tau^{(1)}, \tau^{(2)}, \ldots, \tau^{(N)}\}$, where each task $\tau^{(i)}=(\mathcal{D}^{(i)}_{\text{support}}, D^{(i)}_{\text{query}}, \mathcal{L}^{(i)}) \in \mathcal{T}$ consists of a set of support (training) data points $\mathcal{D}^{(i)}_{\text{support}}$, a set of query (test) data points $\mathcal{D}^{(i)}_{\text{query}}$, and a loss function $\mathcal{L}^{(i)}$.

During \emph{inner optimization}, we adapt a parametrized model $M\left[\boldsymbol{\theta}\right]$ with learnable parameters $\boldsymbol{\theta}$ on the support dataset by optimizing 
\begin{align*}
  \boldsymbol{\theta}^{(i)}_\text{opt} 
  &= \mathcal{A} \left(M,\mathcal{D}^{(i)}_{\text{support}},\mathcal{L}^{(i)},\boldsymbol{\omega} \right)  \\
  &\approx \operatornamewithlimits{argmin}_{\boldsymbol{\theta}}\mathcal{L}^{(i)} \left(M\left[\boldsymbol{\theta}\right], \mathcal{D}^{(i)}_{\text{support}} \right),
\end{align*}
where $\mathcal{A}$ is the chosen training algorithm, \eg mini-batch \emph{stochastic gradient descent (SGD)}. 
This algorithm depends on hyperparameters $\boldsymbol{\omega}$, such as the learning rate or the initialization $\boldsymbol{\theta}_0$ of the model parameters.

The resulting model can be evaluated for the task $\tau^{(i)}$ on the query dataset according to the corresponding loss function $\mathcal{L}^{(i)}(M[\boldsymbol{\theta}^{(i)}_\text{opt}], \mathcal{D}^{(i)}_\text{query})$.

The goal of the \emph{outer optimization} is then to improve these task evaluations by adapting the hyperparameters $\boldsymbol{\omega}$. 
The optimal values for all tasks are found by (approximately) identifying
\begin{align*}
  \boldsymbol{\omega}_\text{opt} &= 
  \operatornamewithlimits{argmin}_{\boldsymbol{\omega}} 
  \mathop{\mathbb{E}}_{\tau^{(i)}\sim p(\mathcal{T})} \left[
  \mathcal{L}_\text{meta} \left(%
    \mathcal{A}\left(M,\mathcal{D}^{(i)}_{\text{support}},\mathcal{L}^{(i)},\boldsymbol{\omega} \right), \mathcal{D}^{(i)}_\text{query} 
  \right) \right],
\end{align*}  
where $\mathcal{L}_\text{meta}$ is some suitable meta-learning loss function.

One common meta-learning framework is \emph{$n$-way $k$-shot classification} \citep{Hospedales2022MetaLearningSurvey}, which instantiates a specific variant of few-shot learning. 
In this case, each of the task training sets $\mathcal{D}^{(i)}_\text{support}$ consists of $n \cdot k$ data points from $n$ distinct classes ($k$ data samples per class) and the same classification loss $\mathcal{L}^{(i)}$, \eg the softmax cross-entropy loss, is used for all tasks.

In our experiments, \cf \cref{sec:experiments}, we follow this $n$-way $k$-shot setup.
We use a softmax cross-entropy classification loss for the inner optimization over the support sets. 
For the meta objective $\mathcal{L}_\text{meta}$, we use an averaged evaluation softmax cross-entropy classification loss on the query sets.

\subsubsection{MAML}
\label{sec:maml}
Building on this general concept, a widely adopted meta-learning approach is \gls{gl:maml}.
Specifically, it focuses on finding optimal initializations of the model that can quickly adapt to new tasks, measured by the number of training steps.
In other words, the outer optimization over hyperparameters is restricted to the initialization of the model weights.
Further, \gls{gl:maml} relies on iterative gradient-based optimization methods, such as mini-batch \gls{gl:sgd}, for both the inner and the outer optimizations. 
Here, the number of adaptation steps $s$ for inner optimization is chosen as a fixed, typically rather small, constant. 
The respective step sizes (learning rates) for the two optimization iterations are consequently referred to as the inner and outer learning rates.

The interplay between these learning rates is complex, making it challenging to find suitable combinations in the \glsxtrshort{gl:maml} algorithm.
Different approaches addressing this include the choice of a very fine-grained separate learning rate for each task in the inner optimization \citep{li_2017_metasgd}, adaptive learning of inner and outer learning rates \citep{singhbehl_2019_alphamaml}, as well as the training of a separate model to generate per-layer learning rates \citep{baik_2020_metalearning_adaptivehypers}.

\subsubsection{FOMAML}
\label{sec:fomaml}
Due to the nested structure of the \gls{gl:maml} bi-level optimization, obtaining gradients for the outer optimization involves the computation of second-order derivatives with respect to the model parameters. This comes with significant computational costs.

Discarding higher-order derivatives by setting them equal to zero, we arrive at a variation of \gls{gl:maml} with reduced computational costs.
This is called \gls{gl:fomaml}.

\subsubsection{ANIL}
\label{sec:anil}
The \gls{gl:anil} algorithm is another variation that streamlines the \gls{gl:maml} algorithm in order to reduce its computational cost.
This is achieved by restricting the inner optimization of \gls{gl:maml} to only adapt the parameters $\boldsymbol{\theta}_\text{head}$ associated with the classification head of the model while keeping the backbone parameters $\boldsymbol{\theta}_\text{backbone}$ fixed.

This was motivated by the observations of \citet{Raghu2019RapidLO} that for \gls{gl:maml}, in many cases, the backbone parameters remained relatively stable across tasks. 
In their analysis for natural image classification tasks, the \gls{gl:anil} algorithm typically achieved a performance comparable to \gls{gl:maml}.

\subsubsection{TIML}
\label{sec:timl}
The \gls{gl:timl} algorithm represents an extension of \gls{gl:maml}.
It incorporates task-specific information, such as geospatial locations of a region or parcel, in the form of cartesian coordinates.
More precisely, given the geographic coordinates of a surface point $\boldsymbol{p}_\text{polar} = (\varphi, \lambda) \in \left[ -\tfrac{\uppi}{2}, \tfrac{\uppi}{2} \right] \times \left[ -\uppi, \uppi  \right]$  represented by its spherical longitude $\varphi$ and latitude $\lambda$, spatial information is encoded by mapping it into a normalized three-dimensional Cartesian coordinate space 
\begin{align*}
  \boldsymbol{p}_\text{cart} &= 
  \left( \cos(\lambda) \cdot \cos(\varphi), \cos(\lambda) \cdot \sin(\varphi), \sin(\lambda) \right)^\top \in \mathbb{R}^3.
\end{align*}

The original \gls{gl:timl} algorithm only includes the spatial coordinates at the task level, more precisely, the central coordinates of the region or county associated with each task.
Furthermore, \citet{tseng_2022_timl} propose to add a one-hot encoded vector of ten higher-level crop categories to the task information. 
Thus, the task information remains constant within each individual task but changes across different tasks. 
We decided to make the following adjustments to \gls{gl:timl} in our benchmark study:
\begin{enumerate}
    \item We do not incorporate additional high-level information on crop categories since we replace the model head $M_\text{head}$ to always exactly match the classes available in any task during each inner optimization. 
    Hence, the model does not need more information on available crop categories.
    \item We consider fine-grained spatial coordinates at the parcel level: We incorporate the coordinates of the centroid of each individual parcel present in a task instead of using the center of the county. 
    The rationale behind this choice is that utilizing the location only at the county level can potentially result in spatial oversimplification.
    To illustrate, consider the case where a county exhibits an uneven distribution of crop classes, with certain crop-type clusters in the North and others in the South. Incorporating the same location information to all parcels would fail to resolve these distinct clusters.
\end{enumerate}
This leaves us with a unique 3-dimensional task information vector for each data point.
This is either directly concatenated with the remaining time series (no encoder) or encoded into specific layers of the model architecture (encoder); see also \citet{tseng_2022_timl} for details. 
For the latter, these encodings are treated as learnable model parameters during outer optimization.
In accordance with the approach outlined in \citet{tseng_2022_timl}, we apply them once prior to the model backbone and once prior to the model head.
In contrast to the model architecture originally considered by \citet{tseng_2022_timl}, we only use one layer for the classification head, while they allow multiple layers and apply the task encoding before each of them.

Additionally, when using the task encoder, we pad all input sequences to \num{366} days and mask out the padded values since the encoder requires fixed length input sequences.
All remaining parameters can be obtained from \citet{tseng_2022_timl}.

Due to the less streamlined nature of the \CropHarvest dataset considered by \citet{tseng_2022_timl}, the authors also employ a forgetful meta-learning routine. 
This is proposed to overcome the challenge that the majority of \CropHarvest data points only allow for a binary classification into crop-\vs-non-crop categories, which creates a bias for meta-learning algorithms toward learning only this specific task. 
However, we skip the incorporation of forgetful meta-learning, since the nature of the \EuroCropsML dataset already inherently provides room for extensive and diverse multi-class tasks.

\subsection{Self-supervised learning}
\label{sec:ssl_methods}
We compare various lightweight variations of \glspl{gl:mae} in this study.
As our baseline, we adopt the \gls{gl:presto} model \citep{Tseng2023LightweightPT}, an efficient time-resolved \gls{gl:mae} designed for remote sensing applications.
Building on this, we introduce several modifications aimed at improving temporal and spectral resolution, as well as computational efficiency.

\subsubsection{Masked autoencoders}
\label{sec:maes}
\Glspl{gl:mae} are \gls{gl:vit}-based models for \gls{gl:ssl}.
They consist of an encoder-decoder architecture in which random patches of input images are masked out for the encoder, to later be reconstructed from the decoder.
\Glspl{gl:mae} are trained using a regression loss, \eg the \gls{gl:mse}, matching decoder reconstructions against original unmasked inputs.

After \gls{gl:ssl} pre-training, \glspl{gl:mae} can be used for various downstream tasks. 
For this, typically only the encoder part of the \gls{gl:mae} is retained while the decoder part is dropped and replaced by a task-specific partial model.
For our classification task this means that after pre-training, we replace the initial decoder with a simple linear classification layer which is subsequently fine-tuned to predict the class log-probabilities.

\subsubsection{Cross-attention masked autoencoders}
\label{sec:crossmae}
\citet{fu2024_crossMAE} investigated the necessity of masked tokens interacting via self-attention during the reconstruction process in \glspl{gl:mae}, versus solely relying on cross-attention between masked and unmasked tokens. 
They found that \gls{gl:mae} decoders can learn strong global representations from unmasked inputs alone, making masked-to-masked interactions unnecessary.
Consequently, they introduced \gls{gl:crossmae}, which replaces the decoder's self-attention with a simpler, computationally less expensive cross-attention mechanism.
This allows the decoder's masked tokens (queries $Q$) to attend exclusively to visible tokens (keys $K$ and values $V$) from the encoder's output.

\subsubsection{The PRESTO model}
\label{sec:Presto}
Adapting the \glspl{gl:mae} framework to pixel-wise time-resolved remote sensing data, \gls{gl:presto} \citep{Tseng2023LightweightPT} follows the encoder-decoder architecture with masking but processes individual pixel time-series data instead of static image patches.
It leverages domain-specific features, such as temporal structure and multi-modal data, as found in datasets like \PixelSenMSCRTS.
\citet{Tseng2023LightweightPT} observed that \gls{gl:presto} could match the performance of much larger models while being significantly more computationally efficient.
The main strengths of \gls{gl:presto} are:
\begin{description}
    \item[Parameter efficiency:]
    \gls{gl:presto} achieves accuracy on par with large-scale models like \gls{gl:vit} or \gls{gl:resnet}, while requiring up to \num{1000} times less learnable parameters.
    \item[Versatility across remote sensing sources:] Trained on diverse inputs (\eg Sentinel-1 and Sentinel-2), \gls{gl:presto} can handle data from multiple sensors concurrently.
    \item[Robustness to missing data:]
    During the masked pre-training, \gls{gl:presto} adapts to process incomplete input data, such as data with temporal or spatial gaps, which makes it more versatile for different downstream tasks.
\end{description}

The \gls{gl:presto} model is pre-trained using the same combination of dynamic (time-resolved) data (Sentinel-1, Sentinel-2, \gls{gl:ndvi}, \gls{gl:era}) and static data (location) as described in \cref{sec:pixelsen12mscrts}, with the addition of nine \emph{Dynamic World Land Cover} classes \citep[\ie \class{water}, \class{trees}, \class{grass}, \class{crops}, \class{shrub \& scrub}, \class{flooded vegetation}, \class{built-up area}, \class{bare ground}, and \class{snow \& ice}; \cf][]{Brown22:DW} for each non-cloudy Sentinel-2 observation, and (static) \emph{topography data}, specifically elevation and slope, derived from the \gls{gl:srtm} \gls{gl:dem}.
Moreover, from the \num{13} Sentinel-2 bands, the bands \Bone, \Bnine, and \Bten are removed and not processed by the \gls{gl:presto} model.

\paragraph{Input encoding}
In order to handle different variable types (categorical and numerical) and varying numbers of input channels, \gls{gl:presto} introduces a specialized unified input encoding pipeline:
\begin{sloppypar}
\begin{description}
    \item[Channel grouping:]
    The input variables are divided into $C$ data source (\eg sensor) specific channel groups, with $\sum_{c=1}^{C}d_c = D$, where $D$ is the total number of channels and $d_c$ is the number of channels belonging to group $c$.
    All Sentinel-1 bands, \eg form one group.
    See the original work of \citet{Tseng2023LightweightPT} for the detailed description of the channel groups.
    \item[Channel group embedding:]
    For each channel group $c$, the input data with $d_c$ channels is passed through a learned linear projection $h^c\colon\R^{d_c}\to\R^{d_{\text{emb}}}$, mapping the group's channels into a shared latent embedding space of dimension $d_{\text{emb}}$. 
    For dynamic channel groups, this transformation is independently applied to all time steps.
    Categorical variables (such as land cover class) are implicitly one-hot-encoded, making the linear transformation equivalent to a learned embedding matrix.
    \item[Contextual encodings:]
    Where applicable, additive encodings are used for token contextualization:\\
    \textit{Temporal position encoding:}
     A sinusoidal function $p_\text{sin}\colon\{1,\dots,T\}\to\R^{d_\text{sin}}$, where $d_\text{sin}=\frac{1}{2}\cdot d_\text{emb}$, and $T$ denotes the number of all time steps, encodes each token's temporal position in the time series, as common in Transformer models \citep{vaswani_vanillatransformer}.\\
    \textit{Month encoding:} Similarly, a sinusoidal encoding $p_\text{month}\colon\{1,\dots,T\}\to\R^{d_\text{month}}$, with $d_\text{month}=\frac{1}{4}\cdot d_\text{emb}$, that only depends on the month of an observation's time step is intended to capture seasonal patterns.\\
    \textit{Channel group encoding:} Finally, a learned embedding $p_\text{channel}\colon\{1,\dots,C\}\to\R^{d_\text{channel}}$, with $d_\text{channel}=\frac{1}{4}\cdot d_\text{emb}$, indicates the channel group (\ie data source) of each token.\\[0.3em]
    The full contextual embedding dimensions are chosen to sum up to the same shared embedding dimension, \ie $d_\text{sin}+d_\text{month}+d_\text{channel} = d_\text{emb}$.
\end{description}
\end{sloppypar}
    
A full input example is constructed by combining the previously defined embeddings as follows:
For a dynamic variable channel group $c$, the raw data $t^c_i$ for a time step $i$ is encoded as 
$h^c(t^c_i) + [p_\text{channel}(c); p_\text{sin}(i); p_\text{month}(i)] \in \R^{d_\text{emb}}$.
For static variable groups, the temporal encodings  are set to zero, since these variables are independent of time, resulting correspondingly in $h^c(t^c) + [p_\text{channel}(c); \bfzero; \bfzero] \in \R^{d_\text{emb}}$.
Finally, all embeddings for static variable groups and dynamic variable groups are concatenated to form the input $\bfx\in\R^{(C_\text{static} + C_\text{dynamic}\cdot T)\times d_\text{emb}}$, where $C_\text{static}$ and $C_\text{dynamic}$ denote the number of static and dynamic groups and together sum to the total number of groups $C_\text{static} + C_\text{dynamic} = C$.

\paragraph{Masking}
Given the inherent challenges of remote sensing data, which is often partially missing information, \eg due to sensor noise or cloud coverage, \gls{gl:presto} employs four structured masking strategies during pre-training aiming at masking a fraction of \SI{75}{\percent} of input tokens:
\begin{description}[itemsep=0.1em, topsep=0.2em, style=sameline, wide]
    \item[Random:] Masking random pairs of time steps and channels.
    \item[Channel groups:] Masking a random subset of channel groups. In the case of dynamic channel groups, this will mask all time steps from the groups.
    \item[Contiguous time steps:] Mask a random subset of consecutive time steps across all dynamic channels.
    \item[Random time steps:] Mask a random subset of time steps across all dynamic channels.
\end{description}

In the context of channel group and time step masking, if not exactly \SI{75}{\percent} of all tokens can be masked with these strategies, a slightly smaller fraction is selected first, and the remaining difference is randomly masked.
Using a random combination of all four masking strategies in the encoder-decoder architecture, \gls{gl:presto} is trained exactly as \glspl{gl:mae}, using an \gls{gl:mse} regression loss matching the decoder reconstructions against the original unmasked input.

\subsubsection{Modifications to PRESTO}
\label{sec:modifications}
\begin{figure*}
    \centering
    \includegraphics[width=0.9\linewidth]{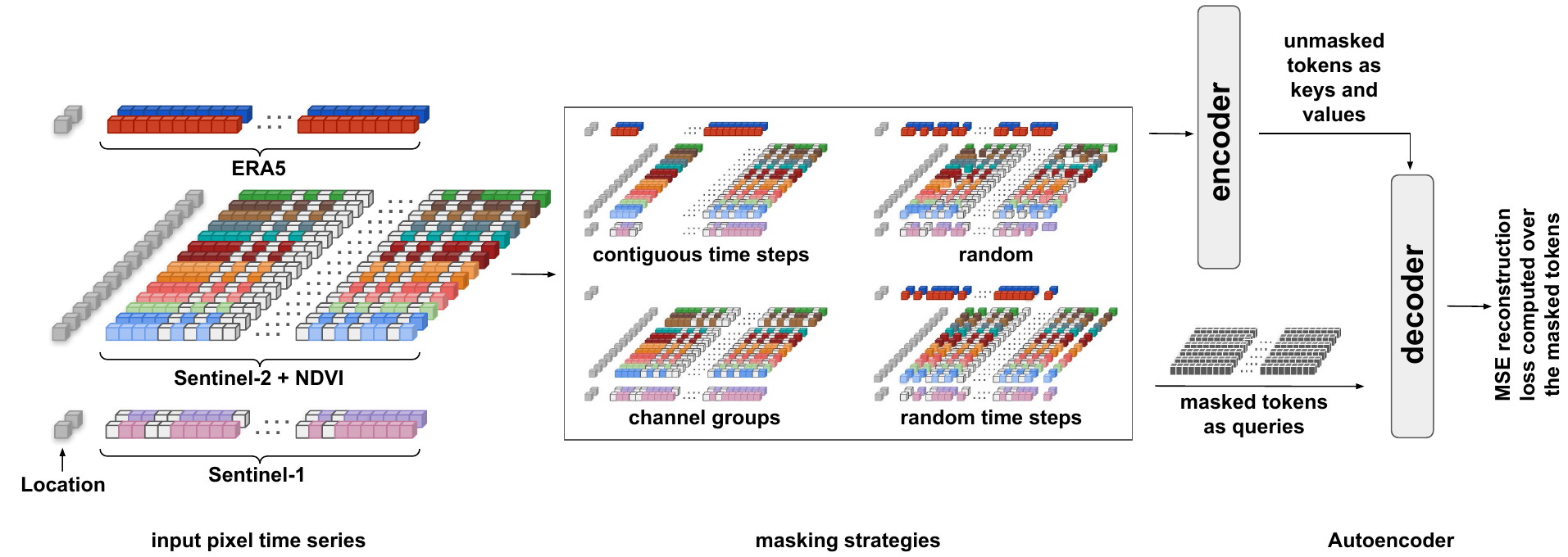}
    \caption{Visualization of \glsxtrshort{gl:crossprestoxts} using the full pre-training dataset described in \cref{sec:pixelsen12mscrts}.
    The static location variable is appended as an additional time step to the dynamic variables, as proposed by \citet{Tseng2023LightweightPT}.
    The unmasked tokens serve as inputs for the encoder before its output is used as keys and values to perform cross-attention in the decoder with the masked tokens as queries.
    Finally, the loss is computed over the reconstructed masked tokens only.
    The white cubes in the data input represent padded time steps.
    These tokens are also masked out in the encoder but not taken into account during loss computation.}
    \label{fig:crossprestoxts}
\end{figure*}

We propose the following variations to the original \gls{gl:presto} model to maximize temporal and spectral flexibility for downstream tasks and better align it with our pre-train dataset \PixelSenMSCRTS (\cf \cref{sec:pixelsen12mscrts}) and experimental setup (\cref{sec:experiments}).
In particular, we make the following modifications:
\begin{description}
    \item[Input encoding:] We no longer restrict the model to \num{12} monthly time steps, and thus remove the month encoding $p_\text{month}$ from the contextual embedding.
    Hence, the shared shared embedding dimension becomes $d_\text{emb} = d_\text{sin}+d_\text{channel}$, with $d_\text{sin} = \frac{3}{4} \cdot d_\text{emb}$ and $d_\text{channel} = \frac{1}{4} \cdot d_\text{emb}$.
    \item[Masking strategy:]
    Unlike \gls{gl:presto}, which uses random fallback masking to maintain a fixed ratio, we strictly adhere to the initial structured mask, even if fewer tokens are masked, with a masking ratio of \SI{70}{\percent}.
    \item[Selection of spectral bands:]
    To preserve the full spectral range of Sentinel-2, all bands (including bands \Bone, \Bnine, and \Bten) are retained throughout.
    If needed, individual bands are masked during fine-tuning rather than removed, improving model versatility without imposing predefined constraints.
\end{description}

We refer to this version of the model as \gls{gl:prestoxts}.

While additional spectral bands and higher temporal resolution enhance representational capacity, they also increase computational demands.
To mitigate this, we introduce \gls{gl:crossprestoxts}---a second, computationally more efficient variant---by integrating the cross-attention mechanism from \cref{sec:crossmae} into the \gls{gl:prestoxts} model.
The final architecture is illustrated in \cref{fig:crossprestoxts}.

\section{Experimental setup}
\label{sec:experiments}
All models are trained end-to-end, following a standard pre-training and fine-tuning paradigm.
In particular, all supervised methods are pre-trained on the \EuroCropsML pre-training dataset, before fine-tuning each model on the \EuroCropsML Estonia data.
On the other hand, all \gls{gl:ssl} algorithms are pre-trained on \PixelSenMSCRTS, before also fine-tuning them on the Estonia data.

In addition, we also consider training a randomly initialized Transformer encoder with a linear classifier from scratch without any pre-training on the same fine-tuning tasks as a baseline comparison.
This is referred to as \emph{no pre-training}.

Hyperparameters are optimized by computing the loss (during pre-training) or classification accuracy (during fine-tuning) on a held-out validation dataset.
\cref{sec:appendix_hyperparams} features a detailed outline of the hyperparameter tuning setup.

\subsection{Supervised pre-training }
\label{sec:exp_pretraining_supervised}
We use all training data for the transfer learning baseline to pre-train the Transformer model and validate it on the validation data.
From the pre-processed \EuroCropsML data, we keep all 13 Sentinel-2 spectral bands except band \Bten (cirrus SWIR), which is typically used for cloud detection. 
This band is not further necessary here since the data \EuroCropsML pre-processing pipeline \citep{Reuss25:EML} already performs cloud removal.
In order to identify the optimal setting for transfer learning, we consider batch size and cosine annealing as additional hyperparameters to the learning rate. 
The detailed configuration can be found in \cref{sec:appendix_pretraining}.

\begin{figure}
\centering
  \begin{subfigure}[t]{0.58\linewidth}
  \centering
    \includegraphics[width=\linewidth]{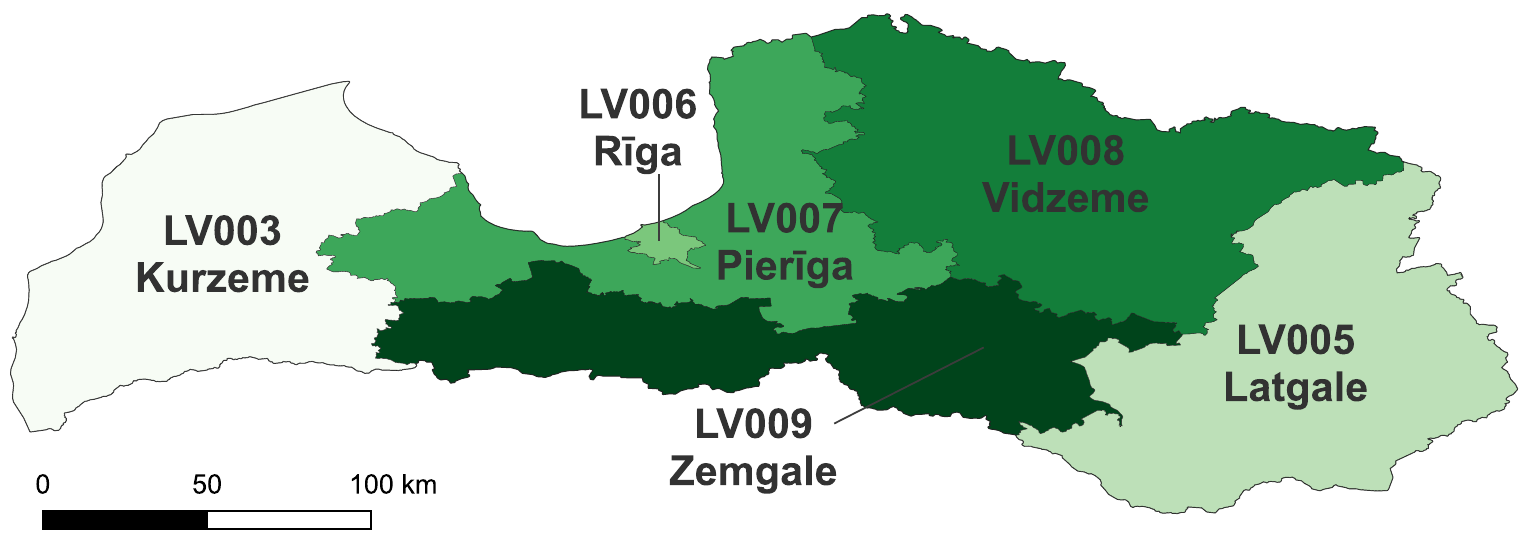}
    \label{fig:nuts3_lv}
  \end{subfigure}
  \centering
  \begin{subfigure}[t]{0.4\linewidth}
  \centering
    \includegraphics[width=1\linewidth]{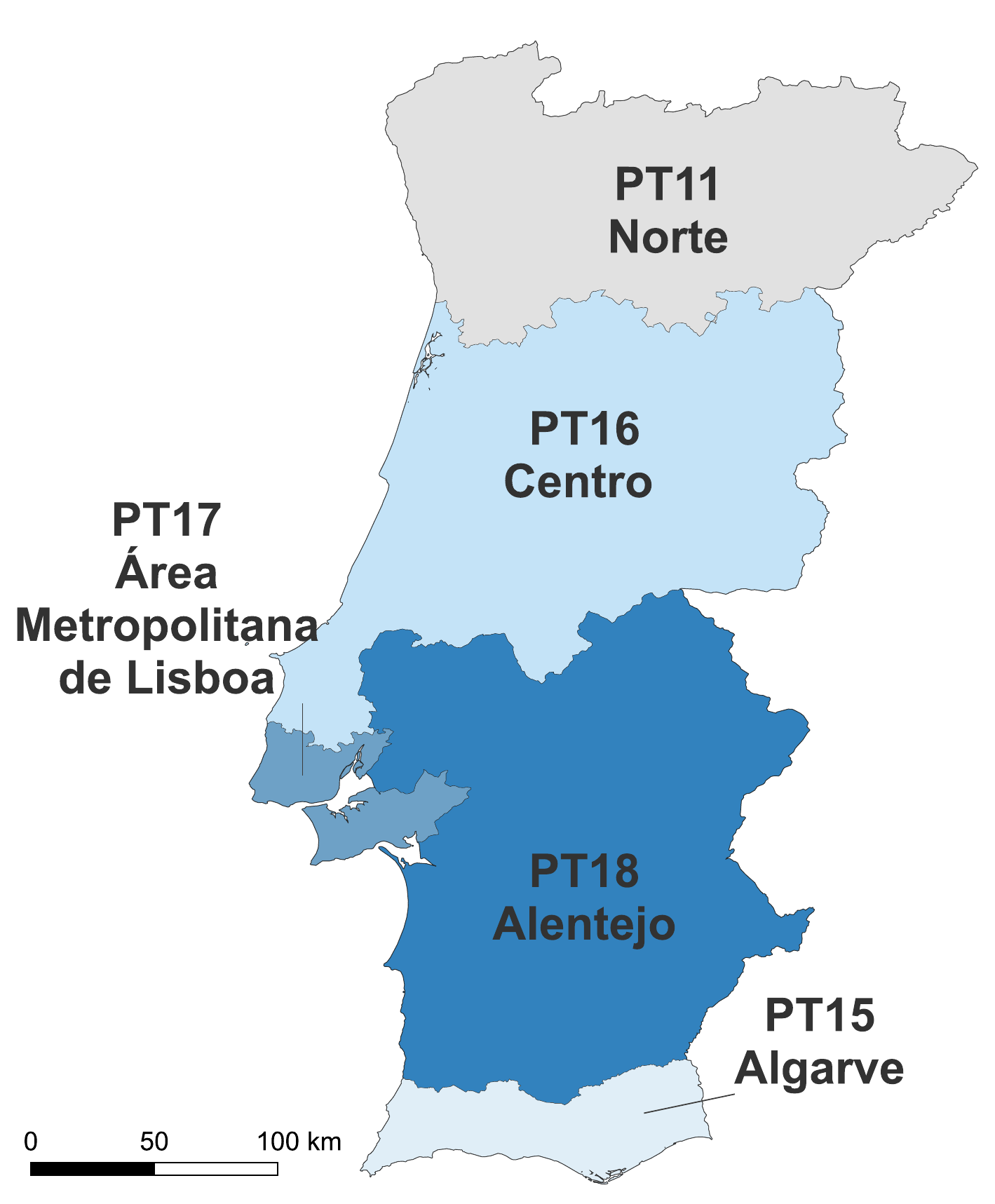}
    \label{fig:nuts2_pt}
  \end{subfigure}
  \caption{NUTS regions for Latvia (left) and Portugal (right) used for sampling meta-learning tasks.} 
  \label{fig:NUTS_regions}
\end{figure}

For meta-learning algorithms, we further sample meta-training and meta-validation tasks $\tau \in \mathcal{T}$ following the $n$-way $k$-shot paradigm for few-shot learning as follows:
\begin{description}
    \item[Step 1: Sampling a NUTS-region]
    With a probability given by the ratio of the number of data points per region to the total number of data points, we sample one of the regions
    \begin{itemize*}[label={}, itemjoin={{,\xspace}}, itemjoin*={{, and}}, after={,\xspace}]
        \item \region{LV003}%
        \item \region{LV005}%
        \item \region{LV006}%
        \item \region{LV007}%
        \item \region{LV008}%
        \item \region{LV009}%
        \item \region{PT11}%
        \item \region{PT15}%
        \item \region{PT16}%
        \item \region{PT17}%
        \item \region{PT18}%
    \end{itemize*}
    where the Portuguese regions (\portugal) are only considered for the second use case (\lvptee) mentioned above.
    For Latvia (\latvia) the regions correspond to the country's NUTS level-3 regions.
    For Portugal (\portugal) they reflect NUTS level-2 regions, as the Portuguese NUTS level-3 regions are relatively small and would hinder the creation of meaningful meta-learning tasks.
    A visualization of the NUTS regions can be seen in \cref{fig:NUTS_regions}.
    
    \item[Step 2: Sampling a task]
    We uniformly sample $n$ random distinct classes that are present in the previously selected region.
    For each class, we then uniformly sample $k_{\text{query}}$ random data points for the query set $\mathcal{D}_\text{query}$. 
    Similarly, from the remaining samples, for each of the $n$ classes, we uniformly select $k_{\text{support}}$ random data points for the support set $\mathcal{D}_\text{support}$ of the task.
    If after sampling the query set, there are only fewer than $k_\text{support}$ data points left for sampling from a specific class, then we include all of them in the support set.
\end{description}

Experiments are conducted for $n \in \{4, 10\}$ and $k \in \{1, 10\}$.
For all meta-learning algorithms and $n$-way $k$-shot settings, we consider the simplest algorithm variant with one gradient step (\ie $s=1$) in the inner optimization. 
For the setting $n=4$ and $k=1$, we also consider variants with multiple inner gradient steps $s\in\{1, 4, 10\}$.
A more detailed outline and analysis of the meta-learning settings can be found in \cref{sec:appendix_metalearning}.

During meta-learning, we sample a fixed set of \num{100} meta-validation tasks for model validation and hyperparameter tuning, \cf \cref{sec:appendix_hyperparams}, from the pre-training validation subset following steps 1 and 2 so that model validation is always performed on the same set of tasks.

In order to assess the ability of different transfer and meta-learning algorithms to cope with data-scarce situations, we fine-tune the models in various few-shot settings with 
\numlist[list-final-separator = {, or }]{1;5;10;20;100;200;500} shots
\citep[\cf][]{Reuss25:EML}.

\subsubsection{Self-supervised pre-training}
\label{sec:exp_pretraining_ssl}
For the baseline \gls{gl:presto} model, we obtained the original publicly available model parameters pre-trained by \citet{Tseng2023LightweightPT}.
\Gls{gl:prestoxts} and \gls{gl:crossprestoxts} were pre-trained from scratch on all channels of the \PixelSenMSCRTS using \gls{gl:mse} loss over the masked tokens only.

\subsection{Fine-tuning setup}
\label{sec:exp_finetuning}
Classification performance during fine-tuning is evaluated using \emph{overall classification accuracy} $a_\text{OA}$ (micro-averaged across instances from all classes).
To assess minority-class performance, we also report the accuracy on the fine-tuning sets excluding the \meadow majority class ($a_\text{MCA}$).
Additional analyses, including hyperparameter choices and evaluations with \emph{Cohen's kappa coefficient} $\kappa$ metric, are deferred to \cref{sec:appendix_additionalresults}.

All models are fine-tuned for up to \num{200} epochs with a batch size of \num{16}.
We use the \num{1000} fixed data points sampled from the validation subset for model validation and hyperparameter tuning.
Early stopping is triggered if the validation loss does not improve for five epochs.
The final models are evaluated on all \num{35182} test samples.
We evaluate three learning rate strategies:
\begin{enumerate}[itemsep=0pt, topsep=0pt, partopsep=0pt, parsep=0pt]
    \item The same learning rate applies to the backbone model and the randomly initialized classification head.
    \item Separate, independently optimized learning rates are used for the backbone and the classification head.
    \item Only the classification head is fine-tuned with a tuned learning rate, while the backbone learning rate is fixed to zero, effectively freezing the backbone.
\end{enumerate}

In order to evaluate the robustness of the results and reproducibility of the algorithms' performances, we conduct all experiments five times, repeated with different random seeds (influencing, \eg random parameter initialization and mini-batch sampling, \cf \cref{sec:appendix_finetuning}).

\begin{figure}[t]
\centering
\includegraphics[width=.95\linewidth]{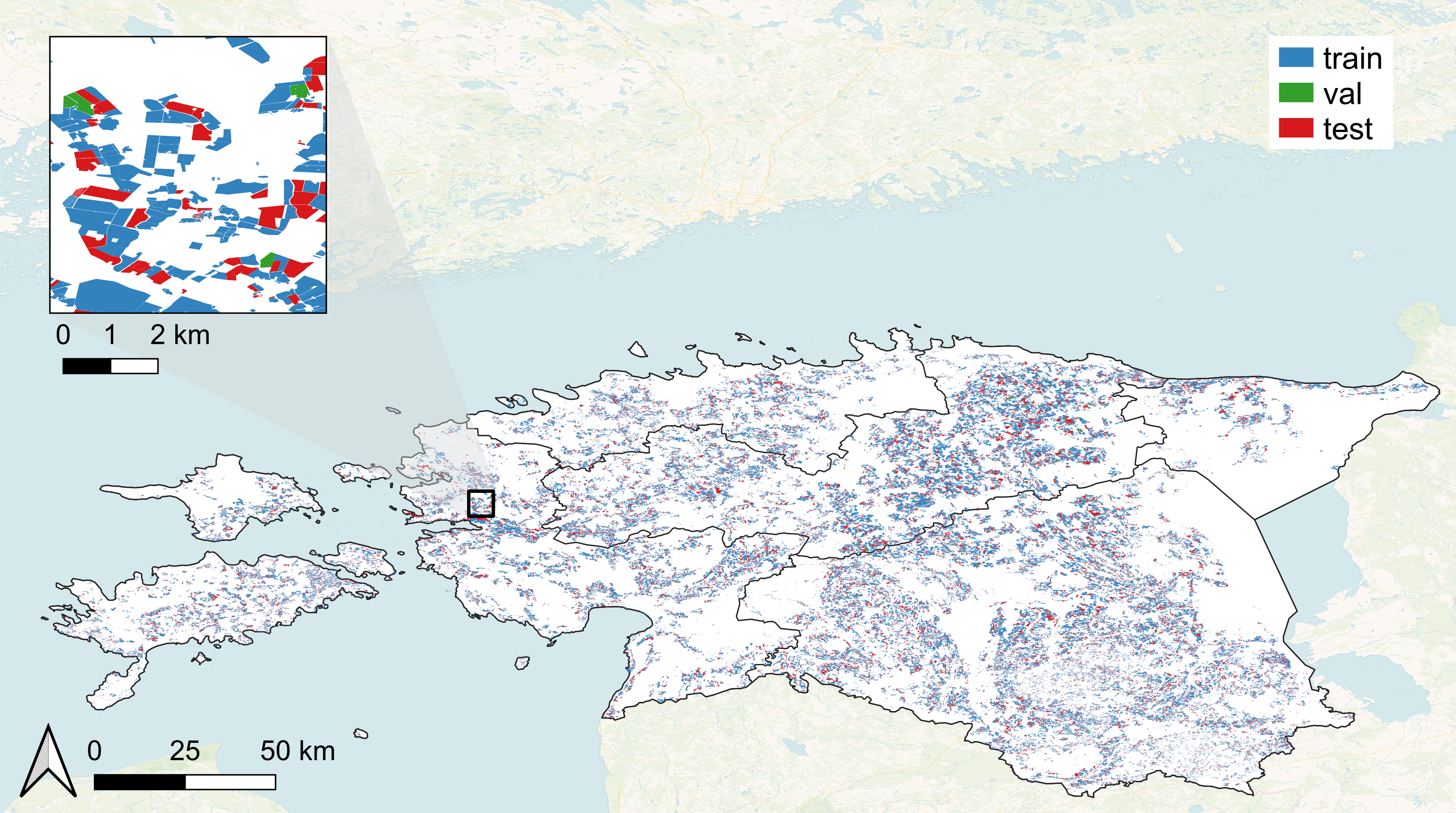}
\caption{Spatial distribution of the fine-tuning dataset.
The entire Estonia data is randomly split into train (\qty{60}{\percent}), validation (\qty{20}{\percent}), and test (\qty{20}{\percent}). 
For the final validation set, only \num{1000} samples are used.
As can be seen in the map, the distribution of the data is spatially uniform throughout the country.}
\label{fig:estonia_spatialdistr}
\end{figure}

In addition, we train the baseline on all of the training data from Estonia to quantify the complexity of the downstream task in a non-few-shot setting.
\Cref{fig:estonia_spatialdistr} visualizes the spatial distribution of the fine-tuning dataset, including all training samples.

In order to evaluate model performance, we define two fine-tuning scenarios with different input modalities:
\begin{description}
    \item[\taskStwo:] The fine-tuning data exclusively contains Sentinel-2 imagery (and \gls{gl:ndvi} in the case of \gls{gl:ssl}).
    For \gls{gl:ssl}, this reflects only a subset of the data available during pre-training.
    \item[\taskalldata:] The fine-tuning data contains additional data modalities, such as Sentinel-1 and \gls{gl:era} data.
    This scenario is only evaluated for \gls{gl:ssl} and demonstrates benchmarking on an extended version of the \EuroCropsML dataset.
    For \gls{gl:prestoxts} and \gls{gl:crossprestoxts}, this reflects a setup, where all input modalities from pre-training (\cref{sec:pixelsen12mscrts}) are present during fine-tuning.
\end{description}

Again, in terms of spectral resolution, in general, we keep all \num{13} Sentinel-2 bands except band \Bten. 
Only the already pre-trained \gls{gl:presto} is fine-tuned on an even smaller subset of bands, by further removing bands \Bone and \Bnine.

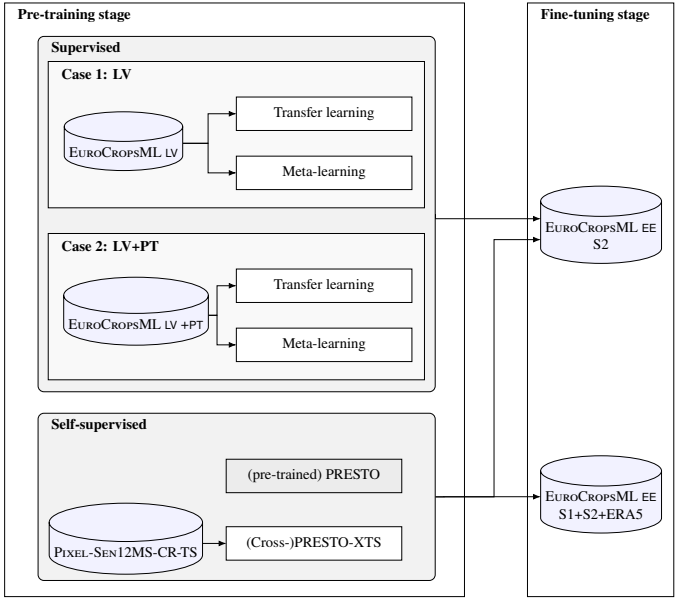
\begin{figure}[t]
    \centering
    \resizebox{\linewidth}{!}{\begin{tikzpicture}[
    node distance=1cm and 1cm,
    box/.style={draw, rectangle, minimum width=4.2cm, minimum height=0.8cm, align=center, fill=white},
    sectionbox/.style={draw, rectangle, rounded corners, inner sep=0.5cm, fill=gray!10},
    subbox/.style={draw, rectangle, minimum height=3.5cm, inner sep=0.5em, fill=gray!5},
    dataset/.style={draw, cylinder, shape border rotate=90, aspect=0.25, minimum height=1.2cm, minimum width=2.8cm, align=center, fill=blue!5},
    ->, >=Latex
]

\node[draw, rectangle, minimum width=11cm, minimum height=14.2cm, anchor=north west] (pretrainbox) at (0,0) {};
\node[draw, rectangle, minimum width=3.5cm, minimum height=14.2cm, anchor=north west] (finetunebox) at ([xshift=7cm]pretrainbox.north) {};

\node[anchor=west] at ([xshift=0.2cm,yshift=-0.3cm]pretrainbox.north west) {\textbf{Pre-training stage}};
\node[anchor=west] at ([xshift=0.2cm,yshift=-0.3cm]finetunebox.north west) {\textbf{Fine-tuning stage}};

\node[sectionbox, minimum width=9.5cm, minimum height=8.5cm, anchor=north west] (supbox) at ([xshift=0.8cm,yshift=-0.8cm]pretrainbox.north west) {};
\node[anchor=west] at ([xshift=0.2cm,yshift=-0.3cm]supbox.north west) {\textbf{Supervised}};

\node[subbox, minimum width=9cm] (supcase1) at ($(supbox.north west)!0.5!(supbox.south west)+(4.75,1.9)$) {};
\node[anchor=west] at ([xshift=0.2cm,yshift=-0.3cm]supcase1.north west) {\textbf{Case 1: LV}};
\node[box, right=4.5cm of supcase1.west, yshift=0.5cm] (transfer1) {Transfer learning};
\node[box, below=0.6cm of transfer1] (meta1) {Meta-learning};
\node[dataset] (data_lv) at ($(transfer1)!0.5!(meta1) + (-4.8, -0.2)$) {\EuroCropsML\ \latvia};
\draw[->] ([yshift=0.2cm]data_lv.east) -- ++(0.6,0) |- (transfer1.west);
\draw[->] ([yshift=0.2cm]data_lv.east) -- ++(0.6,0) |- (meta1.west);

\node[subbox, below=0.6cm of supcase1.south west, anchor=north west, minimum width=9cm] (supcase2) {};
\node[anchor=west] at ([xshift=0.2cm,yshift=-0.3cm]supcase2.north west) {\textbf{Case 2: LV+PT}};
\node[box, right=4.5cm of supcase2.west, yshift=0.5cm] (transfer2) {Transfer learning};
\node[box, below=0.6cm of transfer2] (meta2) {Meta-learning};
\node[dataset] (data_lvpt) at ($(transfer2)!0.5!(meta2) + (-4.5, -0.2)$) {\EuroCropsML\ \latvia+\portugal};
\draw[->] ([yshift=0.2cm]data_lvpt.east) -- ++(0.2,0) |- (transfer2.west);
\draw[->] ([yshift=0.2cm]data_lvpt.east) -- ++(0.2,0) |- (meta2.west);

\node[sectionbox, minimum width=9.5cm, minimum height=4cm, anchor=north west] (sslbox) at ([yshift=-0.5cm]supbox.south west) {};
\node[anchor=west] at ([xshift=0.2cm,yshift=-0.3cm]sslbox.north west) {\textbf{Self-supervised}};
\node[box, fill=gray!15, right=4.5cm of sslbox.west, yshift=0.5cm] (presto) {(pre-trained) PRESTO};
\node[box, below=0.8cm of presto] (prestoxts) {(Cross-)PRESTO-XTS};
\node[dataset] (data_ssl) at ($(prestoxts) + (-4.5, -0.2)$) { \PixelSenMSCRTS};
\draw[->] ([yshift=0.2cm]data_ssl.east) -- (prestoxts.west);

\node[dataset, anchor=north west] (finetune_s2) at ([xshift=0.7cm,yshift=-4.5cm]finetunebox.north west) {\EuroCropsML\ \estonia\\ S2};
\node[dataset, anchor=north west] (finetune_s1s2era5) at ([xshift=0.7cm,yshift=-10.95cm]finetunebox.north west) {\EuroCropsML\ \estonia\\ S1+S2+ERA5};

\draw[->] (supbox.east) -- ++(0,-0.1) |- ([yshift=0.4cm]finetune_s2.west);
\draw[->] (sslbox.east) -- ++(1.4,0) |- ([yshift=-0.1cm]finetune_s2.west);
\draw[->] (sslbox.east) -- ++(0.2,0) |- ([yshift=0.2cm]finetune_s1s2era5.west);

\end{tikzpicture}}
    \hspace{.03\linewidth}
    \caption{Overview of the pre-training and fine-tuning setups.
    The supervised algorithms are either pre-trained first on \EuroCropsML \latvia or \EuroCropsML\ \latvia + \portugal, before being fine-tuned on \EuroCropsML\ \estonia S2.
    While \gls{gl:prestoxts} and \gls{gl:crossprestoxts} are pre-trained on \PixelSenMSCRTS via reconstruction, for the original \gls{gl:presto}, we use the pre-trained weights \citep{Tseng2023LightweightPT}.
    Hence, no additional pre-training is conducted with the \gls{gl:presto} model.
    All \gls{gl:ssl} models are subsequently fine-tuned on either \EuroCropsML\ \estonia S2 or \EuroCropsML\ \estonia S1+S2+ERA5.}
    \label{fig:training_pipeline}
\end{figure}

\Cref{fig:training_pipeline} gives an overview over the full training processes.

\section{Results}
\label{sec:results}
We present the best results for each algorithm and fine-tuning task in this section and defer a detailed analysis of algorithm variations and hyperparameter influences to \ref{sec:appendix_hyperparams} and \ref{sec:appendix_additionalresults}.
The tables in this section compare the randomly initialized baseline model (no pre-training) to transfer learning, the meta-learning algorithms \gls{gl:anil}, \gls{gl:maml}, \gls{gl:fomaml}, and \gls{gl:timl} (with and without encoder), and variants of the \gls{gl:ssl} model \gls{gl:presto}.
In each case, we report the mean and standard deviation of $a_\text{OA}$ estimated from the repeated experiments with varying random seeds, as detailed in \cref{sec:exp_finetuning,sec:appendix_finetuning}.

\subsection{Results on \taskStwo}
\label{sec:results_taskstwo}
When training the randomly initialized baseline model on the full training data of the \taskStwo\ fine-tuning task, we receive a final accuracy of $a_\text{OA}=\num{0.787(0.004)}$ on the test set.
This can serve as a general reference value for the overall difficulty of the learning task on the Estonian data.

\begin{table*}[ht!]
    \caption{%
    Results for the \taskStwo\ benchmark.
    We report \emph{classification accuracy} on the test set for each algorithm and few-shot task.
    Test metrics are shown as mean $\pm$ standard deviation over five runs, \cf \cref{sec:exp_finetuning}. 
    The best result per few-shot scenario is marked in {\color{blue}\textbf{blue bold}}.
    \textbf{Black bold} values highlight the best result for supervised or \gls{gl:ssl} pre-training.
    Additionally, cases in which the \lvptee was an improvement over the corresponding \lvee result are highlighted in \textcolor{mediumseagreen}{green color}.
    }
    \label{tab:comparison_results_eurocropsml_S2}

    \begin{subtable}{\linewidth}
        \caption{Fine-tuning classification results for \taskStwo.
        \label{tab:best_results_S2}}
        \centering\scriptsize

\robustify\bfseries
\sisetup{detect-all=true,uncertainty-mode=separate,table-align-uncertainty=true,round-mode=uncertainty,round-precision=2}

\begin{tabular}{llSSSSSSS}
\toprule
{} & {benchmark task ($k$-shot)} & {1} & {5} & {10} & {20} & {100} & {200} & {500} \\
{} & {algorithm} & {} & {} & {} & {} & {} & {} & {} \\

\cmidrule(r){1-2} \cmidrule(lr){3-3} \cmidrule(lr){4-4} \cmidrule(lr){5-5} \cmidrule(lr){6-6} \cmidrule(lr){7-7} \cmidrule(lr){8-8} \cmidrule(l){9-9}

\multirow[c]{17}{*}{\rotatebox[origin=c]{90}{accuracy}}
 & no pre-training & \num{0.170667 +- 0.1403229} & \bfseries \color{blue} \num{0.331130 +- 0.107528} & \bfseries \color{blue} \num{0.438207 +- 0.027004} & \num{0.490370 +- 0.019933} & \num{0.518043 +- 0.033152} & \num{0.572810 +- 0.018726} & \num{0.631692 +- 0.006193} \\
\cmidrule(r){2-9}
 & transfer learning (\latvia) & \num{0.143215 +- 0.069705} & \num{0.203286 +- 0.081175} & \num{0.253817 +- 0.060033} & \num{0.367171 +- 0.086884} & \num{0.464601 +- 0.043019} & \num{0.502945 +- 0.032297} & \num{0.574510 +- 0.047607} \\
 & \gls{gl:anil} (\latvia) & \num{0.189199 +- 0.047364} & \num{0.250378 +- 0.075744} & \bfseries \num{0.419027 +- 0.045249} & \num{0.490296 +- 0.068658} & \bfseries \color{blue} \num{0.561480 +- 0.033554} & \num{0.589193 +- 0.019552} & \bfseries \color{blue} \num{0.652282 +- 0.011135} \\
 & \gls{gl:maml} (\latvia) & \num{0.176300 +- 0.077477} & \bfseries \num{0.259235 +- 0.088080} & \num{0.390080 +- 0.034725} & \bfseries \color{blue} \num{0.506549 +- 0.056659} & \num{0.552163 +- 0.022856} & \bfseries \color{blue} \num{0.611131 +- 0.042565} & \num{0.647877 +- 0.016656} \\
 & \gls{gl:fomaml} (\latvia) & \num{0.206378 +- 0.110366} & \num{0.228213 +- 0.051797} & \num{0.408169 +- 0.064589} & \num{0.452919 +- 0.080190} & \num{0.515650 +- 0.035827} & \num{0.556864 +- 0.017883} & \num{0.619050 +- 0.006883} \\
 & \gls{gl:timl} (encoder) (\latvia) & \bfseries \color{blue} \num{0.247098 +- 0.091937} & \num{0.247007 +- 0.105412} & \num{0.368365 +- 0.040482} & \num{0.423427 +- 0.067324} & \num{0.493593 +- 0.045185} & \num{0.517202 +- 0.021845} & \num{0.591223 +- 0.021084} \\
 & \gls{gl:timl} (no encoder) (\latvia) & \num{0.176289 +- 0.122604} & \num{0.263817 +- 0.100001} & \num{0.336701 +- 0.068644} & \num{0.456290 +- 0.058360} & \num{0.530152 +- 0.044469} & \num{0.535490 +- 0.018198} & \num{0.637354 +- 0.010053} \\
 \arrayrulecolor{gray} 
 \cmidrule(r){2-9}
 \arrayrulecolor{black}
 & transfer learning (\latvia + \portugal) & \num{0.078728 +- 0.009102} & \num{0.123285 +- 0.005764} & \num{0.249059 +- 0.097862} & \num{0.292206 +- 0.089407} & \num{0.405184 +- 0.022854} & \num{0.473322 +- 0.033044} & \color{mediumseagreen} \num{0.583457 +- 0.027695} \\
    & \gls{gl:anil} (\latvia + \portugal) & \num{0.153868 +- 0.050276} & \num{0.200495 +- 0.019973} & \num{0.321346 +- 0.089598} & \num{0.479802 +- 0.044880} & \num{0.538497 +- 0.027642} & \num{0.576050 +- 0.009840} & \num{0.623080 +- 0.019012} \\
    & \gls{gl:maml} (\latvia + \portugal) & \num{0.116651 +- 0.013054} & \num{0.183372 +- 0.018023} & \num{0.365721 +- 0.085958} & \num{0.452282 +- 0.093832} & \num{0.516611 +- 0.016206} & \num{0.567955 +- 0.017392} & \num{0.636922 +- 0.014028} \\
    & \gls{gl:fomaml} (\latvia + \portugal) & \num{0.103286 +- 0.010742} & \num{0.192144 +- 0.034912} & \num{0.333284 +- 0.101718} & \num{0.445086 +- 0.058282} & \num{0.511597 +- 0.014298} & \color{mediumseagreen} \num{0.562089 +- 0.027844} & \color{mediumseagreen} \num{0.623393 +- 0.011230} \\
    & \gls{gl:timl} (encoder) (\latvia + \portugal) & \num{0.128014 +- 0.031495} & \num{0.163311 +- 0.030520} & \num{0.344199 +- 0.084231} & \color{mediumseagreen} \num{0.438446 +- 0.030566} & \num{0.488017 +- 0.029107} & \num{0.472890 +- 0.038095} & \num{0.586726 +- 0.024490} \\
    & \gls{gl:timl} (no encoder) (\latvia + \portugal) & \num{0.162282 +- 0.073580} & \num{0.188551 +- 0.027405} & \num{0.299284 +- 0.018234} & \num{0.441015 +- 0.069749} & \num{0.516349 +- 0.062820} & \color{mediumseagreen} \num{0.537843 +- 0.020560} & \num{0.622421 +- 0.011621} \\
 \cmidrule(r){2-9}
 & \gls{gl:presto} & \bfseries \num{0.156603 +- 0.077573} & \num{0.236115 +- 0.119967} & \num{0.235501 +- 0.029735} & \num{0.322085 +- 0.063678} & \num{0.457927 +- 0.045003} & \bfseries \num{0.490586 +- 0.054895} & \num{0.552788 +- 0.003684} \\
 & \gls{gl:prestoxts} & \num{0.049662 +- 0.024817} & \num{0.182468 +- 0.147091} & \num{0.275885 +- 0.094361} & \num{0.442584 +- 0.023787} & \bfseries \num{0.501433 +- 0.035497} & \num{0.486368 +- 0.016917} & \bfseries \num{0.621903 +- 0.021139} \\
 & \gls{gl:crossprestoxts} & \num{0.094077 +- 0.102261} & \bfseries \num{0.320158 +- 0.181785} & \bfseries \num{0.299864 +- 0.091585} & \bfseries \num{0.474390 +- 0.011035} & \num{0.495134 +- 0.017532} & \num{0.467773 +- 0.022158} & \num{0.598533 +- 0.017522} \\
\bottomrule
\end{tabular}

    \end{subtable}%

    \medskip
    
   \begin{subtable}{\linewidth}
        \caption{Fine-tuning classification results for \taskStwo\ excluding the majority class \meadow.
        \label{tab:best_results_S2_nomeadow}}
        \centering\scriptsize

\robustify\bfseries
\sisetup{detect-all=true,uncertainty-mode=separate,table-align-uncertainty=true,round-mode=uncertainty,round-precision=2}
\begin{tabular}{llSSSSSSS}
\toprule
{} & {benchmark task ($k$-shot)} & {1} & {5} & {10} & {20} & {100} & {200} & {500} \\
{} & {algorithm} & {} & {} & {} & {} & {} & {} & {} \\

\cmidrule(r){1-2} \cmidrule(lr){3-3} \cmidrule(lr){4-4} \cmidrule(lr){5-5} \cmidrule(lr){6-6} \cmidrule(lr){7-7} \cmidrule(lr){8-8} \cmidrule(l){9-9}

\multirow[c]{17}{*}{\rotatebox[origin=c]{90}{accuracy}}
 & no pre-training & \num{0.049561 +- 0.044801} & \num{0.036514 +- 0.038025} & \num{0.060069 +- 0.045342} & \num{0.141643 +- 0.097123} & \num{0.274424 +- 0.114192} & \num{0.455055 +- 0.027361} & \num{0.516660 +- 0.022612} \\
 \cmidrule(r){2-9}
 & transfer learning (\latvia) & \num{0.154853 +- 0.107352} & \num{0.195727 +- 0.128969} & \num{0.236427 +- 0.171956} & \num{0.295199 +- 0.094578} & \num{0.388751 +- 0.097871} & \num{0.430508 +- 0.072361} & \num{0.530950 +- 0.058361} \\
 & \gls{gl:anil} (\latvia) & \num{0.102981 +- 0.132144} & \num{0.197733 +- 0.112034} & \num{0.178429 +- 0.118757} & \bfseries \color{blue} \num{0.308213 +- 0.051649} & \num{0.355256 +- 0.087896} & \num{0.451185 +- 0.038663} & \bfseries \color{blue} \num{0.547812 +- 0.009635} \\
 & \gls{gl:maml} (\latvia) & \num{0.143724 +- 0.106018} & \num{0.065061 +- 0.086016} & \bfseries \color{blue} \num{0.262652 +- 0.045418} & \num{0.281857 +- 0.059751} & \num{0.299504 +- 0.055557} & \num{0.385024 +- 0.069862} & \num{0.480048 +- 0.083422} \\
 & \gls{gl:fomaml} (\latvia) & \num{0.078119 +- 0.106641} & \num{0.087798 +- 0.099027} & \num{0.073813 +- 0.080087} & \num{0.169971 +- 0.081076} & \num{0.324116 +- 0.093051} & \num{0.440144 +- 0.026306} & \num{0.486163 +- 0.023181} \\
 & \gls{gl:timl} (encoder) (\latvia) & \num{0.042062 +- 0.031523} & \num{0.116530 +- 0.050631} & \num{0.067491 +- 0.058294} & \num{0.152880 +- 0.036337} & \num{0.299341 +- 0.054728} & \num{0.289345 +- 0.069738} & \num{0.449888 +- 0.030791} \\
 & \gls{gl:timl} (no encoder) (\latvia) & \num{0.108834 +- 0.103156} & \num{0.208556 +- 0.092152} & \num{0.205613 +- 0.085402} & \num{0.261333 +- 0.090260} & \num{0.377100 +- 0.053216} & \num{0.420590 +- 0.061563} & \num{0.546177 +- 0.018731} \\
 \arrayrulecolor{gray} 
 \cmidrule(r){2-9}
 \arrayrulecolor{black}
  & transfer learning (\latvia + \portugal) & \num{0.144433 +- 0.018332} & \color{mediumseagreen} \num{0.226792 +- 0.020206} & \num{0.202343 +- 0.050647} & \num{0.255818 +- 0.041097} & \num{0.328203 +- 0.095213} & \color{mediumseagreen} \num{0.438596 +- 0.037101} & \num{0.488942 +- 0.041675} \\
 & ANIL (\latvia + \portugal) & \color{mediumseagreen} \num{0.165252 +- 0.082997} & \bfseries \color{blue} \num{0.271513 +- 0.031508} & \color{mediumseagreen} \num{0.223369 +- 0.086776} & \num{0.247719 +- 0.063074} & \num{0.325271 +- 0.067134} & \num{0.435402 +- 0.071152} & \num{0.501488 +- 0.040766} \\
 & MAML (\latvia + \portugal) & \bfseries \color{blue} \num{0.190637 +- 0.055759} & \num{0.229800 +- 0.017797} & \num{0.175345 +- 0.087016} & \num{0.262990 +- 0.033488} & \color{mediumseagreen} \num{0.320791 +- 0.078009} & \color{mediumseagreen} \num{0.442978 +- 0.023009} & \color{mediumseagreen} \num{0.527135 +- 0.016512} \\
 & FOMAML (\latvia + \portugal) & \color{mediumseagreen} \num{0.158919 +- 0.083956} & \color{mediumseagreen} \num{0.189656 +- 0.105281} & \color{mediumseagreen} \num{0.139517 +- 0.095506} & \color{mediumseagreen} \num{0.283427 +- 0.075700} & \color{mediumseagreen} \num{0.335604 +- 0.074263} & \bfseries \color{blue} \num{0.458292 +- 0.030590} & \color{mediumseagreen} \num{0.512922 +- 0.030446} \\
 & TIML (encoder) (\latvia + \portugal) & \color{mediumseagreen} \num{0.106742 +- 0.082976} & \color{mediumseagreen} \num{0.117521 +- 0.077575} & \color{mediumseagreen} \num{0.111341 +- 0.082230} & \num{0.120846 +- 0.068977} & \num{0.244427 +- 0.060754} & \color{mediumseagreen} \num{0.299123 +- 0.068688} & \num{0.404186 +- 0.071995} \\
 & TIML (no encoder) (\latvia + \portugal) & \color{mediumseagreen} \num{0.181536 +- 0.103951} & \num{0.199891 +- 0.070253} & \color{mediumseagreen} \num{0.230770 +- 0.046004} & \color{mediumseagreen} \num{0.281182 +- 0.035545} & \bfseries \color{blue} \num{0.398398 +- 0.042862} & \color{mediumseagreen} \num{0.451207 +- 0.052176} & \num{0.539910 +- 0.024869} \\
\cmidrule(r){2-9}
 & \gls{gl:presto} & \num{0.027685 +- 0.029748} & \num{0.064156 +- 0.046668} & \num{0.083939 +- 0.076904} & \num{0.098806 +- 0.056404} & \num{0.266772 +- 0.057758} & \num{0.323189 +- 0.013983} & \num{0.386670 +- 0.037515} \\
 & \gls{gl:prestoxts} & \bfseries \num{0.095220 +- 0.047584} & \bfseries \num{0.074522 +- 0.043841} & \num{0.031021 +- 0.032720} & \num{0.013167 +- 0.010200} & \bfseries \num{0.277977 +- 0.165034} & \bfseries \num{0.396000 +- 0.055134} & \bfseries \num{0.510164 +- 0.028063} \\
 & \gls{gl:crossprestoxts} & \num{0.070914 +- 0.046183} & \num{0.064679 +- 0.086660} & \bfseries \num{0.094316 +- 0.071616} & \bfseries \num{0.118797 +- 0.020113} & \num{0.255295 +- 0.084629} & \num{0.206638 +- 0.160447} & \num{0.498033 +- 0.044230} \\
\bottomrule
\end{tabular}

  \end{subtable}%

    \medskip

  \begin{subtable}{\linewidth}
        \caption{Fine-tuning classification results for \taskStwo\ on the \meadow class.
        \label{tab:best_results_S2_meadow}}
        \centering\scriptsize

\robustify\bfseries
\sisetup{detect-all=true,uncertainty-mode=separate,table-align-uncertainty=true,round-mode=uncertainty,round-precision=2}
\begin{tabular}{llSSSSSSS}
\toprule
{} & {benchmark task ($k$-shot)} & {1} & {5} & {10} & {20} & {100} & {200} & {500} \\
{} & {algorithm} & {} & {} & {} & {} & {} & {} & {} \\

\cmidrule(r){1-2} \cmidrule(lr){3-3} \cmidrule(lr){4-4} \cmidrule(lr){5-5} \cmidrule(lr){6-6} \cmidrule(lr){7-7} \cmidrule(lr){8-8} \cmidrule(l){9-9}

\multirow[c]{17}{*}{\rotatebox[origin=c]{90}{accuracy}}
 & no pre-training & \num{0.302679 +- 0.323045} & \bfseries \color{blue} \num{0.652278 +- 0.262302} & \bfseries \color{blue} \num{0.850401 +- 0.059134} & \num{0.870504 +- 0.079983} & \num{0.783604 +- 0.099509} & \num{0.701170 +- 0.034364} & \num{0.757084 +- 0.035518} \\
  \cmidrule(r){2-9}
 & transfer learning (\latvia) & \num{0.131123 +- 0.218191} & \num{0.211525 +- 0.232999} & \num{0.272774 +- 0.187787} & \num{0.442262 +- 0.145285} & \num{0.540320 +- 0.088008} & \num{0.581905 +- 0.075460} & \num{0.621993 +- 0.047007} \\
 & ANIL (\latvia) & \num{0.283182 +- 0.211827} & \num{0.307765 +- 0.275517} & \num{0.681293 +- 0.207355} & \num{0.688778 +- 0.189508} & \num{0.786277 +- 0.130411} & \num{0.739630 +- 0.030068} & \num{0.766162 +- 0.026314} \\
 & MAML (\latvia) & \num{0.211810 +- 0.267501} & \bfseries \num{0.470896 +- 0.238401} & \num{0.528985 +- 0.099418} & \num{0.751476 +- 0.120563} & \bfseries \color{blue} \num{0.827577 +- 0.061712} & \bfseries \color{blue} \num{0.857601 +- 0.033305} & \bfseries \color{blue} \num{0.830820 +- 0.069141} \\
 & FOMAML (\latvia) & \num{0.346189 +- 0.326887} & \num{0.381275 +- 0.204355} & \bfseries \num{0.785612 +- 0.115493} & \bfseries \num{0.810123 +- 0.100549} & \num{0.724434 +- 0.119284} & \num{0.684097 +- 0.044850} & \num{0.763904 +- 0.031612} \\
 & TIML (encoder) (\latvia) & \bfseries \color{blue} \num{0.470599 +- 0.169052} & \num{0.389235 +- 0.265594} & \num{0.696335 +- 0.099691} & \num{0.718339 +- 0.173808} & \num{0.705341 +- 0.133379} & \num{0.765580 +- 0.058895} & \num{0.745286 +- 0.051590} \\
 & TIML (no encoder) (\latvia) & \num{0.249379 +- 0.331914} & \num{0.324054 +- 0.282702} & \num{0.479594 +- 0.226895} & \num{0.670421 +- 0.172804} & \num{0.696988 +- 0.140300} & \num{0.660738 +- 0.083692} & \num{0.736743 +- 0.038171} \\
 \arrayrulecolor{gray} 
 \cmidrule(r){2-9}
 \arrayrulecolor{black}
  & transfer learning (\latvia + \portugal) & \num{0.007105 +- 0.014638} & \num{0.010456 +- 0.010526} & \color{mediumseagreen} \num{0.299982 +- 0.234083} & \num{0.331872 +- 0.191900} & \num{0.489099 +- 0.078321} & \num{0.511174 +- 0.063559} & \color{mediumseagreen} \num{0.686485 +- 0.067737} \\
 & ANIL (\latvia + \portugal) & \num{0.141460 +- 0.191027} & \num{0.123436 +- 0.073447} & \num{0.428147 +- 0.202489} & \color{mediumseagreen} \num{0.732787 +- 0.122771} & \num{0.770926 +- 0.099699} & \num{0.729365 +- 0.086079} & \num{0.755623 +- 0.028642} \\
 & MAML (\latvia + \portugal) & \num{0.036001 +- 0.050311} & \num{0.132763 +- 0.025804} & \color{mediumseagreen} \num{0.573243 +- 0.205711} & \num{0.658623 +- 0.196983} & \num{0.730066 +- 0.087304} & \num{0.704188 +- 0.020506} & \num{0.756597 +- 0.025376} \\
 & FOMAML (\latvia + \portugal) & \num{0.042642 +- 0.078258} & \num{0.194855 +- 0.174007} & \num{0.544502 +- 0.281093} & \num{0.621303 +- 0.170445} & \num{0.703440 +- 0.096628} & \num{0.675233 +- 0.068681} & \num{0.743813 +- 0.042933} \\
 & TIML (encoder) (\latvia + \portugal) & \num{0.151203 +- 0.139258} & \num{0.213224 +- 0.138252} & \num{0.598028 +- 0.252360} & \color{mediumseagreen} \num{0.784649 +- 0.114895} & \color{mediumseagreen} \num{0.753544 +- 0.119688} & \num{0.662306 +- 0.151356} & \color{mediumseagreen} \num{0.785707 +- 0.044031} \\
 & TIML (no encoder) (\latvia + \portugal) & \num{0.141294 +- 0.264010} & \num{0.176190 +- 0.098467} & \num{0.373968 +- 0.059963} & \num{0.615244 +- 0.159824} & \num{0.644924 +- 0.148123} & \num{0.632282 +- 0.077136} & \num{0.712363 +- 0.039879} \\
\cmidrule(r){2-9}
 & \gls{gl:presto} & \bfseries \num{0.297131 +- 0.192151} & \num{0.423561 +- 0.298759} & \num{0.400713 +- 0.134044} & \num{0.565473 +- 0.146334} & \num{0.666298 +- 0.138955} & \num{0.673059 +- 0.109046} & \num{0.733868 +- 0.043204} \\
 & \gls{gl:prestoxts} & \num{0.000000 +- 0.000000} & \num{0.300137 +- 0.353039} & \bfseries \num{0.542803 +- 0.220247} & \bfseries \color{blue} \num{0.910675 +- 0.060606} & \num{0.745013 +- 0.181187} & \num{0.584875 +- 0.056615} & \bfseries \num{0.743706 +- 0.018720} \\
 & \gls{gl:crossprestoxts} & \num{0.119325 +- 0.255484} & \bfseries \num{0.598313 +- 0.472879} & \num{0.523923 +- 0.234602} & \num{0.885522 +- 0.056767} & \bfseries \num{0.756573 +- 0.067163} & \bfseries \num{0.752427 +- 0.213565} & \num{0.708085 +- 0.060650} \\
\bottomrule
\end{tabular}

  \end{subtable}%
\end{table*}

\Cref{tab:best_results_S2} highlights the results when fine-tuning the pre-trained models on the data from Estonia. 
\Cref{tab:best_results_S2_meadow} and \cref{tab:best_results_S2_nomeadow} provide additional insights into a more comprehensive analysis of the impact of the \meadow class and minority-class performance.

\subsubsection{Supervised learning}
\begin{table*}[t]
    \caption{
        Detailed analysis of supervised algorithms in the \num{20}-shot \taskStwo\ benchmark setting. 
        The metric \emph{classification accuracy} was evaluated on subsets of the test dataset corresponding to classes that are overlapping, overlapping excluding \meadow, or non-overlapping (\cf \cref{fig:overlap_diagrams}) between the countries for the best performing pre-training scenario (in terms of accuracy on the validation dataset) per algorithm.
        We report the mean $\pm$ standard deviation over five repeated experimental runs in all cases on the test set.
        The best result per subset is marked in \textbf{bold}.\\
        \textsuperscript{\textdaggerdbl}\,The results on all Estonia classes are identical with the results in \cref{tab:best_results_S2}.%
    }
    \label{tab:best_results_overlap}

  \begin{subtable}{\linewidth}
    \caption{\num{20}-shot \lvee benchmark task.}
    \label{tab:overlap_results_lv}
    
    \centering\scriptsize

\robustify\bfseries
\sisetup{detect-all=true,uncertainty-mode=separate,table-align-uncertainty=true,round-mode=uncertainty,round-precision=2}

\begin{tabular}{llSSSSSS}
\toprule
{} & {class subset} & {EE\textsuperscript{\textdaggerdbl}} & {EE $\cap$ LV} & {(EE $\cap$ LV) $\setminus$ $\{\text{meadow}\}$} & {EE $\setminus$ LV} \\
{} & {algorithm} & {\raisebox{-.25\height}{\resizebox{!}{0.35cm}{\definecolor{mediumseagreen}{RGB}{85,168,104}
\definecolor{peru}{RGB}{221,132,82}
\definecolor{steelblue}{RGB}{76,114,176}
\begin{tikzpicture}
\scriptsize
\coordinate (LV) at (0,0);
\coordinate (EE) at (1,0);
\coordinate (cap) at ($(LV)!0.5!(EE)$);

\begin{scope}[blend group=soft light]
    \clip (EE) circle[radius=1.5];
    \fill[steelblue, opacity=0.45] (LV) circle[radius=1.5];
    \fill[peru, blend mode=soft light, opacity=0.45] (EE) circle[radius=1.5];
\end{scope}

\draw (LV) circle[radius=1.5];
\draw (EE) circle[radius=1.5];
\end{tikzpicture}}}} & {\raisebox{-.25\height}{\resizebox{!}{0.35cm}{\definecolor{mediumseagreen}{RGB}{85,168,104}
\definecolor{peru}{RGB}{221,132,82}
\definecolor{steelblue}{RGB}{76,114,176}
\begin{tikzpicture}
\scriptsize
\coordinate (LV) at (0,0);
\coordinate (EE) at (1,0);
\coordinate (cap) at ($(LV)!0.5!(EE)$);

\begin{scope}[blend group=soft light]
    \clip (EE) circle[radius=1.5];
    \clip (LV) circle[radius=1.5];
    \fill[steelblue, opacity=0.45] (LV) circle[radius=1.5];
    \fill[peru, opacity=0.45] (EE) circle[radius=1.5];
\end{scope}

\draw (LV) circle[radius=1.5];
\draw (EE) circle[radius=1.5];
\end{tikzpicture}}}} & {\raisebox{-.25\height}{\resizebox{!}{0.35cm}{\definecolor{mediumseagreen}{RGB}{85,168,104}
\definecolor{peru}{RGB}{221,132,82}
\definecolor{steelblue}{RGB}{76,114,176}
\begin{tikzpicture}
\scriptsize
\coordinate (LV) at (0,0);
\coordinate (EE) at (1,0);
\coordinate (cap) at ($(LV)!0.5!(EE)$);

\begin{scope}[blend group=soft light]
    \clip (EE) circle[radius=1.5];
    \clip (LV) circle[radius=1.5];
    \fill[steelblue, opacity=0.45] (LV) circle[radius=1.5];
    \fill[peru, opacity=0.45] (EE) circle[radius=1.5];
\end{scope}

\draw (LV) circle[radius=1.5];
\draw (EE) circle[radius=1.5];
\end{tikzpicture}}} $-$ \raisebox{-.25\height}{\resizebox{!}{0.35cm}{\definecolor{mediumseagreen}{RGB}{85,168,104}
\definecolor{peru}{RGB}{221,132,82}
\definecolor{steelblue}{RGB}{76,114,176}
\begin{tikzpicture}
\scriptsize
\fill[shading=axis, left color=mediumseagreen!75!white, right color=mediumseagreen!5!black, shading angle=45] (-1.25,0) to[bend right] (-2.5,3) to[bend left] (-0.5,0) -- cycle;
\fill[shading=axis, left color=mediumseagreen!75!white, right color=mediumseagreen!5!black, shading angle=45] (-.75,0) to[bend right] (-1.5,3)  to[bend left] (0.25,0);
\fill[shading=axis, left color=mediumseagreen!75!white, right color=mediumseagreen!5!black, shading angle=45] (.75,0) to[bend left] (1.5,3) to[bend right] (-.25,0);
\fill[shading=axis, left color=mediumseagreen!75!white, right color=mediumseagreen!5!black, shading angle=45] (1.25,0) to[bend left] (2.5,3) to[bend right] (0.5,0);
\end{tikzpicture}}}} & {\raisebox{-.25\height}{\resizebox{!}{0.35cm}{
\definecolor{mediumseagreen}{RGB}{85,168,104}
\definecolor{peru}{RGB}{221,132,82}
\definecolor{steelblue}{RGB}{76,114,176}
\begin{tikzpicture}
\scriptsize
\coordinate (LV) at (0,0);
\coordinate (EE) at (1,0);
\coordinate (cap) at ($(LV)!0.5!(EE)$);

\begin{scope}[blend group=soft light,even odd rule]
    \clip (EE) circle[radius=1.5] (LV) circle[radius=1.5];
    \fill[peru, opacity=0.45] (EE) circle[radius=1.5];
\end{scope}

\draw (LV) circle[radius=1.5];
\draw (EE) circle[radius=1.5];
\end{tikzpicture}
}}} \\

\cmidrule(r){1-2} \cmidrule(lr){3-3} \cmidrule(lr){4-4} \cmidrule(lr){5-5} \cmidrule(l){6-6}

\multirow[c]{7}{*}{\rotatebox[origin=c]{90}{accuracy}}
    & no pre-training & \num{0.490370 +- 0.019933} & \num{0.541523 +- 0.021532} & \num{0.169588 +- 0.116429} & \num{0.021387 +- 0.021787} \\
    \cmidrule{2-6}
    & transfer learning & \num{0.364550 +- 0.081318} & \num{0.390606 +- 0.102377} & \num{0.328404 +- 0.106270} & \bfseries \num{0.152312 +- 0.117436} \\
    & ANIL & \num{0.490296 +- 0.068658} & \num{0.536240 +- 0.076901} & \bfseries \num{0.363785 +- 0.063677} & \num{0.069075 +- 0.013981} \\
    & MAML & \bfseries \num{0.506549 +- 0.056659} & \bfseries \num{0.555854 +- 0.062598} & \num{0.334690 +- 0.058117} & \num{0.054509 +- 0.069458} \\
    & FOMAML & \num{0.475675 +- 0.039138} & \num{0.500132 +- 0.089555} & \num{0.204809 +- 0.097100} & \num{0.020058 +- 0.022021} \\
    & TIML (encoder) & \num{0.423427 +- 0.067324} & \num{0.464712 +- 0.078335} & \num{0.177970 +- 0.040149} & \num{0.044913 +- 0.037296} \\
    & TIML (no encoder) & \num{0.456290 +- 0.058360} & \num{0.499256 +- 0.068170} & \num{0.307569 +- 0.104793} & \num{0.062370 +- 0.044433} \\
\bottomrule
\end{tabular}

  \end{subtable}

  \bigskip

  \begin{subtable}{\linewidth}
    \caption{\num{20}-shot \lvptee benchmark task. 
    Cases in which the \lvptee was an improvement over the corresponding \lvee result in \cref{tab:overlap_results_lv} are highlighted  with \textcolor{mediumseagreen}{green color}.}
    \label{tab:overlap_results_lvpt}
    \centering\scriptsize

\robustify\bfseries
\sisetup{detect-all=true,uncertainty-mode=separate,table-align-uncertainty=true,round-mode=uncertainty,round-precision=2}

\begin{tabular}{llSSSSSS}
\toprule
{} & {class subset} & {EE\textsuperscript{\textdaggerdbl}} & {EE $\cap$ LV $\cap$ PT} & {(EE $\cap$ LV $\cap$ PT) $\setminus$ $\{\text{meadow}\}$} & {(EE $\cap$ LV) $\setminus$ PT} & {(EE $\cap$ PT) $\setminus$ LV} & {EE $\setminus$ (LV $\cup$ PT)} \\
{} & {algorithm} & {\raisebox{-.25\height}{\resizebox{!}{0.5cm}{\definecolor{mediumseagreen}{RGB}{85,168,104}
\definecolor{peru}{RGB}{221,132,82}
\definecolor{steelblue}{RGB}{76,114,176}
\begin{tikzpicture}
\scriptsize
\coordinate (LV) at (210:0.75);
\coordinate (PT) at (330:0.75);
\coordinate (EE) at (90:0.75);

\begin{scope}[blend group=soft light]
    \clip (EE) circle[radius=1.5];
    \fill[steelblue, opacity=0.45] (LV) circle[radius=1.5];
    \fill[mediumseagreen, opacity=0.45] (PT) circle[radius=1.5];
    \fill[peru, opacity=0.45] (EE) circle[radius=1.5];
\end{scope}

\draw (LV) circle[radius=1.5];
\draw (PT) circle[radius=1.5];
\draw (EE) circle[radius=1.5];

\end{tikzpicture}}}} & {\raisebox{-.25\height}{\resizebox{!}{0.5cm}{\definecolor{mediumseagreen}{RGB}{85,168,104}
\definecolor{peru}{RGB}{221,132,82}
\definecolor{steelblue}{RGB}{76,114,176}
\begin{tikzpicture}
\scriptsize
\coordinate (LV) at (210:0.75);
\coordinate (PT) at (330:0.75);
\coordinate (EE) at (90:0.75);

\begin{scope}[blend group=soft light]
    \clip (LV) circle[radius=1.5];
    \clip (PT) circle[radius=1.5];
    \clip (EE) circle[radius=1.5];
    \fill[steelblue, opacity=0.45] (LV) circle[radius=1.5];
    \fill[mediumseagreen, opacity=0.45] (PT) circle[radius=1.5];
    \fill[peru, opacity=0.45] (EE) circle[radius=1.5];
\end{scope}

\draw (LV) circle[radius=1.5];
\draw (PT) circle[radius=1.5];
\draw (EE) circle[radius=1.5];

\end{tikzpicture}}}} & {\raisebox{-.25\height}{\resizebox{!}{0.5cm}{\definecolor{mediumseagreen}{RGB}{85,168,104}
\definecolor{peru}{RGB}{221,132,82}
\definecolor{steelblue}{RGB}{76,114,176}
\begin{tikzpicture}
\scriptsize
\coordinate (LV) at (210:0.75);
\coordinate (PT) at (330:0.75);
\coordinate (EE) at (90:0.75);

\begin{scope}[blend group=soft light]
    \clip (LV) circle[radius=1.5];
    \clip (PT) circle[radius=1.5];
    \clip (EE) circle[radius=1.5];
    \fill[steelblue, opacity=0.45] (LV) circle[radius=1.5];
    \fill[mediumseagreen, opacity=0.45] (PT) circle[radius=1.5];
    \fill[peru, opacity=0.45] (EE) circle[radius=1.5];
\end{scope}

\draw (LV) circle[radius=1.5];
\draw (PT) circle[radius=1.5];
\draw (EE) circle[radius=1.5];

\end{tikzpicture}}} $-$ \raisebox{-.25\height}{\resizebox{!}{0.35cm}{\definecolor{mediumseagreen}{RGB}{85,168,104}
\definecolor{peru}{RGB}{221,132,82}
\definecolor{steelblue}{RGB}{76,114,176}
\begin{tikzpicture}
\scriptsize
\fill[shading=axis, left color=mediumseagreen!75!white, right color=mediumseagreen!5!black, shading angle=45] (-1.25,0) to[bend right] (-2.5,3) to[bend left] (-0.5,0) -- cycle;
\fill[shading=axis, left color=mediumseagreen!75!white, right color=mediumseagreen!5!black, shading angle=45] (-.75,0) to[bend right] (-1.5,3)  to[bend left] (0.25,0);
\fill[shading=axis, left color=mediumseagreen!75!white, right color=mediumseagreen!5!black, shading angle=45] (.75,0) to[bend left] (1.5,3) to[bend right] (-.25,0);
\fill[shading=axis, left color=mediumseagreen!75!white, right color=mediumseagreen!5!black, shading angle=45] (1.25,0) to[bend left] (2.5,3) to[bend right] (0.5,0);
\end{tikzpicture}}}} & {\raisebox{-.25\height}{\resizebox{!}{0.5cm}{\definecolor{mediumseagreen}{RGB}{85,168,104}
\definecolor{peru}{RGB}{221,132,82}
\definecolor{steelblue}{RGB}{76,114,176}
\begin{tikzpicture}
\scriptsize
\coordinate (LV) at (210:0.75);
\coordinate (PT) at (330:0.75);
\coordinate (EE) at (90:0.75);

\begin{scope}[blend group=soft light,even odd rule]
    \clip (LV) circle[radius=1.5] (PT) circle[radius=1.5];
    \clip (EE) circle[radius=1.5] (PT) circle[radius=1.5];
    \fill[steelblue, opacity=0.45] (LV) circle[radius=1.5];
    \fill[peru, opacity=0.45] (EE) circle[radius=1.5];
\end{scope}

\draw (LV) circle[radius=1.5];
\draw (PT) circle[radius=1.5];
\draw (EE) circle[radius=1.5];

\end{tikzpicture}}}} & {\raisebox{-.25\height}{\resizebox{!}{0.5cm}{\definecolor{mediumseagreen}{RGB}{85,168,104}
\definecolor{peru}{RGB}{221,132,82}
\definecolor{steelblue}{RGB}{76,114,176}
\begin{tikzpicture}
\scriptsize
\coordinate (LV) at (210:0.75);
\coordinate (PT) at (330:0.75);
\coordinate (EE) at (90:0.75);

\begin{scope}[blend group=soft light,even odd rule]
    \clip (PT) circle[radius=1.5] (LV) circle[radius=1.5];
    \clip (EE) circle[radius=1.5] (LV) circle[radius=1.5];
    \fill[mediumseagreen, opacity=0.45] (PT) circle[radius=1.5];
    \fill[peru, opacity=0.45] (EE) circle[radius=1.5];
\end{scope}

\draw (LV) circle[radius=1.5];
\draw (PT) circle[radius=1.5];
\draw (EE) circle[radius=1.5];

\end{tikzpicture}}}} & {\raisebox{-.25\height}{\resizebox{!}{0.5cm}{\definecolor{mediumseagreen}{RGB}{85,168,104}
\definecolor{peru}{RGB}{221,132,82}
\definecolor{steelblue}{RGB}{76,114,176}
\begin{tikzpicture}
\scriptsize
\coordinate (LV) at (210:0.75);
\coordinate (PT) at (330:0.75);
\coordinate (EE) at (90:0.75);

\begin{scope}[blend group=soft light,even odd rule]
    \clip (EE) circle[radius=1.5] (LV) circle[radius=1.5];
    \clip (EE) circle[radius=1.5] (PT) circle[radius=1.5];
    \fill[peru, opacity=0.45] (EE) circle[radius=1.5];
\end{scope}

\draw (LV) circle[radius=1.5];
\draw (PT) circle[radius=1.5];
\draw (EE) circle[radius=1.5];

\end{tikzpicture}}}} \\

\cmidrule(r){1-2} \cmidrule(lr){3-3} \cmidrule(lr){4-4} \cmidrule(lr){5-5} \cmidrule(lr){6-6} \cmidrule(lr){7-7} \cmidrule(l){8-8}

\multirow[c]{7}{*}{\rotatebox[origin=c]{90}{accuracy}}
    & no pre-training & \bfseries \num{0.490370 +- 0.019933} & \bfseries \num{0.647769 +- 0.041757} & \num{0.078658 +- 0.077138} & \num{0.241754 +- 0.152647} & \num{0.015724 +- 0.014936} & \num{0.045917 +- 0.057862} \\
    \cmidrule{2-8}
    & transfer learning & \num{0.292206 +- 0.089407} & \num{0.275633 +- 0.136768} & \num{0.131937 +- 0.027773} & \num{0.415299 +- 0.051816} & \bfseries \num{0.094984 +- 0.082368} & \num{0.170108 +- 0.032277} \\
    & ANIL & \num{0.479802 +- 0.044880} & \num{0.562914 +- 0.082066} & \num{0.128871 +- 0.059741} & \num{0.412697 +- 0.090594} & \num{0.052864 +- 0.059624} & \num{0.187982 +- 0.107561} \\
    & MAML &  \num{0.452282 +- 0.093832} & \num{0.514735 +- 0.135319} & \bfseries \num{0.147086 +- 0.043765} & \num{0.437297 +- 0.103329} & \num{0.035076 +- 0.029090} & \num{0.197227 +- 0.105516} \\
    & FOMAML & \num{0.445086 +- 0.058282} & \num{0.487853 +- 0.103609} & \num{0.146873 +- 0.078055} & \num{0.474666 +- 0.083065} & \num{0.063892 +- 0.092618} & \num{0.174422 +- 0.141850} \\
    & TIML (encoder) & \color{mediumseagreen} \num{0.438446 +- 0.030566} & \num{0.581880 +- 0.071610} & \num{0.063783 +- 0.047986} & \num{0.203927 +- 0.123764} & \num{0.028887 +- 0.036597} & \num{0.035747 +- 0.053336} \\
    & TIML (no encoder) & \num{0.441015 +- 0.069749} & \num{0.475624 +- 0.109906} & \num{0.118883 +- 0.036085} & \bfseries \num{0.493916 +- 0.056214} & \num{0.047314 +- 0.035781} & \bfseries \num{0.220647 +- 0.071416} \\
\bottomrule
\end{tabular}

  \end{subtable}
  
\end{table*}

\Cref{tab:best_results_S2} shows the overall accuracy for each algorithm and task.
For \lvee, the standard transfer learning approach did not lead to any improvement in precision, compared to training from scratch.
Algorithms belonging to the \gls{gl:maml} family (\ie \gls{gl:anil}, (FO-)\gls{gl:maml}) tend to perform better as the number of fine-tuning shots increases, reaching a maximum of $a_\text{OA} \approx \qty{65}{\percent}$ for the \num{500}-shot benchmark task with \gls{gl:maml}.
Furthermore, \gls{gl:timl} (without an encoder) shows comparable results.
The encoder version of \gls{gl:timl} achieves the highest classification accuracy for the extremely data-scarce \num{1}-shot benchmark task, outperforming alternative approaches by a significant margin.
However, this approach exhibits suboptimal performance for tasks with \num{100} or more shots, in comparison to the remaining \lvee algorithms.

Compared to pre-training on \latvia only, the incorporation of additional data from Portugal during pre-training (\lvptee) did not result in general improvements during fine-tuning.
This becomes quite evident when examining the baseline model that has not been exposed to any data from Portugal.
It outperforms all supervised \lvptee algorithms for the \num{1}-, \num{5}-, \num{10}- and \num{20}-shot benchmark tasks.

Surprisingly, pre-training on Portuguese data often boosts performance on minority classes (\cref{tab:best_results_S2_nomeadow}).
While the majority of methods show a drop in accuracy, models exposed to Portugal data in \num{1}- and \num{5}-shot settings achieve even higher minority-class accuracies than overall accuracies (\cf  \cref{tab:best_results_S2}).
Most notably, \lvptee attains the highest scores for \num{1}-, \num{5}-, \num{100}-, and \num{200}-shots through various meta-learning algorithms.

As might be reasonably anticipated, the results displayed in \cref{tab:best_results_S2_meadow} show a contrary behavior.
While the no-pretraining baseline is leading the \num{5}- and \num{10}-shot tasks, pre-training on Latvian data yields the highest performances, with \gls{gl:maml} winning from \num{100} shots onward.
Regular transfer learning, while achieving strong results on minority-classes, is encountering difficulties when being evaluated exclusively on the meadow class.

\Cref{tab:overlap_results_lv} (\lvee) and \cref{tab:overlap_results_lvpt} (\lvptee) enhance the understanding through a more comprehensive analysis of shared and non-shared class subsets between pre-training and fine-tuning countries.
In particular, they demonstrate the dominance of the \meadow class in the learned data.%
\footnote{\label{ftn:20shot} The selected \num{20}-shot scenario serves as a representative case, as presenting these results for all $k$-shot settings would exceed the scope of this section.}
For the overlapping classes (\estonia $\cap$ \latvia) in the learning task \lvee (\cf \cref{tab:overlap_results_lv}), the \gls{gl:maml} algorithm demonstrates the best performance, attaining an accuracy of approximately $56\%$.
In the context of the learning task pertaining to Portuguese data, as indicated in \cref{tab:overlap_results_lvpt}, the accuracy is generally lower than in the \lvee case (\cf \cref{tab:overlap_results_lv}), with the exception of \gls{gl:timl} (with encoder) in the \num{1}-shot task.
Furthermore, taking into account all classes (\estonia) as well as only overlapping classes (\estonia $\cap$ \latvia $\cap$ \portugal), all Portugal pre-training methods are even outperformed by the no-pretraining baseline.
In both cases (\lvee and \lvptee), the few-shot performance of all algorithms is significantly lower when evaluated on those classes not previously seen during pre-training (\estonia $\setminus$ \latvia  and \estonia $\setminus$ (\latvia $\cup$ \portugal)).

\subsubsection{Self-supervised learning}
The \gls{gl:ssl} methods do not fully match the overall accuracy of the meta-learning methods, \cf \cref{tab:best_results_S2,tab:best_results_S2_nomeadow,tab:best_results_S2_meadow}.
However, in terms of overall accuracy (\cref{tab:best_results_S2}) and meadow-class accuracy (\cref{tab:best_results_S2_meadow}), \gls{gl:crossprestoxts} consistently outperforms standard transfer learning, except in two settings.
Moreover, it shows the strongest \gls{gl:ssl} performance when it comes to overall accuracy, in particular in low-shot regimes ($ 2\leq k \leq 20$).

When excluding the dominant class from the evaluation (\cref{tab:best_results_S2_nomeadow}), all models drop in accuracy, yet \gls{gl:ssl} often outperforms the no-pretraining baseline on the minority classes.
Notably, in the \num{500}-shot regime, \gls{gl:prestoxts} and \gls{gl:crossprestoxts} even exceed several meta-learning algorithms, with \gls{gl:prestoxts} achieving up to \num{13}\% higher accuracy.

On the contrary, while \gls{gl:prestoxts} shows strong performance on minority classes for \num{1}- and \num{5}-shots, it struggles to capture the \meadow class for these scenarios, compared to its \gls{gl:ssl} counterparts, misclassifying all samples for the \num{1}-shot task. 
Nevertheless, the algorithm's performance improves with an increasing number of samples per class, even showing the overall highest score for the \num{20}-shot task.

\subsection{Results on \taskalldata}
\label{sec:results_taskalldata}

The baseline model trained on all data of the \taskalldata\ task yields $a_\text{OA} = \num{0.768(0.010)}$, again serving as a general reference for evaluating the task's difficulty.

\subsubsection{Self-supervised learning}
\begin{table*}[ht!]
    \caption{%
    Results for the \taskalldata\ fine-tuning benchmark.
    We report \emph{classification accuracy} on the test set for each algorithm and few-shot task.
    Metrics are shown as mean $\pm$ standard deviation over five runs, \cf \cref{sec:exp_finetuning}.
    The best result per few-shot task is marked in \textbf{bold}.
    }
    \label{tab:comparison_results_eurocropsml_S1S2ERA5}
    \begin{subtable}{\linewidth}
        \caption{Fine-tuning classification results for \taskalldata.
        \label{tab:best_results_S1S2ERA5}}
        \centering\scriptsize

\robustify\bfseries
\sisetup{detect-all=true,uncertainty-mode=separate,table-align-uncertainty=true,round-mode=uncertainty,round-precision=2}
\begin{tabular}{llSSSSSSS}
\toprule
{} & {benchmark task ($k$-shot)} & {1} & {5} & {10} & {20} & {100} & {200} & {500} \\
{} & {algorithm} & {} & {} & {} & {} & {} & {} & {} \\

\cmidrule(r){1-2} \cmidrule(lr){3-3} \cmidrule(lr){4-4} \cmidrule(lr){5-5} \cmidrule(lr){6-6} \cmidrule(lr){7-7} \cmidrule(lr){8-8} \cmidrule(l){9-9}

\multirow[c]{4}{*}{\rotatebox[origin=c]{90}{accuracy}}
  & no pre-training & \bfseries \num{0.223717 +- 0.166993} & \bfseries \num{0.432835 +- 0.061911} & \bfseries \num{0.285055 +- 0.104447} & \bfseries \num{0.457336 +- 0.020676} & \num{0.459201 +- 0.036465} & \num{0.493372 +- 0.022448} & \num{0.597084 +- 0.033760} \\
  \cmidrule{2-9}
 & \gls{gl:presto} & \num{0.121147 +- 0.046465} & \num{0.229129 +- 0.098702} & \num{0.249099 +- 0.028409} & \num{0.360741 +- 0.054982} & \num{0.475163 +- 0.044415} & \bfseries \num{0.516304 +- 0.031219} & \num{0.611102 +- 0.014753} \\
 & \gls{gl:prestoxts} & \num{0.181877 +- 0.190998} & \num{0.398084 +- 0.071412} & \num{0.236166 +- 0.052803} & \num{0.433921 +- 0.064601} & \bfseries \num{0.478529 +- 0.023061} & \num{0.497334 +- 0.030869} & \bfseries \num{0.614610 +- 0.025749} \\
 & \gls{gl:crossprestoxts} & \num{0.126497 +- 0.197812} & \num{0.083042 +- 0.012605} & \num{0.146461 +- 0.031735} & \num{0.367694 +- 0.068855} & \num{0.433204 +- 0.047995} & \num{0.452913 +- 0.060877} & \num{0.595168 +- 0.021535} \\
\bottomrule
\end{tabular}

    \end{subtable}

    \medskip

    \begin{subtable}{\linewidth}
        \caption{Fine-tuning classification results for \taskalldata\ excluding the majority class \meadow. 
        \label{tab:best_results_S1S2ERA5_nomeadow}}
        \centering\scriptsize

\robustify\bfseries
\sisetup{detect-all=true,uncertainty-mode=separate,table-align-uncertainty=true,round-mode=uncertainty,round-precision=2}
\begin{tabular}{llSSSSSSS}
\toprule
{} & {benchmark task ($k$-shot)} & {1} & {5} & {10} & {20} & {100} & {200} & {500} \\
{} & {algorithm} & {} & {} & {} & {} & {} & {} & {} \\

\cmidrule(r){1-2} \cmidrule(lr){3-3} \cmidrule(lr){4-4} \cmidrule(lr){5-5} \cmidrule(lr){6-6} \cmidrule(lr){7-7} \cmidrule(lr){8-8} \cmidrule(l){9-9}

\multirow[c]{4}{*}{\rotatebox[origin=c]{90}{accuracy}}
 & no pre-training & \num{0.031380 +- 0.061680} & \num{0.002485 +- 0.005345} & \num{0.030083 +- 0.053043} & \num{0.005613 +- 0.006823} & \num{0.144586 +- 0.125279} & \num{0.280691 +- 0.182681} & \num{0.472048 +- 0.030169} \\
   \cmidrule{2-9}
 & \gls{gl:presto} & \num{0.024633 +- 0.022671} & \num{0.052450 +- 0.044786} & \num{0.087438 +- 0.081226} & \bfseries \num{0.123996 +- 0.072112} & \bfseries \num{0.339779 +- 0.087877} & \num{0.388152 +- 0.058635} & \num{0.462161 +- 0.057911} \\
 & \gls{gl:prestoxts} & \num{0.004316 +- 0.005937} & \num{0.001155 +- 0.001090} & \num{0.043087 +- 0.048405} & \num{0.036710 +- 0.048871} & \num{0.265148 +- 0.170441} & \num{0.319680 +- 0.212782} & \bfseries \num{0.539986 +- 0.016195} \\
 & \gls{gl:crossprestoxts} & \bfseries \num{0.058521 +- 0.053332} & \bfseries \num{0.093902 +- 0.046781} & \bfseries \num{0.119418 +- 0.034722} & \num{0.086948 +- 0.074571} & \num{0.225560 +- 0.149602} & \bfseries \num{0.427620 +- 0.087591} & \num{0.490196 +- 0.017875} \\
\bottomrule
\end{tabular}

    \end{subtable}
    
    \medskip

    \begin{subtable}{\linewidth}
        \caption{Fine-tuning classification results for \taskalldata\ on the \meadow class.
        \label{tab:results_S1S2ERA5_meadow}}
        \centering\scriptsize

\robustify\bfseries
\sisetup{detect-all=true,uncertainty-mode=separate,table-align-uncertainty=true,round-mode=uncertainty,round-precision=2}
\begin{tabular}{llSSSSSSS}
\toprule
{} & {benchmark task ($k$-shot)} & {1} & {5} & {10} & {20} & {100} & {200} & {500} \\
{} & {algorithm} & {} & {} & {} & {} & {} & {} & {} \\

\cmidrule(r){1-2} \cmidrule(lr){3-3} \cmidrule(lr){4-4} \cmidrule(lr){5-5} \cmidrule(lr){6-6} \cmidrule(lr){7-7} \cmidrule(lr){8-8} \cmidrule(l){9-9}

\multirow[c]{4}{*}{\rotatebox[origin=c]{90}{accuracy}} & No pretraining & \bfseries \num{0.433375 +- 0.388963} & \bfseries \num{0.901943 +- 0.131033} & \bfseries \num{0.562989 +- 0.269844} & \bfseries \num{0.949742 +- 0.049380} & \bfseries \num{0.802151 +- 0.158618} & \bfseries \num{0.725206 +- 0.171113} & \num{0.733381 +- 0.059773} \\
 \cmidrule(r){2-9}
 & \gls{gl:presto} & \num{0.226353 +- 0.109299} & \num{0.421719 +- 0.237309} & \num{0.425319 +- 0.120005} & \num{0.618808 +- 0.185503} & \num{0.622741 +- 0.162318} & \num{0.655997 +- 0.074844} & \bfseries \num{0.773457 +- 0.076384} \\
 & \gls{gl:prestoxts} & \num{0.375429 +- 0.400068} & \num{0.830761 +- 0.149500} & \num{0.446635 +- 0.155873} & \num{0.866904 +- 0.175286} & \num{0.711127 +- 0.201800} & \num{0.690988 +- 0.209047} & \num{0.695954 +- 0.056890} \\
 & \gls{gl:crossprestoxts} & \num{0.200594 +- 0.446386} & \num{0.071205 +- 0.070144} & \num{0.175940 +- 0.100282} & \num{0.673724 +- 0.211427} & \num{0.659550 +- 0.252393} & \num{0.480485 +- 0.058832} & \num{0.709594 +- 0.035132} \\
\bottomrule
\end{tabular}

  \end{subtable}%

\end{table*}

The \gls{gl:ssl} methods struggle in low few-shot tasks when evaluated on overall accuracy, exhibiting a consistent deficit from the no-pretraining baseline (\cref{tab:best_results_S1S2ERA5}).
However, their performances improve significantly when exposed to a greater number of labeled samples.
Consequently, both \gls{gl:presto} and \gls{gl:prestoxts} ultimately surpass training from scratch.

Furthermore, an evaluation of minority-class performance (\cref{tab:best_results_S1S2ERA5_nomeadow}) reveals that all \gls{gl:ssl} models exceed the baseline, often by a significant margin.
It is noteworthy that \gls{gl:crossprestoxts} demonstrates a strong advantage in low few-shot tasks ($k\leq10$) as well as for \num{100} shots for minority classes, outperforming all other methods.

Concurrently, \gls{gl:prestoxts} demonstrates the highest overall and minority-class accuracy in the \num{500}-shot task, with $a_{\text{OA}}=\num{0.615(0.026)}$ and $a_{\text{MCA}}=\num{0.54(0.016)}$, surpassing no-pretraining by \num{14}\% on minority classes.
However, the model struggles significantly in low-shot regimes.

Notably, while all \gls{gl:ssl} are outperformed by the no-pretraining baseline for $1 \leq k \leq 20$ in the \taskalldata\ scenario (\cref{tab:best_results_S1S2ERA5}), this advantage is largely due to the baseline's strong bias toward the dominant meadow class (\cref{tab:results_S1S2ERA5_meadow}).

\section{Discussion}
\label{sec:discussion}
The results in \cref{sec:results}, while showing that no single algorithm dominants across all scenarios, highlight that meta-learning approaches demonstrate distinct strengths in specific tasks.

\subsection{Supervised learning}
\label{sec:discussion_supervised}
This section first analyzes the impact of pre-training on Latvian data and then examines how the inclusion of Portuguese data during pre-training affects fine-tuning performance.

\subsubsection{Effect of pre-training on data from Latvia}
\label{sec:discussion_lv}
\Cref{tab:best_results_S2} provides an overview of the benchmark results when transferring knowledge gained from Latvian data to Estonian data.
As illustrated in \cref{fig:overlap_diagrams}, certain crop-type classes are shared between Latvia, Portugal, and Estonia, while others are not.
It is therefore of interest to ascertain whether models demonstrate enhanced performance on classes they have already encountered during pre-training.

\Cref{tab:overlap_results_lv} compares performance on classes overlapping between Latvia and Estonia (\estonia $\cap$ \latvia) and those exclusive to Estonia (\estonia $\setminus$ \latvia).
The findings suggest that, in general, meta-learning approaches, specifically those belonging to the \gls{gl:maml} family, offer slight advantage in both overall accuracy and performance on overlapping classes (\estonia $\cap$ \latvia), compared to both the no-pretraining baseline and standard transfer learning.
This highlights the importance of domain similarity in few-shot learning.
Most pre-training-based methods (with the exception of \gls{gl:fomaml}) even exhibit better accuracy on classes exclusively present within Estonia (\estonia $\setminus$ \latvia) in comparison to the no-pretraining baseline.
This indicates that during pre-training, the model backbones learn transferable features useful for novel class adaptation.
This effect is most pronounced for standard transfer learning, which achieves the highest accuracy (\qty{13.6}{\percent}, compared to the next best approach \gls{gl:anil} with \qty{9.8}{\percent}) on the class \class{legumes harvested green}---the most abundant class in Estonia that is not present in Latvia.

The relative importance of the class distribution becomes clearer upon analysis of the classification accuracy on various subsets of classes\ (\cf \cref{tab:best_results_S2_nomeadow,tab:best_results_S2_meadow,tab:overlap_results_lv}).
Overall, the performance of all algorithms drops when the dominant class \meadow is excluded, and increases when evaluated exclusively on this class.
Nonetheless, transfer learning and \gls{gl:anil} appear to learn the most stable representation across all fine-tuning classes, particularly for those previously seen during pre-training (\cf \cref{tab:overlap_results_lv}), rather than disproportionately relying on the majority class.
Although both algorithms show increased accuracy on \meadow, they also maintain the most stable performance on minority classes (\cref{tab:best_results_S2_nomeadow,tab:overlap_results_lv} (\estonia $\cap$ \latvia and (\estonia $\cap$ \latvia) $\setminus$ $\{\text{meadow}\}$), compared to overall accuracy.
In particular, transfer learning achieves nearly equal performance on majority and minority class(es) in low-shot regimes.
Only with an increasing number of samples per class, the model tends to focus more heavily on the dominant class.

The steepest performance decline upon removing the majority class is observed for \gls{gl:timl} (with encoder) and the baseline model.
This suggests that \gls{gl:timl}'s strong performance in the \num{1}-shot task is largely due to its focus on the \meadow class, which is further confirmed by its high accuracy for that class in \cref{tab:best_results_S2_meadow}.
Similarly, in the \num{5}-, \num{10}-, and \num{20}-shot tasks, the no-pretraining baseline outperforms all supervised methods on the meadow class, implying it primarily defaults to predicting the majority class---explaining its strong overall accuracy in low-shot regimes.

\subsubsection{Impact of adding Portugal pre-training data}
\label{sec:discussion_lvpt}
\Cref{tab:best_results_S2} further evaluates the results of pre-training on data from both Latvia and Portugal, followed by fine-tuning on data from Estonia.
Adding Portugal to the pre-training data exposes limitations in geographical knowledge transfer: in terms of overall accuracy, none of the algorithms successfully bridges the domain gap between Portuguese and Estonian data, likely due to geographic and spectral variations.
There are several potential reasons why the addition of Portugal data may not lead to improved performance:
\begin{description}
    \item[Climate differences] Estonia and Latvia share similar climates, facilitating cross-region generalization.
    Portugal’s Mediterranean climate introduces a greater feature distribution shift, potentially leading to model bias and reduced pre-training effectiveness.
    \item[Crop diversity] As visualized in \cref{fig:overlap_diagrams}, Estonia and Latvia share more common crop types, while Portugal introduces classes absent in the Baltic states.
    This potentially leads to less effective knowledge transfer. 
    \item[Agricultural practices] Differences in crop management, such as irrigation and soil treatment, vary significantly between Portugal and the Baltic regions, making it difficult for models trained on one region to adapt to the other.
    Portuguese soils tend to be fairly arid, acidic, and rocky \citep{SOIL_Portugal} in contrast to the humid, sandy, or clay soils in Latvia \citep{Aldis2002ACS} and Estonia \citep{Kmoch21:estsoil_eh}.
    \item[Growth cycle differences] Climate-driven differences in crop phenology result in mismatched temporal patterns for otherwise similar crops, making it challenging for models to align temporal features between regions.
\end{description}

Nevertheless, \cref{tab:best_results_S2_nomeadow,tab:best_results_S2_meadow}
reveal that adding Portuguese data reduces the tendency of models to overfit to the dominant class \meadow.
In fact, the performance on minority classes (\cref{tab:best_results_S2_nomeadow}) often improves with the addition of Portugal data.
However, these gains are insufficient to offset the broader generalization gap introduced by the inclusion of geographically and phenologically dissimilar data.

\Cref{tab:overlap_results_lvpt} shows once more that when evaluating on Estonia-specific classes (\estonia $\setminus$ (\latvia $\cup$ \portugal)), most meta-learning algorithms---except \gls{gl:timl} with encoder---and conventional transfer learning outperform the baseline by a significant margin.
While a performance drop remains (as also seen in \cref{tab:overlap_results_lv} for \estonia $\setminus$ \latvia), it is less severe.
Along with the improved performance on minority classes, this suggests that in the \lvptee setting, the models shift their focus toward classes unique to Estonia.
A likely explanation is that, while adding Portuguese data increases the total number of shared classes between pre-training and fine-tuning, the spectral characteristics of classes shared between Portugal and Estonia can differ substantially.
This domain mismatch may introduce ambiguity, encouraging the models to focus more on domain-consistent classes---\ie those specific to Estonia.
At the same time, exposure to more diverse pre-training data may help the models to learn broader, more generalizable features, mitigating overfitting to the ubiquitous dominant class \meadow---shared across all countries---improving their ability to adapt to previously unseen Estonian classes.

\subsubsection{Limitations of the TIML family}
\label{sec:discussion_timl}
The algorithms of the \gls{gl:timl} family, despite having incorporated additional parcel location information, show rather mediocre performance on our multi-class cross-regional benchmark.
Although the original implementation \citep{tseng_2022_timl} successfully applied \gls{gl:timl} on a binary classification task, our extension of applying it on a multi-class benchmark reveals significant challenges.

This might be attributed to several compounding factors.
First, cross-country transfer already introduces a substantial domain shift, which seems to be further amplified by adding precise parcel location features.
The models tend to overfit to region-specific patterns that do not generalize well---especially when incorporating Portuguese parcel locations, which may encode localized agricultural characteristics that are not representative of Estonian conditions.
This issue is particularly pronounced in the \gls{gl:timl} encoder, which represents a learnable component in itself.
The encoder’s architecture likely prioritizes capturing fine-grained, region-specific features at the expense of spatial generalization.
As a result, the encoder may overfit to the pre-training regions and struggle to adapt to new geographical domains during fine-tuning.
This raises the question to what extent the encoder can effectively leverage the knowledge acquired from the pre-training on distant location information during fine-tuning.

Likewise, adding static location data as features to temporally structured time-series data, as done in \gls{gl:timl} (without encoder), can obscure the inherent temporal patterns.

Moreover, \Gls{gl:timl}---in particular with encoder---appears to struggle slightly more with the dataset's class imbalance, compared to transfer and other meta-learning variants.
A likely explanation is that the dominant meadow class is geographically ubiquitous in Estonia.
Hence, from the model's perspective, many different locations correlate with the same label, making location features a poor discriminator.
Rather than aiding class separation, they become a reinforcing signal to default to the majority class.
This suggests that the algorithm does not explicitly learn features that help distinguish minority classes and instead emphasizes broadly applicable, but ultimately uninformative, patterns.

In summary, the combination of these factors---cross-regional domain shifts, over-reliance on geographically anchored features, and encoder architectural limitations---hinders \gls{gl:timl}'s effectiveness in transferring knowledge across such heterogeneous datasets.
These findings also raise broader concerns about the utility of geographical location features in classifying different crop types, given that similar latitudes can show divergent agricultural practices.
However, a more detailed study of the \gls{gl:timl} encoder algorithm is beyond the scope of this work and will be left for future research. 

\subsubsection{Run-time analysis}
\label{sec:discussion_runtime}
Despite their superior performance, meta-learning algorithms tend to require longer training periods and greater computational resources compared to baselines.
While \gls{gl:anil} and \gls{gl:fomaml} significantly reduce computational costs compared to the standard \gls{gl:maml} algorithm, run-time remains one of their major drawbacks.

Run-times of the meta-learning algorithms are fairly stable across the two tasks with and without incorporating data from Portugal.
In contrast, the run-times of transfer learning increase substantially when incorporating the additional data from Portugal. 
A detailed run-time analysis can be found in \cref{tab:runtimes} in \cref{sec:runtimeanalysis}.

\subsection{Self-supervised learning}
\label{sec:discussion_ssl}
The results presented in \cref{sec:results} indicate that, while \glsxtrlong{gl:ssl} does not consistently match the overall accuracy of meta-learning approaches, they nonetheless offer meaningful advantages over certain methods, particularly in low-shot and minority-class scenarios.

All \gls{gl:presto} variants seem to be fairly robust to missing input data.
The robustness to incomplete spectral information is most evident in the \taskStwo\ task, where fewer spectral bands are available than during pre-training.
Despite the reduced spectral information, the \gls{gl:ssl} models maintain competitive performance, suggesting that their learned representations are resilient to input degradation.

\gls{gl:crossprestoxts} stands out as a particularly effective variant.
Despite being significantly computationally more efficient during pre-training (\cf \cref{sec:runtimeanalysis}), \gls{gl:crossprestoxts} also frequently outperforms \gls{gl:prestoxts} in the \taskStwo\ task.
Furthermore, it consistently (with only one exception) achieves higher minority-class accuracy in the \taskalldata\ scenario (\cref{tab:best_results_S1S2ERA5_nomeadow}), indicating improved generalization to underrepresented classes.

\subsection{Comparison of supervised and self-supervised methods}
\label{sec:discussion_comparison}
While \gls{gl:ssl} on general-purpose land cover data enables moderate generalization, it generally lags behind meta-learning algorithms pre-trained on crop-specific data---epecially in lower-shot regimes ($k \leq 100$).
This reflects the importance of task-specific inductive biases and domain relevance in few-shot settings.

Supervised and, in particular, meta-learning algorithms are explicitly designed to adapt rapidly to new tasks using minimal labeled data, making them particularly well-suited for crop-type classification problems, which often involve new regions, growing seasons, or crop varieties.
In contrast, \gls{gl:ssl} focuses on extracting broad, task-agnostic features from unlabeled data, which, while powerful, may lack the semantic specificity required for effective domain adaptation in agricultural settings with limited samples.

However, \gls{gl:ssl} often outperforms the no-pretraining baseline, with the greatest improvements seen for underrepresented classes as well as higher-shot settings.
It also tends to surpass traditional transfer learning methods in overall performance.

\section{Conclusion}
\label{sec:conclusion}
This study presented a comprehensive benchmark on the newly introduced
\EuroCropsML dataset, demonstrating its use for benchmarking
few-shot crop-type classification by evaluating the effectiveness
of both supervised and self-supervised methods. 
It sets a precedent for future benchmarks in agricultural remote sensing, providing a valuable resource to evaluate novel machine learning algorithms or models on real-world, multi-class and multi-region crop type data.

The evaluation spans diverse European regions, offering new insights into the effectiveness of these methods for real-world remote sensing applications in agriculture.

The results indicate that while meta-learning algorithms from the \gls{gl:maml} family generally outperform transfer learning in terms of accuracy, they come with increased computational costs and training time.
The \gls{gl:timl} family of algorithms did not show a significant improvement in prediction accuracy.
Incorporating additional data from a more distant geographical country into the pre-training stage did not lead to enhanced performance.
This indicates that regional disparities may not be conducive to effective knowledge transfer and serves to illustrate the inherent difficulties associated with the transfer of knowledge between disparate geographical regions.

In general, the study demonstrates the complexity of applying meta-learning techniques to geospatial problems.
Although meta-learning can adapt to variations within similar regions, transferring knowledge between regions with substantial differences requires careful consideration of the characteristics of the underlying data.
These findings highlight the trade-offs between computational efficiency and prediction accuracy in machine learning applications for remote sensing and crop type classification.
Hence, we encourage researchers to first investigate regular transfer learning approaches to establish a baseline for cross-country adaptation before exploring whether meta-learning techniques could improve classification performance.

Furthermore, crop classification often requires distinguishing between visually and spectrally similar classes, such as \wheat and \barley.
Prior studies \citep{STILLER2024100064,isprs-archives-XLIII-B3-2022-1327-2022} show that \gls{gl:ssl} models tend to underperform on such fine-grained distinctions, as they are not trained to optimize for class-level separation.
Similar trends have been observed in, for instance, plant phenotyping tasks, where \gls{gl:ssl} models were more sensitive to redundancy in pre-training datasets \citep{OGIDI20230037}.
Thus, while \gls{gl:ssl} is effective at leveraging unlabeled data, supervised methods currently provide a more robust solution for crop-type classification under real-world constraints.

Nevertheless, in the absence of labeled pre-training data, \gls{gl:ssl} remains a valuable strategy.
Our experiments show that especially for minority classes and with an increasing number of training samples, \gls{gl:ssl} often yields substantial improvement over training from scratch.
In addition, it frequently outperforms conventional transfer learning.

Finally, the limitations of pre-training in general are compounded by this study's real-world evaluation design.
While training splits are few-shot and mostly class-balanced, validation and test sets retain their original, highly imbalanced distributions---reflecting realistic deployment scenarios.
This poses a substantial challenge for both supervised and self-supervised methods to maintain performance on minority classes.

\section*{CRediT authorship contribution statement}
\textbf{J{.} Reuss:} Conceptualization, Data curation, Formal analysis, Investigation, Methodology, Resources, Software, Validation, Visualization, Writing -- original draft, Writing -- review \& editing. \textbf{J{.} Macdonald:} Conceptualization, Formal analysis, Investigation, Methodology, Resources, Software, Validation, Visualization, Writing -- original draft, Writing -- review \& editing. \textbf{S{.} Becker:} Conceptualization, Formal analysis, Investigation, Validation, Visualization, Writing -- original draft, Writing -- review \& editing. \textbf{E{.} Gikalo:} Data curation, Methodology, Software, Visualization, Writing -- original draft. \textbf{K{.} Schultka:} Conceptualization, Methodology, Software. \textbf{L{.} Richter:} Conceptualization, Funding acquisition, Supervision, Writing -- review \& editing. \textbf{M{.} K{\"o}rner:} Conceptualization, Funding acquisition, Project administration, Supervision, Writing -- review \& editing.

\section*{Declaration of competing interest}

The authors declare that they have no known competing financial interests or personal relationships that could have appeared to influence the work reported in this paper.

\section*{Acknowledgments}
\label{sec:acknowledgments}
The project is funded by the German Federal Ministry for Economic Affairs and Energy based on a decision by the German Bundestag under the funding reference 50EE2007B (J{.} Reuss, E{.} Gikalo, and M{.} K{\"o}rner), 50EE2007A (J{.} Macdonald, S{.} Becker, K{.} Schultka, and L{.} Richter) and 50EE2105 (M{.} K{\"o}rner). 
S{.} Becker acknowledges support by SNF Grant PZ00P2 216019.

\bibliographystyle{plainnat}
\bibliography{main}

\appendix
\section{Implementation details}
\label{sec:appendix_implementationdetails}
We implement all algorithms within the \texttt{PyTorch} framework \citep{Paszke_2019_PyTorch} for machine learning and automatic differentiation.

\subsection{Neural network architectures}
\label{sec:appendix_networkarchitecture}

For all experiments, we use a state-of-the-art Transformer architecture.
During supervised pre-training, we use a single Transformer encoder with a linear classification layer on top, which is predicts the class log-probabilities.
When pre-training in an \gls{gl:ssl} manner, we use a full Transformer encoder-decoder architecture.
Specifically, we use a single Transformer block for both the encoder (and decoder), comprising four attention heads.
Each token in the input sequence is represented by an internal embedding vector of dimension \num{128}. 
The fully connected network within a Transformer block uses a hidden dimension of $d_\text{hidden}=256$, expanding the initial embedding dimension by a factor of two.
Additive sinusoidal temporal positional encoding \citep{vaswani_vanillatransformer} with a maximum sequence length of \num{366}, accommodating daily samples over a full-year of temporal coverage, including leap years, is used.%
\footnote{\label{ftn:positonalencoding} Even though we are only looking at data from 2021, setting the maximum length of the positional encoding to 366 (leap year) provides a more flexible approach for further use cases. For instance, it allows fine-tuning of models that were pre-trained on leap year data}

\subsubsection{PRESTO}
\label{sec:appendix_presto}
The original \gls{gl:presto} implementation uses a Transformer encoder-decoder architecture with an embedding dimension of $d_\text{emb}=128$, defining the size of feature vectors throughout the Transformer blocks.
The model comprises four Transformer blocks in total---two in the encoder and two in the decoder---each using eight attention heads.
The fully connected network within each block uses a hidden dimension of $d_\text{hidden}=512$, expanding the initial embedding dimension by a factor of four.
Additive sinusoidal temporal positional encoding \citep{vaswani_vanillatransformer} with a maximum sequence length of \num{24} is used, corresponding to a monthly sampling over two consecutive years.

\subsection{Model training}
\label{sec:appendix_hyperparams}
The general training follows the two-phase approach of pre-training and fine-tuning.
Besides the actual learnable parameters of the Transformer model (weights), there are a number of non-learned hyperparameters that influence the model.
These include hyperparameters that were manually/empirically selected (batch size for the mini-batch gradient descent, annealing schedules for learning rates), as well as hyperparameters that were automatically tuned over numerous trial runs (learning rates).
The term "empirically selected" denotes that experiments were conducted with a range of values for the respective hyperparameter, with the optimal value being determined on the basis of the validation accuracy.
The manually set values, along with the valid ranges for tuned and empirically determined hyperparameters, were established on the basis of initial experiments, which showed the greatest improvements with these settings (\cf \cref{tab:hyperparameters_metalearning,tab:hyperparameters_pretraining,tab:hyperparameters_finetuning}).
The following subsections each contain an overview of the (tuned/selected) hyperparameters.
For the tuning trial runs, we employ a \gls{gl:tpe} with percentile pruning within the \texttt{Optuna} framework \citep{akiba2019optunanextgenerationhyperparameteroptimization}.
We employ a total of \num{100} warm-up steps and keep the top \num{75}th percentile of runs.
In all cases, the selection of the best hyperparameters is based on the classification accuracy on the validation dataset. 
This is also used to determine the best pre-training epoch.
Its model state is used to initialize the model backbone weights before fine-tuning.
Finally, the best fine-tuning epoch is determined by the classification accuracy on the validation dataset of the fine-tuning task.
Its model state is used to compute the final reported model accuracy on the fine-tuning test dataset.

\subsection{Meta-Learning}
\label{sec:appendix_metalearning}
All meta-learning algorithms are trained using a mini-batch stochastic gradient descent in the inner optimization loop and the Adam optimizer \citep{kingma2017adammethodstochasticoptimization} in the outer optimization loop. 
The learning rates for the inner and outer optimization loop are individually tuned on \num{20000} meta-train tasks, as described in \cref{sec:appendix_hyperparams}.
We then meta-train the models on \num{100000} meta-training tasks and validate after every 100 tasks on the meta-validation tasks (corresponding to epochs for regular training).
The classification head layer is adapted to predict a vector in $ \mathbb{R}^{n_c}$ of $n_c$ class logits that correspond to the distinct classes. 
The weights of the classification head layer are reset at the start of each new inner optimization loop to allow it to adapt to each new batch of tasks.
As mentioned in \cref{sec:experiments}, we carry out a series of experiments with different $n$-way $k$-shot-settings and inner loop steps $s$, \cf \cref{tab:hyperparameters_metalearning}.
Subsequently, all meta-trained models undergo fine-tuning on the Estonia data.

\begin{table}
    \caption{Meta-learning hyperparameters.
    The hyperparameters employed during meta-learning, along with their corresponding valid values, were either set manually/empirically, or were automatically tuned.\\
    \textsuperscript{\S}\,The encoder learning rate corresponds to the learning rate of the \gls{gl:timl} encoder and, hence, was solely employed for the purpose of \gls{gl:timl} (with encoder) training.}
    \centering\scriptsize
    \begin{tabular}[t]{@{}Xrr@{}}
      \toprule
      \multicolumn{1}{@{}X}{hyperparameter} & \multicolumn{1}{X}{value / value range} & \multicolumn{1}{X@{}}{assignment}\\
      \cmidrule(r){1-1} \cmidrule(lr){2-2} \cmidrule(l){3-3}
      tuning tasks & \num{100000} & manual \\
      tuning trials per setup & \num{36} & manual \\
      inner learning rate $\alpha$ & $\alpha \in \left[0.01, 10.0\right]$ & tuned \\
      outer learning rate $\beta$ & $\beta \in \left[0.0001, 0.1\right]$ & tuned \\
      encoder learning rate $\eta$\textsuperscript{\S} & $\eta \in \left[0.0001, 0.1\right]$ & tuned \\
      training tasks & \num{100000} & manual \\
      number of tasks per batch & \num{4} & manual \\
      \parbox{2.9cm}{number of classes per task $n$} & $n \in \{4, 10\}$ & empirical \\
      \parbox{2.9cm}{number of samples per class $k$} & $k \in \{1, 10\}$ & empirical \\
      \parbox{2.9cm}{number of inner loop gradient steps $s$} & \parbox{2.2cm}{$s \in \{1, 4, 10\}$ for $n=4$, $s=1$ for $n=10$} & empirical\\
      \bottomrule
    \end{tabular}
\label{tab:hyperparameters_metalearning}
\end{table}

\subsection{Transfer learning}
\label{sec:appendix_pretraining}
\begin{table}
    \caption{Transfer learning hyperparameters.
    The hyperparameters utilized during transfer learning, along with their corresponding valid values. 
    They were either set manually or empirically, or alternatively, were automatically tuned.}
    \centering\scriptsize
    \begin{tabular}[t]{@{}Xrr@{}}
      \toprule
      \multicolumn{1}{@{}X}{hyperparameter} & \multicolumn{1}{X}{value / value range} & \multicolumn{1}{X@{}}{assignment}\\
      \cmidrule(r){1-1} \cmidrule(lr){2-2} \cmidrule(l){3-3}
        tuning epochs & \num{150} & manual \\
        tuning trials per setup & \num{36} & manual \\
        training epochs & \num{150} & manual \\
        early stopping patience $p$ (epochs) & \num{15} & manual \\
        batch size $b$ & $b \in \{1,4,16,32,64,128,264\}$ & empirical \\
        learning rate $\alpha$ & $\alpha \in \left[0.0001, 0.01\right]$ & tuned \\
        cosine annealing cycles $c$ & $c \in \{0, 1, 2\}$ & empirical \\
      \bottomrule
    \end{tabular}

\label{tab:hyperparameters_pretraining}
\end{table}

The regular transfer learning experiments are conducted using the Adam optimizer \citep{kingma2017adammethodstochasticoptimization} with a tuned learning rate (\cf \cref{sec:appendix_hyperparams}) and a batch size of \num{128}.
We train for a maximum of \num{150} epochs on the training set, validate after each epoch on the validation set, and stop the training in case the validation loss does not decrease for more than \num{15} epochs (early stopping).
Hyperparameter tuning and training are performed with varying batch sizes $b \in \{4, 8, 16, 32, 64, 128, 264\}$.
Based on the validation accuracy, the batch size \num{128} is selected for further experimentation and subsequent fine-tuning.
Hence, with a batch size of \num{128}, we performed a series of experiments with the cosine annealing learning rate schedule.
The number of cycles is set to zero (corresponding to no cosine annealing), one, or two cycles.
We choose the most promising schedule based on the validation accuracy, which is one annealing cycle when pre-training only on data from Latvia, and no annealing when pre-training on data from Latvia and Portugal.
\Cref{tab:hyperparameters_pretraining} contains an overview of all hyperparameters used during transfer learning.

\subsection{Self-supervised learning}
\label{sec:appendix_ssl}
\begin{table}
    \caption{\Glsxtrlong{gl:ssl} hyperparameters.
    The hyperparameters utilized during \gls{gl:ssl} pre-traning, along with their corresponding valid values. 
    They were either set manually or empirically, or alternatively, were automatically tuned.}
    \centering\scriptsize
    \begin{tabular}[t]{@{}Xrr@{}}
      \toprule
      \multicolumn{1}{@{}X}{hyperparameter} & \multicolumn{1}{X}{value / value range} & \multicolumn{1}{X@{}}{assignment}\\
      \cmidrule(r){1-1} \cmidrule(lr){2-2} \cmidrule(l){3-3}
        tuning trials per setup & \num{8} & manual \\
        early stopping patience $p$ (steps) & \num{15} & manual \\
        batch size $b$ & $b \in \{128, 256\}$ & empirical \\
        learning rate $\alpha$ & $\alpha \in \left[0.00001, 0.01\right]$ & tuned \\
      \bottomrule
    \end{tabular}

\label{tab:hyperparameters_ssl}
\end{table}

In alignment with the aforementioned transfer learning, all \gls{gl:ssl} models are pre-trained using the Adam optimizer with a tuned learning rate (\cf \cref{sec:appendix_hyperparams}) and a batch size of \num{256} for one epoch.
After \num{250000} training batches (steps), we validate on the validation data.
Hyperparameter tuning and training are performed with varying batch sizes $b \in \{128, 256\}$.
The selection of batch sizes and number of epochs is based on the computational constraints provided by the system.
Based on the validation accuracy, the batch size \num{256} is selected for further experimentation (\cf \cref{sec:appendix_additionalresults}) and subsequent fine-tuning.
\Cref{tab:hyperparameters_ssl} contains an overview of all hyperparameters used during \glsxtrlong{gl:ssl}.

\subsection{Fine-tuning}
\label{sec:appendix_finetuning}
For the models that were pre-trained via \gls{gl:ssl}, we replace the models' decoder with a simple linear classification layer to predict class log-probabilities.
For the meta-learned and pre-trained models, we simply reset the classification head layer.
All models are fine-tuned for a maximum of \num{200} epochs on Estonia tasks using an Adam optimizer \citep{kingma2017adammethodstochasticoptimization}.
We validate every epoch on the validation set. 
Training is stopped if the validation loss does not decrease for more than five epochs (early stopping). 
Furthermore, we adjust the learning rate(s) in three different settings as described in \cref{sec:exp_finetuning}.
\Cref{tab:comparison_learningrate_settings_S2,tab:comparison_learningrate_settings_S1S2ERA5} illustrate the results of the \num{20}-shot task, exemplifying the impact of the above learning rate settings.\\
All experiments are repeated five times with five different random seeds $r \in \{0, 1, 42, 123, 1234\}$.
We set the random seed across multiple libraries, including Python’s built-in \texttt{random} module, \texttt{NumPy}'s random number generator, and \texttt{PyTorch}’s CPU random number generator.
Moreover, we mandate that \texttt{PyTorch} utilizes deterministic algorithms for operations on the GPU when using CuDNN, and disable the auto-tuning functionality of CuDNN.
Hence, for each random seed $r$, the experiments are almost entirely deterministic.
The only remaining randomness is introduced by \texttt{Optuna}'s TPE during hyperparameter tuning.
An overview of the relevant hyperparameters is shown in \cref{tab:hyperparameters_finetuning}.

\begin{table*}
    \centering\scriptsize
    \caption{
    Fine-tuning hyperparameters.
    The hyperparameters utilized during fine-tuning, along with their corresponding valid values.
    Their final value was either assigned manually or empirically, or alternatively, was automatically tuned.
    The optimal batch size for fine-tuning, denoted by $b$, was identified through a process of fine-tuning the transfer learning models with varying fine-tuning batch sizes.
    Subsequently, the final batch size of \num{16} was selected based on the validation accuracy observed during the fine-tuning process.}
    \begin{tabular}[t]{@{}Xrrrr@{}}
      \toprule
      \multicolumn{1}{@{}X}{lr tuning} & \multicolumn{1}{X}{only head} & \multicolumn{1}{X}{different head \& backbone} & \multicolumn{1}{X}{same head \& backbone} \\
      \multicolumn{1}{@{}X}{hyperparameter} & \multicolumn{1}{X}{value / value range} & \multicolumn{1}{X}{value / value range} & \multicolumn{1}{X}{value / value range} & \multicolumn{1}{X@{}}{assignment}\\
      \cmidrule(r){1-1} \cmidrule(lr){2-2} \cmidrule(lr){3-3} \cmidrule(lr){4-4} \cmidrule(l){5-5}
      tuning epochs & \num{200} & \num{200} & \num{200} & manual \\
      tuning trials per setup & \num{8} & \num{32} & \num{8} & manual \\
      training epochs & \num{200} & \num{200} & \num{200} & manual \\
      early stopping patience $k$ (epochs) & \num{5}  & \num{5}  & \num{5} & manual \\
      batch size $b$ & \num{16} & \num{16} & \num{16} & empirical \\
      learning rate $\alpha$ & 
      \parbox{2.9cm}{$\alpha_\text{head} \in \left[\num{e-6}, \num{e-2}\right]$, $\alpha_\text{backbone}=0$} & 
      \parbox{2.9cm}{$\alpha_\text{head} \in \left[\num{e-6}, \num{e-2}\right]$, $\alpha_\text{backbone} \in \left[\num{e-6}, \num{e-3}\right]$} & 
      $\alpha \in \left[\num{e-6}, \num{e-2}\right]$ 
      & tuned \\
      \bottomrule
    \end{tabular}
\label{tab:hyperparameters_finetuning}
\end{table*}

\section{Additional results}
\label{sec:appendix_additionalresults}
The results in \cref{tab:comparison_results_eurocropsml_S2} and \cref{tab:comparison_results_eurocropsml_S1S2ERA5} show the best-performing setup for each model and algorithm, selected based on overall accuracy $a_\text{OA}$ on the validation set.
This section provides a more comprehensive overview and evaluation.
In particular, we 
\begin{itemize}
    \item evaluate different strategies for hyperparameter tuning,
    \item analyze model performance on parent-level classes,
    \item assess the impact of meta-learning parameters in an $n$-way $k$-shot $s$-step setting,
    \item investigate the impact of specific masking strategies during self-supervised pre-training,
    \item assess fine-tuning performance using a larger Transformer encoder-decoder architecture for \gls{gl:ssl},
    \item and analyze robustness under class imbalance using additional evaluation metrics.
\end{itemize}

\subsection{Impact of learning rate tuning}
\label{sec:appendix_learningrates}
During fine-tuning, we considered three different hyperparameter tuning scenarios for the learning rate(s) of each algorithm, \cf \cref{sec:exp_finetuning}: using the \emph{same} learning rate for the \emph{head \& backbone}, allowing \emph{different} learning rates for the \emph{head \& backbone}, and \emph{only} tuning a learning rate for the \emph{head} while keeping the backbone fixed.
\Cref{tab:comparison_learningrate_settings_S2,tab:comparison_learningrate_settings_S1S2ERA5} illustrate the results for each algorithm on the \num{20}-shot fine-tuning benchmark tasks for the three learning rate settings.

\subsubsection{Supervised learning}
Fine-tuning both the model backbone and the model head achieved higher prediction accuracies compared to only fine-tuning the model head in all cases.
In addition, the most flexible setting with separately tuned learning rates for the model backbone and head performed best in the majority of cases.
Noteworthy, the \gls{gl:anil} algorithm, which only adapts the learnable parameters of the model head during its inner optimization loop, did not perform better than the other transfer meta-learning approaches when fine-tuning only the model head. 
While in theory using just one learning rate is less flexible than allowing separate ones for head and backbone, using just one might still turn out to be advantageous:
If the backbone and head use different learning rates, the gradient updates may vary significantly between the two, leading to an imbalance in how quickly each part adapts during training. This can cause the model to optimize one part faster than the other, potentially leading to suboptimal representations in the backbone or overfitting in the head. 

\subsubsection{Self-supervised learning}
All methods, except \gls{gl:prestoxts} for the \taskStwo\ task, achieved their highest accuracy by training both head and backbone, either with the same or with separate learning rates.
Moreover, the majority of algorithms demonstrated enhanced performance when equipped with distinct learning rates for backbone and head.
On the \taskalldata\ task in particular, all \gls{gl:ssl} methods performed substantially better with separate learning rates.
At the same time, the no pre-training baseline achieved its best performance with a single, unified learning rate for both head and backbone.
Overall, this behavior is expected: the backbone of the pre-trained models already contains learned representations, while the classification head starts from a random initialization for all models, and hence benefits from a separate, potentially higher, learning rate.

\subsection{Performance on parent-level classes}
\label{sec:appendix_parentlevel}

The hierarchical class structure of \EuroCrops supports performance analysis at multiple class levels.
This section outlines how the \EuroCropsML dataset can be used for more detailed evaluation and error analyses.
For the main evaluation in \cref{sec:results}, we used the final \gls{gl:hcat}3 level \num{6}.
Similarly, \cref{tab:results_S2_parentacc,tab:results_S1S2ERA5_parentacc} present classification accuracy based on \gls{gl:hcat}3 levels \num{3} and \num{4}.
In general, a prediction is considered correct if the predicted parent class at level $d$ matches the ground truth's parent class at the same level.
Level \num{3} includes six parent classes, while \num{4} comprises \num{33}.
It is worth noting that the \meadow class is a standalone category without any parent or child classes.
As such, it is present in level \num{3}, level \num{4}, and the final fine-grained class set used in \cref{sec:results}.
While we do not evaluate these results in detail here, they illustrate one way the dataset can support multi-level performance analysis.

\begin{table*}[ht!]
    \caption{%
    Parent-level classification results for the \taskStwo\ benchmark.
    We report \emph{parent-class classification accuracy} on the test set for each algorithm and few-shot task.
    Metrics are shown as mean $\pm$ standard deviation over five runs, \cf \cref{sec:exp_finetuning}.
    The best result per few-shot scenario is marked in {\color{blue}\textbf{blue bold}}.
    \textbf{Black bold} values highlight the best result for supervised or \gls{gl:ssl} pre-training.
    Additionally, cases in which the \lvptee was an improvement over the corresponding \lvee result are highlighted in \textcolor{mediumseagreen}{green color}.
    }
    \label{tab:results_S2_parentacc}
    \begin{subtable}{\linewidth}
        \caption{Parent-level \num{3} accuracy for fine-tuning runs on \taskStwo\, where a prediction is considered correct if the predicted class shares the same parent category (\gls{gl:hcat}3 level \num{3}) as the ground truth.
        \label{tab:results_S2_parentacc_lv3}}
        \centering\scriptsize

\robustify\bfseries
\sisetup{detect-all=true,uncertainty-mode=separate,table-align-uncertainty=true,round-mode=uncertainty,round-precision=2}

\begin{tabular}{llSSSSSSS}
\toprule
{} & {benchmark task ($k$-shot)} & {1} & {5} & {10} & {20} & {100} & {200} & {500} \\
{} & {algorithm} & {} & {} & {} & {} & {} & {} & {} \\

\cmidrule(r){1-2} \cmidrule(lr){3-3} \cmidrule(lr){4-4} \cmidrule(lr){5-5} \cmidrule(lr){6-6} \cmidrule(lr){7-7} \cmidrule(l){8-8} \cmidrule(l){9-9}

\multirow[c]{17}{*}{\rotatebox[origin=c]{90}{accuracy}}
& no pre-training & \num{0.454977 +- 0.108472} & \num{0.586118 +- 0.069080} & \num{0.731761 +- 0.062051} & \num{0.717844 +- 0.116304} & \num{0.775220 +- 0.031067} & \num{0.783219 +- 0.007926} & \num{0.815354 +- 0.012915} \\
 \cmidrule(r){2-9}
 & transfer learning (\latvia) & \num{0.454039 +- 0.033425} & \num{0.482701 +- 0.066499} & \num{0.545654 +- 0.019645} & \num{0.648047 +- 0.063566} & \num{0.701114 +- 0.029334} & \num{0.723012 +- 0.017259} & \num{0.764561 +- 0.026406} \\
 & ANIL (\latvia) & \num{0.471451 +- 0.156240} & \num{0.552226 +- 0.070581} & \num{0.659093 +- 0.120492} & \num{0.736615 +- 0.062155} & \num{0.778108 +- 0.038984} & \num{0.790245 +- 0.015672} & \num{0.823785 +- 0.013354} \\
 & MAML (\latvia) & \num{0.502734 +- 0.102884} & \bfseries \color{blue} \num{0.599420 +- 0.054559} & \num{0.658411 +- 0.025379} & \num{0.765738 +- 0.043912} & \bfseries \color{blue} \num{0.807629 +- 0.010940} & \bfseries \color{blue} \num{0.822074 +- 0.017549} & \bfseries \color{blue} \num{0.834017 +- 0.005527} \\
 & FOMAML (\latvia) & \num{0.514064 +- 0.116317} & \num{0.554096 +- 0.035019} & \bfseries \color{blue} \num{0.735859 +- 0.038825} & \bfseries \color{blue} \num{0.770394 +- 0.022638} & \num{0.769263 +- 0.029519} & \num{0.773725 +- 0.013640} & \num{0.811568 +- 0.008962} \\
 & TIML (encoder) (\latvia) & \bfseries \color{blue} \num{0.575766 +- 0.123395} & \num{0.586954 +- 0.110027} & \num{0.664465 +- 0.063590} & \num{0.748826 +- 0.060250} & \num{0.757217 +- 0.042913} & \num{0.777790 +- 0.015709} & \num{0.795879 +- 0.012834} \\
 & TIML (no encoder) (\latvia) & \num{0.497919 +- 0.125793} & \num{0.576846 +- 0.089834} & \num{0.630243 +- 0.070785} & \num{0.726514 +- 0.056205} & \num{0.744347 +- 0.044401} & \num{0.757643 +- 0.029164} & \num{0.808493 +- 0.008607} \\
 \arrayrulecolor{gray} 
 \cmidrule(r){2-9}
 & transfer learning (\latvia + \portugal) & \num{0.443136 +- 0.008620} & \num{0.421033 +- 0.020426} & \color{mediumseagreen} \num{0.569774 +- 0.099714} & \num{0.578540 +- 0.085044} & \num{0.677511 +- 0.019116} & \num{0.701421 +- 0.025753} & \color{mediumseagreen} \num{0.784270 +- 0.024468} \\
 & ANIL (\latvia + \portugal) & \color{mediumseagreen} \num{0.500699 +- 0.063672} & \num{0.505179 +- 0.024998} & \num{0.638815 +- 0.076336} & \color{mediumseagreen} \num{0.751890 +- 0.039606} & \num{0.771059 +- 0.029347} & \num{0.788920 +- 0.017986} & \num{0.815696 +- 0.009634} \\
 & MAML (\latvia + \portugal) & \num{0.446296 +- 0.025481} & \num{0.522023 +- 0.021166} & \color{mediumseagreen} \num{0.680672 +- 0.072473} & \num{0.730214 +- 0.081500} & \num{0.777585 +- 0.015432} & \num{0.781002 +- 0.013573} & \num{0.820357 +- 0.011729} \\
 & FOMAML (\latvia + \portugal) & \num{0.456557 +- 0.015933} & \num{0.513308 +- 0.054713} & \num{0.659673 +- 0.090705} & \num{0.721545 +- 0.064980} & \num{0.760588 +- 0.030767} & \num{0.773299 +- 0.025876} & \color{mediumseagreen} \num{0.816736 +- 0.010180} \\
 & TIML (encoder) (\latvia + \portugal) & \num{0.473583 +- 0.037652} & \num{0.457853 +- 0.079997} & \num{0.616668 +- 0.090940} & \num{0.716588 +- 0.084335} & \color{mediumseagreen} \num{0.762919 +- 0.029049} & \num{0.744233 +- 0.051406} & \color{mediumseagreen} \num{0.800159 +- 0.010360} \\
 & TIML (no encoder) (\latvia + \portugal) & \num{0.495816 +- 0.037494} & \num{0.515514 +- 0.047304} & \num{0.609761 +- 0.026576} & \num{0.707316 +- 0.061282} & \num{0.742027 +- 0.045585} & \num{0.741402 +- 0.028326} & \num{0.800159 +- 0.013542} \\
 \arrayrulecolor{black}
  \cmidrule(r){2-9}
& \gls{gl:presto} & \num{0.497556 +- 0.024676} & \bfseries \num{0.559286 +- 0.087984} & \num{0.556370 +- 0.083461} & \num{0.651941 +- 0.029086} & \num{0.735342 +- 0.045571} & \bfseries \num{0.751265 +- 0.038561} & \num{0.792957 +- 0.007095} \\
 & \gls{gl:prestoxts} & \num{0.487818 +- 0.024219} & \num{0.483963 +- 0.050540} & \num{0.557075 +- 0.104838} & \num{0.541652 +- 0.080392} & \num{0.695327 +- 0.125914} & \num{0.720277 +- 0.020358} & \bfseries \num{0.802797 +- 0.009533} \\
 & \gls{gl:crossprestoxts} & \bfseries \num{0.503581 +- 0.055495} & \num{0.489881 +- 0.020349} & \bfseries \num{0.582792 +- 0.059675} & \bfseries \num{0.681229 +- 0.114246} & \bfseries \num{0.779211 +- 0.020459} & \num{0.700830 +- 0.132960} & \num{0.793241 +- 0.019504} \\
\bottomrule
\end{tabular}

  \end{subtable}
  
  \medskip

    \begin{subtable}{\linewidth}
        \caption{Parent-level \num{4} accuracy for fine-tuning runs on \taskStwo\, where a prediction is considered correct if the predicted class shares the same parent category (\gls{gl:hcat}3 level \num{4}) as the ground truth.
        \label{tab:results_S2_parentacc_lv4}}
        \centering\scriptsize

\robustify\bfseries
\sisetup{detect-all=true,uncertainty-mode=separate,table-align-uncertainty=true,round-mode=uncertainty,round-precision=2}

\begin{tabular}{llSSSSSSS}
\toprule
{} & {benchmark task ($k$-shot)} & {1} & {5} & {10} & {20} & {100} & {200} & {500} \\
{} & {algorithm} & {} & {} & {} & {} & {} & {} & {} \\

\cmidrule(r){1-2} \cmidrule(lr){3-3} \cmidrule(lr){4-4} \cmidrule(lr){5-5} \cmidrule(lr){6-6} \cmidrule(lr){7-7} \cmidrule(l){8-8} \cmidrule(l){9-9}

\multirow[c]{17}{*}{\rotatebox[origin=c]{90}{accuracy}}
& no pre-training & \num{0.242670 +- 0.097836} & \num{0.389557 +- 0.068434} & \num{0.532141 +- 0.054936} & \num{0.566363 +- 0.059259} & \num{0.621784 +- 0.036491} & \num{0.683969 +- 0.012150} & \num{0.725934 +- 0.011307} \\
\cmidrule(r){2-9}
 & transfer learning (\latvia) & \num{0.231255 +- 0.052137} & \num{0.304173 +- 0.053274} & \num{0.355153 +- 0.035923} & \num{0.494656 +- 0.081156} & \num{0.585180 +- 0.036251} & \num{0.617799 +- 0.029685} & \num{0.669365 +- 0.038751} \\
 & ANIL (\latvia) & \num{0.262134 +- 0.082029} & \num{0.373418 +- 0.091136} & \num{0.527719 +- 0.059678} & \num{0.615082 +- 0.066994} & \num{0.669069 +- 0.045391} & \num{0.690984 +- 0.020959} & \num{0.741959 +- 0.012830} \\
 & MAML (\latvia) & \num{0.276624 +- 0.076845} & \num{0.344477 +- 0.078468} & \num{0.500216 +- 0.032863} & \bfseries \color{blue} \num{0.636570 +- 0.050570} & \bfseries \color{blue} \num{0.680791 +- 0.025387} & \bfseries \color{blue} \num{0.713820 +- 0.040164} & \bfseries \color{blue} \num{0.745398 +- 0.010864} \\
 & FOMAML (\latvia) & \num{0.280695 +- 0.072051} & \num{0.354249 +- 0.051316} & \bfseries \color{blue} \num{0.543079 +- 0.047951} & \num{0.589813 +- 0.048465} & \num{0.648917 +- 0.028891} & \num{0.669496 +- 0.014207} & \num{0.714075 +- 0.008610} \\
 & TIML (encoder) (\latvia) & \bfseries \color{blue} \num{0.306793 +- 0.092384} & \num{0.337786 +- 0.092389} & \num{0.422489 +- 0.024540} & \num{0.525155 +- 0.044431} & \num{0.605202 +- 0.041330} & \num{0.607942 +- 0.043521} & \num{0.696385 +- 0.013544} \\
 & TIML (no encoder) (\latvia) & \num{0.288329 +- 0.137554} & \bfseries \color{blue} \num{0.394747 +- 0.112558} & \num{0.465551 +- 0.075178} & \num{0.577705 +- 0.057340} & \num{0.640771 +- 0.038098} & \num{0.661480 +- 0.021521} & \num{0.723768 +- 0.007931} \\
 \arrayrulecolor{gray} 
 \cmidrule(r){2-9}
& transfer learning (\latvia + \portugal) & \num{0.172628 +- 0.014479} & \num{0.226423 +- 0.009252} & \color{mediumseagreen} \num{0.376585 +- 0.091416} & \num{0.411080 +- 0.084929} & \num{0.536502 +- 0.006054} & \num{0.588511 +- 0.027938} & \color{mediumseagreen} \num{0.680820 +- 0.023676} \\
 & ANIL (\latvia + \portugal) & \color{mediumseagreen} \num{0.270360 +- 0.050541} & \num{0.325541 +- 0.025910} & \num{0.431732 +- 0.092723} & \num{0.609516 +- 0.033046} & \num{0.641362 +- 0.036507} & \num{0.690057 +- 0.014815} & \num{0.724183 +- 0.011448} \\
 & MAML (\latvia + \portugal) & \num{0.234904 +- 0.020354} & \num{0.340208 +- 0.022147} & \num{0.478262 +- 0.091128} & \num{0.588375 +- 0.080625} & \num{0.653704 +- 0.019366} & \num{0.678807 +- 0.017941} & \num{0.726178 +- 0.017125} \\
 & FOMAML (\latvia + \portugal) & \num{0.213780 +- 0.026372} & \num{0.322062 +- 0.049191} & \num{0.440146 +- 0.063937} & \num{0.557109 +- 0.058166} & \num{0.637309 +- 0.025832} & \color{mediumseagreen} \num{0.672963 +- 0.020972} & \color{mediumseagreen} \num{0.723955 +- 0.007583} \\
 & TIML (encoder) (\latvia + \portugal) & \num{0.201251 +- 0.038113} & \num{0.254033 +- 0.008581} & \color{mediumseagreen} \num{0.429219 +- 0.064291} & \num{0.524228 +- 0.024593} & \num{0.578353 +- 0.014453} & \num{0.591894 +- 0.041104} & \num{0.687772 +- 0.032920} \\
 & TIML (no encoder) (\latvia + \portugal) & \num{0.271701 +- 0.025109} & \num{0.311665 +- 0.028910} & \num{0.439054 +- 0.021876} & \num{0.544449 +- 0.050587} & \num{0.638491 +- 0.041164} & \num{0.642573 +- 0.020391} & \num{0.715946 +- 0.015661} \\
 \arrayrulecolor{black}
  \cmidrule(r){2-9}
& \gls{gl:presto} & \num{0.198772 +- 0.067700} & \num{0.300784 +- 0.095341} & \num{0.321124 +- 0.049337} & \num{0.398096 +- 0.054611} & \num{0.569041 +- 0.029402} & \num{0.605258 +- 0.057480} & \num{0.652885 +- 0.009127} \\
 & \gls{gl:prestoxts} & \num{0.129430 +- 0.061547} & \num{0.258541 +- 0.105565} & \num{0.318629 +- 0.083828} & \num{0.463419 +- 0.032234} & \num{0.584998 +- 0.065658} & \bfseries \num{0.606702 +- 0.026595} & \bfseries \num{0.713058 +- 0.018574} \\
 & \gls{gl:crossprestoxts} & \bfseries \num{0.214524 +- 0.077589} & \bfseries \num{0.366011 +- 0.127214} & \bfseries \num{0.401830 +- 0.044130} & \bfseries \num{0.546552 +- 0.047029} & \bfseries \num{0.595105 +- 0.040873} & \num{0.564931 +- 0.058244} & \num{0.698226 +- 0.018134} \\

\bottomrule
\end{tabular}

  \end{subtable}%
\end{table*}

\begin{table*}[ht!]
    \vspace{12pt}
    \caption{%
    Parent-level classification results for the \taskalldata\ benchmark.
    We report \emph{parent-class classification accuracy} on the test set for each algorithm and few-shot task.
    Metrics are shown as mean $\pm$ standard deviation over five runs, \cf \cref{sec:exp_finetuning}.
    The best result per few-shot scenario is marked in \textbf{bold}.
    }
    \label{tab:results_S1S2ERA5_parentacc}
    \begin{subtable}{\linewidth}
        \caption{Parent-level \num{3} accuracy for fine-tuning runs on \taskalldata\, where a prediction is considered correct if the predicted class shares the same parent category (\gls{gl:hcat}3 level \num{3}) as the ground truth.
        \label{tab:results_S1S2ERA5_parentacc_lv3}}
        \centering\scriptsize

\robustify\bfseries
\sisetup{detect-all=true,uncertainty-mode=separate,table-align-uncertainty=true,round-mode=uncertainty,round-precision=2}
\begin{tabular}{llSSSSSSS}
\toprule
{} & {benchmark task ($k$-shot)} & {1} & {5} & {10} & {20} & {100} & {200} & {500} \\
{} & {algorithm} & {} & {} & {} & {} & {} & {} & {} \\

\cmidrule(r){1-2} \cmidrule(lr){3-3} \cmidrule(lr){4-4} \cmidrule(lr){5-5} \cmidrule(lr){6-6} \cmidrule(l){7-7} \cmidrule(l){8-8} \cmidrule(l){9-9}

\multirow[c]{4}{*}{\rotatebox[origin=c]{90}{accuracy}}
& No pretraining & \num{0.484668 +- 0.024219} & \num{0.518578 +- 0.059325} & \bfseries \num{0.530777 +- 0.075970} & \num{0.530373 +- 0.087093} & \num{0.681337 +- 0.107369} & \num{0.704161 +- 0.098623} & \num{0.791433 +- 0.019044} \\
 \cmidrule(r){2-9}
 & \gls{gl:presto} & \num{0.472656 +- 0.019195} & \bfseries \num{0.562259 +- 0.053310} & \num{0.519300 +- 0.110707} & \bfseries \num{0.673424 +- 0.049645} & \bfseries \num{0.725416 +- 0.044971} & \bfseries \num{0.741402 +- 0.030773} & \bfseries \num{0.807009 +- 0.017133} \\
 & \gls{gl:prestoxts} & \num{0.456796 +- 0.050009} & \num{0.447274 +- 0.043906} & \num{0.454511 +- 0.129027} & \num{0.576010 +- 0.136424} & \num{0.669985 +- 0.113064} & \num{0.686095 +- 0.120069} & \num{0.791018 +- 0.017057} \\
 & \gls{gl:crossprestoxts} & \bfseries \num{0.497823 +- 0.010916} & \num{0.464470 +- 0.040092} & \num{0.492650 +- 0.026989} & \num{0.616577 +- 0.084820} & \num{0.659309 +- 0.118193} & \num{0.685521 +- 0.028284} & \num{0.794628 +- 0.012252} \\
\bottomrule
\end{tabular}

  \end{subtable}
  
  \medskip
  
    \begin{subtable}{\linewidth}
        \caption{Parent-level \num{4} accuracy for fine-tuning runs on \taskalldata\, where a prediction is considered correct if the predicted class shares the same parent category (\gls{gl:hcat}3 level \num{4}) as the ground truth.
        \label{tab:results_S1S2ERA5_parentacc_lv4}}
        \centering\scriptsize

\robustify\bfseries
\sisetup{detect-all=true,uncertainty-mode=separate,table-align-uncertainty=true,round-mode=uncertainty,round-precision=2}
\begin{tabular}{llSSSSSSS}
\toprule
{} & {benchmark task ($k$-shot)} & {1} & {5} & {10} & {20} & {100} & {200} & {500} \\
{} & {algorithm} & {} & {} & {} & {} & {} & {} & {} \\

\cmidrule(r){1-2} \cmidrule(lr){3-3} \cmidrule(lr){4-4} \cmidrule(lr){5-5} \cmidrule(lr){6-6} \cmidrule(l){7-7} \cmidrule(l){8-8} \cmidrule(l){9-9}

\multirow[c]{4}{*}{\rotatebox[origin=c]{90}{accuracy}}
& No pretraining & \bfseries \num{0.239742 +- 0.165766} & \bfseries \num{0.448690 +- 0.048373} & \bfseries \num{0.299017 +- 0.100494} & \num{0.464897 +- 0.020778} & \num{0.482338 +- 0.040419} & \num{0.522699 +- 0.029221} & \num{0.626411 +- 0.031621} \\
 \cmidrule(r){2-9}
 & \gls{gl:presto} & \num{0.172981 +- 0.045281} & \num{0.308783 +- 0.077392} & \num{0.320221 +- 0.038572} & \num{0.447877 +- 0.030291} & \bfseries \num{0.593139 +- 0.033741} & \bfseries \num{0.627389 +- 0.039034} & \num{0.706151 +- 0.017718} \\
 & \gls{gl:prestoxts} & \num{0.218970 +- 0.165207} & \num{0.403559 +- 0.073410} & \num{0.321426 +- 0.041940} & \bfseries \num{0.483628 +- 0.018908} & \num{0.558013 +- 0.050112} & \num{0.585771 +- 0.063602} & \bfseries  \num{0.706628 +- 0.021903} \\
 & \gls{gl:crossprestoxts} & \num{0.202308 +- 0.168332} & \num{0.188574 +- 0.045065} & \num{0.218322 +- 0.019904} & \num{0.425991 +- 0.039125} & \num{0.538423 +- 0.047188} & \num{0.577909 +- 0.048975} & \num{0.701046 +- 0.015808} \\
\bottomrule
\end{tabular}

  \end{subtable}%
\end{table*}

\subsection{Evaluation of meta-learning hyperparameters}

\begin{table*}  
    \centering
     \caption{%
     Detailed analysis of the \num{20}-shot benchmark task.
     The metric \emph{classification accuracy} was evaluated for six different $n$-way $k$-shot $s$-step meta-learning variants per algorithm.
     We report the mean $\pm$ standard deviation over five repeated experimental runs in all cases.
     The best result per algorithm is marked in \textbf{bold}.
     }
    \scriptsize

\robustify\bfseries
\sisetup{detect-all=true,uncertainty-mode=separate,table-align-uncertainty=true,round-mode=uncertainty,round-precision=2}

\begin{tabular}{llSSSSSS}
\toprule
{} & {$n$-way} & {4} & {4} & {4} & {4} & {10} & {10} \\
{} & {$k$-shot} & {1} & {1} & {1} & {10} & {1} & {10} \\
{} & {$s$-step} & {1} & {4} & {10} & {1} & {1} & {1} \\
{} & {algorithm} & {} & {} & {} & {} & {} & {} \\

\cmidrule(r){1-2} \cmidrule(lr){3-3} \cmidrule(lr){4-4} \cmidrule(lr){5-5} \cmidrule(lr){6-6} \cmidrule(lr){7-7} \cmidrule(l){8-8} 

\multirow[c]{10}{*}{\rotatebox[origin=c]{90}{accuracy}}
    & \gls{gl:anil} (\latvia) & \num{0.404804 +- 0.061107} & \bfseries \num{0.472952 +- 0.067690} & \num{0.439429 +- 0.065195} & \num{0.369041 +- 0.083332} & \num{0.403138 +- 0.093792} & \num{0.358854 +- 0.071813} \\
    & \gls{gl:maml} (\latvia) & \num{0.464084 +- 0.060093} & \bfseries \num{0.465113 +- 0.067180} & \num{0.418896 +- 0.040983} & \num{0.417077 +- 0.084073} & \num{0.361435 +- 0.077880} & \num{0.462879 +- 0.058459} \\
    & \gls{gl:fomaml} (\latvia) & \bfseries \num{0.448530 +- 0.030289} & \num{0.447541 +- 0.059859} & \num{0.403144 +- 0.051727} & \num{0.398960 +- 0.045439} & \num{0.435666 +- 0.047671} & \num{0.364561 +- 0.102510} \\
    & \gls{gl:timl} (encoder) (\latvia) & \bfseries \num{0.414115 +- 0.054506} & \num{0.366722 +- 0.061878} & \num{0.372270 +- 0.074187} & \num{0.327218 +- 0.089545} & \num{0.375948 +- 0.085217} & \num{0.296373 +- 0.061411} \\
    & \gls{gl:timl} (no encoder) (\latvia) & \num{0.385083 +- 0.063266} & \num{0.390870 +- 0.072602} & \bfseries \num{0.432636 +- 0.081063} & \num{0.379137 +- 0.077030} & \num{0.371349 +- 0.050179} & \num{0.339577 +- 0.063553} \\
     \arrayrulecolor{gray} 
     \cmidrule(r){2-8}
     \arrayrulecolor{black}
    & \gls{gl:anil} (\latvia + \portugal) & \num{0.355085 +- 0.060242} & \num{0.395850 +- 0.067661} & \num{0.376744 +- 0.075071} & \bfseries \num{0.462174 +- 0.069365} & \num{0.374419 +- 0.084040} & \num{0.330862 +- 0.049867} \\
    & \gls{gl:maml} (\latvia + \portugal) & \num{0.379001 +- 0.072980} & \bfseries \num{0.443670 +- 0.107331} & \num{0.347098 +- 0.075535} & \num{0.361639 +- 0.080775} & \num{0.388392 +- 0.063147} & \num{0.356989 +- 0.037022} \\
    & \gls{gl:fomaml} (\latvia + \portugal) & \num{0.408249 +- 0.051342} & \bfseries \num{0.410073 +- 0.073538} & \num{0.332397 +- 0.091658} & \num{0.391871 +- 0.068699} & \num{0.377875 +- 0.083426} & \num{0.341913 +- 0.051141} \\
    & \gls{gl:timl} (encoder) (\latvia + \portugal) & \num{0.342232 +- 0.093165} & \num{0.318873 +- 0.040374} & \num{0.351896 +- 0.078424} & \num{0.374567 +- 0.046718} & \bfseries \num{0.394429 +- 0.062370} & \num{0.366426 +- 0.060150} \\
    & \gls{gl:timl} (no encoder) (\latvia + \portugal) & \num{0.361492 +- 0.030127} & \bfseries \num{0.423853 +- 0.047674} & \num{0.417594 +- 0.093636} & \num{0.313939 +- 0.051693} & \num{0.313729 +- 0.059432} & \num{0.355767 +- 0.052903} \\
\bottomrule
\end{tabular}

    \label{tab:lv_lv_pt_ee_meta}
\end{table*}
\cref{tab:lv_lv_pt_ee_meta} shows the classification accuracy of the \num{20}-shot benchmarking tasks for the six $n$-way $k$-shot $s$-step settings of the meta-learning algorithms.
Although there was no single setting that performed best in all algorithms and benchmark tasks, in most cases the best prediction accuracies could be achieved for meta-learning in \num{4}-way \num{1}-shot \num{4}-step.
This shows that all meta-learning algorithms could take advantage of taking more than one gradient step during the inner optimization for the task adaptation, but a small number of steps is sufficient.
Furthermore, allowing for a larger number of classes ($n=10$) and thus potentially a higher diversity between classes in tasks during pre-training did not improve downstream fine-tuning performance for any algorithm.
Similarly, presenting the models with more samples per class ($k=10$) and thus potentially a higher intra-class diversity in the tasks during pre-training also did not improve their fine-tuning performance in all but one case.

\subsection{Effect of masking design and model capacity on PRESTO}
\label{sec:appendix_modelsize_masking}
In addition to using a mixture of all masking strategies defined in \cref{sec:Presto}, we also experimented with restricting the model to deterministic masking only, \ie masking entire channel groups and contiguous time steps.
This design choice is motivated by real-world remote sensing deployment scenarios, where entire input modalities may be missing at test time, \eg due to persistent cloud cover in Sentinel-2, and temporal gaps often occur in contiguous blocks.
By pre-training the model under these deterministic masking conditions, we aim to encourage the encoder to learn more robust representations.

Moreover, along with the model defined in \cref{sec:appendix_networkarchitecture}, we pre-train a larger variant of \gls{gl:prestoxts} and \gls{gl:crossprestoxts}, with a size that matches the original \gls{gl:presto} model (\cf \cref{sec:appendix_presto}) but uses an increased embedding size of $d_{\text{emb}}=256$.
This choice was made based on the fact that bigger models have the capacity to often learn richer representations during self-supervised pre-training.
\Cref{tab:comparison_additional_S2,tab:comparison_additional_S1S2ERA5} show the results for both the main models and the bigger variant.
Models using the larger architecture are marked with the postfix \emph{L}, and those using deterministic masking with the postfix \emph{D}.

Across both fine-tuning tasks, increasing the model size and using deterministic masking strategies significantly influence performance.
For \taskStwo\ low-shot regimes (\cref{tab:results_additional_S2}), \gls{gl:prestoxts} (D, L) and \gls{gl:crossprestoxts} (D, L) generally improve overall accuracy compared to \cref{tab:comparison_results_eurocropsml_S2}.
In particular, the increased model size seems to boost performance here.
However, especially \gls{gl:crossprestoxts} tends to focus mostly on the majority class when trained with a larger model, as reflected in \cref{tab:results_additional_S2_nomeadow}.
As the number of training samples increases, the original \gls{gl:prestoxts} model frequently outperforms the larger or deterministic variants.

For the multi-modal \taskalldata\ task, \gls{gl:prestoxts} consistently favors the original model (\cref{sec:results}) from five shots onward, while its minority-class accuracy improves most with the smaller model for higher shot regimes.
In contrast, \gls{gl:crossprestoxts} retains a longer advantage from the larger model size for overall accuracy, suggesting its pre-training architecture scales better with multi-modal complexity.
However, for minority-class performance, the smaller \gls{gl:crossprestoxts} variant again performs best, which is consistent with trends observed in \taskStwo.

All in all, both strategies appear to help increase accuracy in low-shot regimes but often increase the model's tendency of focusing on the majority class instead of improving per-class predictions.

\subsection{Robustness evaluation under class imbalance}
\label{sec:appendix_cohenskappa}
While the main evaluation in \cref{sec:results} focuses on the overall classification accuracy and a restricted accuracy excluding the majority class that addresses the high class imbalance in the data, we report \emph{Cohen’s kappa coefficient} $\kappa$  as an additional evaluation metric.
The kappa score offers a complementary perspective to accuracy by accounting for chance agreements between predicted and ground truth labels, providing a robust metric in scenarios with skewed class distributions.
Divergence between $a_{\text{MCA}}$ (\cf \cref{tab:best_results_S2_nomeadow,tab:best_results_S1S2ERA5_nomeadow}) and $\kappa$ reflects their differing sensitivities: minority-class accuracy evaluates performance exclusively on underrepresented classes, while $\kappa$ captures overall class-wise agreement, penalizing both change alignment and skewed distributions.
Hence, a model might achieve strong performance on underrepresented classes---resulting in high minority-class accuracy---but perform poorly on the majority class.
Since Cohen's kappa evaluates agreement across the full label distribution, the model's weak performance on the majority class would lead to substantially lower kappa scores, despite its strengths on rare classes.

As in \cref{sec:results}, we report the best-performing configuration for each algorithm based on the metric on the validation set in \cref{tab:comparison_results_eurocropsml_S2_CK,tab:comparison_results_eurocropsml_S1S2ERA5_CK}, presenting the mean and standard deviation over five repeated experimental runs.

\subsubsection{\taskStwo}
As evaluated in \cref{sec:discussion_supervised}, the \gls{gl:maml} algorithms show superior behavior, showing the highest Cohen's kappa scores, with \gls{gl:anil} achieving the best overall performance for \num{500} shots with $\kappa = 0.556 \pm 0.01$.
In contrast, despite demonstrating relatively strong accuracy in low ($k\leq 20$) few-shot tasks, the randomly initialized baseline model achieves a rather low Cohen's kappa score. 
While \gls{gl:timl} (without encoder) shows competitive results for \lvee, in particular \gls{gl:timl} (with encoder) seems to struggle massively with the high-class imbalance when being trained on \lvptee, indicating that the model is unable to discern the differences between various crop classes and instead demonstrating a proclivity for the majority class.
This further confirms the results from \cref{tab:best_results_S2_nomeadow,tab:overlap_results_lvpt}, which suggested that both the baseline as well as the \gls{gl:timl} (with encoder) may have mainly learned to predict the dominant crop class \meadow, instead of effectively learning the distinctions between the various crop classes.
Although pre-training on additional Portugal data usually decreases overall accuracy (\cf \cref{tab:best_results_S2}), it often leads to increased $\kappa$ scores and minority-class performance (\cf \cref{tab:best_results_S2_nomeadow}) for meta-learning algorithms (\cf \cref{tab:comparison_results_eurocropsml_S2_CK}).
In particular \gls{gl:fomaml} shows improved Cohen's kappa scores (highlighted in {\color{mediumseagreen}green}).

Nevertheless, most supervised methods outperform the \gls{gl:ssl} models in all scenarios (\cref{tab:comparison_results_eurocropsml_S2_CK}), although the performance gap narrows to approximately 7\%\ between the best meta-learning algorithms and \gls{gl:prestoxts} at \num{500} samples.
\gls{gl:presto} achieves higher scores than both \gls{gl:prestoxts} and \gls{gl:crossprestoxts} in the \num{1}- and \num{5}-shot settings.
Notably, among the \gls{gl:ssl} methods, \gls{gl:crossprestoxts} attains the highest scores with  \numlist[list-final-separator = {, and }]{10;20;100} samples with  $\kappa_{10}=0.122 \pm 0.067$, $\kappa_{20}=0.2486 \pm 0.0097$, and $\kappa_{100}=0.344 \pm 0.038$.
\gls{gl:prestoxts}, while initially underperforming, exhibits steady improvement with more data and outperforms the other \gls{gl:ssl} methods in the \num{500}-shot regime.
Moreover, the results on the (D, L) variants confirm the assumption from \cref{sec:appendix_modelsize_masking} that, while improving overall accuracy, the models mainly focus on the majority class \meadow instead of learning diverse class patterns.
Training the baseline model on all \taskStwo\ data, achieves $\kappa = \num{0.705(0.002)}$ on the test set.

\subsubsection{\taskalldata}
A similar trend for \gls{gl:ssl} is observed in \cref{tab:comparison_results_eurocropsml_S1S2ERA5_CK}, where the smaller \gls{gl:presto} consistently achieves the highest $\kappa$ scores in the \num{1}- to \num{200}-shot range and is only exceeded by \gls{gl:prestoxts} in the \num{500}-shot setting with an increase of 4\%.
\gls{gl:crossprestoxts} shows stable improvement with increasing sample sizes and closely follows the performance of \gls{gl:prestoxts} on \num{500} samples. 
The baseline model trained on all data of the \taskalldata\ task yields $\kappa = \num{0.673(0.016)}$ on the test set.

\begin{table*}[p]
     \caption{%
     Detailed analysis of the \num{20}-shot \taskStwo\ benchmark task.
        The metric \emph{classification accuracy} was evaluated for three different learning rate tuning variants per algorithm. We report the mean $\pm$ standard deviation over five repeated experimental runs in all cases.
        The best result per algorithm is marked in \textbf{bold}.\\
    }
    \vspace{7pt}
    \label{tab:comparison_learningrate_settings_S2}
    \centering\scriptsize

\robustify\bfseries
\sisetup{detect-all=true,uncertainty-mode=separate,table-align-uncertainty=true,round-mode=uncertainty,round-precision=2}
\begin{tabular}{llSSS}
\toprule
{} & {lr tuning} & {only head} & {different head \& backbone} & {same head \& backbone} \\
{} & {algorithm} & {} & {} & {} \\

\cmidrule(r){1-2} \cmidrule(lr){3-3} \cmidrule(lr){4-4} \cmidrule(l){5-5}

\multirow[c]{17}{*}{\rotatebox[origin=c]{90}{accuracy}}
    & no pre-training\textsuperscript{\textdagger} & \num{0.322256 +- 0.178078} & \bfseries \num{0.486112 +- 0.017052} & \num{0.462560 +- 0.035096} \\
    \cmidrule(r){2-5}
    & transfer learning (\latvia) & \num{0.260673 +- 0.081695} & \num{0.325439 +- 0.087240} & \bfseries \num{0.353084 +- 0.085472} \\
    & \gls{gl:anil} (\latvia) & \num{0.372634 +- 0.050698} & \bfseries \num{0.477676 +- 0.064123} & \num{0.445017 +- 0.055907} \\
    & \gls{gl:maml} (\latvia) & \num{0.386021 +- 0.055835} & \bfseries \num{0.492854 +- 0.058410} & \num{0.463453 +- 0.071543} \\
    & \gls{gl:fomaml} (\latvia) & \num{0.420238 +- 0.040899} & \bfseries \num{0.465801 +- 0.022434} & \num{0.450702 +- 0.065603} \\
    & \gls{gl:timl} (encoder) (\latvia) & \num{0.344523 +- 0.083447} & \num{0.401575 +- 0.074192} & \bfseries \num{0.420687 +- 0.063731} \\
    & \gls{gl:timl} (no encoder) (\latvia) & \num{0.378006 +- 0.080300} & \bfseries \num{0.448013 +- 0.062472} & \num{0.423188 +- 0.067041} \\
    \arrayrulecolor{gray} 
    \cmidrule(r){2-5}
    \arrayrulecolor{black}
    & transfer learning  (\latvia + \portugal) & \num{0.171184 +- 0.020892} & \num{0.256222 +- 0.070828} & \bfseries \num{0.260940 +- 0.094762} \\
    & \gls{gl:anil} (\latvia + \portugal) & \num{0.364777 +- 0.059534} & \bfseries \num{0.470206 +- 0.055846} & \num{0.406299 +- 0.084484} \\
    & \gls{gl:maml} on \latvia + \portugal & \num{0.386504 +- 0.073418} & \num{0.411477 +- 0.101939} & \bfseries \num{0.443897 +- 0.096873} \\
    & \gls{gl:fomaml} (\latvia + \portugal) & \num{0.376596 +- 0.095893} & \bfseries \num{0.439952 +- 0.058408} & \num{0.420209 +- 0.058119} \\
    & \gls{gl:timl} (encoder) (\latvia + \portugal) & \num{0.343096 +- 0.024410} & \num{0.397266 +- 0.038999} & \bfseries \num{0.418345 +- 0.048999} \\
    & \gls{gl:timl} (no encoder) (\latvia + \portugal) & \num{0.352049 +- 0.041252} & \bfseries \num{0.438901 +- 0.058611} & \num{0.438508 +- 0.067284} \\
    \cmidrule(r){2-5}
    & \gls{gl:presto} & \num{0.279012 +- 0.046930} & \bfseries \num{0.294389 +- 0.072960} & \num{0.286197 +- 0.051870} \\
    & \gls{gl:prestoxts} & \bfseries \num{0.432608 +- 0.022672} & \num{0.421875 +- 0.062326} & \num{0.337553 +- 0.031338} \\
    & \gls{gl:crossprestoxts} & \num{0.434438 +- 0.040642} & \bfseries \num{0.473549 +- 0.009639} & \num{0.366824 +- 0.112704} \\
\bottomrule
\end{tabular}

\end{table*}

\begin{table*}[p]
    \caption{%
    Detailed analysis of the \num{20}-shot \taskalldata\ benchmark task.
    The \emph{classification accuracy} was evaluated for three different learning rate tuning variants per algorithm.
    We report \emph{classification accuracy} on the test set for each algorithm and all three learning rate tuning variants.
    Metrics are shown as mean $\pm$ standard deviation over five runs, \cf \cref{sec:exp_finetuning}.
    The best result per algorithm is marked in \textbf{bold}.
    }
    \vspace{7pt}
    \label{tab:comparison_learningrate_settings_S1S2ERA5}
    \centering\scriptsize

\robustify\bfseries
\sisetup{detect-all=true,uncertainty-mode=separate,table-align-uncertainty=true,round-mode=uncertainty,round-precision=2}
\begin{tabular}{llSSS}
\toprule
{} & {lr tuning} & {only head} & {different head \& backbone} & {same head \& backbone} \\
{} & {algorithm} & {} & {} & {} \\

\cmidrule(r){1-2} \cmidrule(lr){3-3} \cmidrule(lr){4-4} \cmidrule(l){5-5}

\multirow[c]{4}{*}{\rotatebox[origin=c]{90}{accuracy}}
 & no pre-training & \num{0.312398 +- 0.137417} & \num{0.376823 +- 0.077296} & \bfseries \num{0.428401 +- 0.055142} \\
 \cmidrule(r){2-5}
 & \gls{gl:presto} & \num{0.294912 +- 0.059231} & \bfseries \num{0.357041 +- 0.062558} & \num{0.287516 +- 0.040547} \\
 & \gls{gl:prestoxts} & \num{0.326792 +- 0.133653} & \bfseries \num{0.430629 +- 0.062123} & \num{0.371588 +- 0.081447} \\
 & \gls{gl:crossprestoxts} & \num{0.177722 +- 0.088117} & \bfseries \num{0.345535 +- 0.095933} & \num{0.296953 +- 0.087004} \\
\bottomrule
\end{tabular}

\end{table*}

\begin{table*}[p]
    \caption{%
    Additional results for the \taskStwo\ benchmark.
    We report \emph{classification accuracy} on the test set for each algorithm and few-shot task.
    Metrics are shown as mean $\pm$ standard deviation over five runs, \cf \cref{sec:exp_finetuning}.
    The best result per task is marked in \textbf{bold}.
    }
    \label{tab:comparison_additional_S2}
    \begin{subtable}{\linewidth}
        \caption{Additional fine-tuning results for \taskStwo.
        \label{tab:results_additional_S2}}
        \centering\scriptsize        

\robustify\bfseries
\sisetup{detect-all=true,uncertainty-mode=separate,table-align-uncertainty=true,round-mode=uncertainty,round-precision=2}

\begin{tabular}{llSSSSSSS}
\toprule
{} & {benchmark task ($k$-shot)} & {1} & {5} & {10} & {20} & {100} & {200} & {500} \\
{} & {algorithm} & {} & {} & {} & {} & {} & {} & {} \\

\cmidrule(r){1-2} \cmidrule(lr){3-3} \cmidrule(lr){4-4} \cmidrule(lr){5-5} \cmidrule(lr){6-6} \cmidrule(lr){7-7} \cmidrule(lr){8-8} \cmidrule(l){9-9}

\multirow[c]{4}{*}{\rotatebox[origin=c]{90}{accuracy}}
  & \gls{gl:prestoxts} (D) & \num{0.064169 +- 0.059714} & \num{0.223756 +- 0.139191} & \num{0.265039 +- 0.137030} & \num{0.403240 +- 0.044657} & \num{0.449724 +- 0.041999} & \num{0.452629 +- 0.022681} & \num{0.599693 +- 0.024249} \\
 & \gls{gl:prestoxts} (D, L) & \bfseries \num{0.154744 +- 0.180960} & \num{0.281786 +- 0.184490} & \num{0.312364 +- 0.142282} & \num{0.463891 +- 0.013608} & \num{0.469860 +- 0.021699} & \bfseries \num{0.500466 +- 0.022864} & \bfseries \num{0.606691 +- 0.014175} \\
 & \gls{gl:crossprestoxts} (D) & \num{0.064647 +- 0.079846} & \bfseries \num{0.380388 +- 0.106742} & \num{0.322853 +- 0.127768} & \num{0.461577 +- 0.017042} & \num{0.450367 +- 0.046835} & \num{0.494884 +- 0.029282} & \num{0.599284 +- 0.009128} \\
 & \gls{gl:crossprestoxts} (D, L) & \num{0.150549 +- 0.183381} & \num{0.200654 +- 0.166793} & \bfseries \num{0.376528 +- 0.126588} & \bfseries \num{0.474970 +- 0.008915} & \bfseries \num{0.483543 +- 0.018930} & \num{0.495287 +- 0.026233} & \num{0.589239 +- 0.025756} \\
\bottomrule
\end{tabular}

    \end{subtable}

    \medskip

    \begin{subtable}{\linewidth}
        \caption{Additional fine-tuning results for \taskStwo\ excluding the majority class \meadow. 
        \label{tab:results_additional_S2_nomeadow}}
        \centering\scriptsize

\robustify\bfseries
\sisetup{detect-all=true,uncertainty-mode=separate,table-align-uncertainty=true,round-mode=uncertainty,round-precision=2}

\begin{tabular}{llSSSSSSS}
\toprule
{} & {benchmark task ($k$-shot)} & {1} & {5} & {10} & {20} & {100} & {200} & {500} \\
{} & {algorithm} & {} & {} & {} & {} & {} & {} & {} \\

\cmidrule(r){1-2} \cmidrule(lr){3-3} \cmidrule(lr){4-4} \cmidrule(lr){5-5} \cmidrule(lr){6-6} \cmidrule(lr){7-7} \cmidrule(lr){8-8} \cmidrule(l){9-9}

\multirow[c]{4}{*}{\rotatebox[origin=c]{90}{accuracy}}
 & \gls{gl:prestoxts} (D) & \num{0.060679 +- 0.049118} & \num{0.023478 +- 0.017257} & \num{0.017309 +- 0.023342} & \bfseries \num{0.048711 +- 0.059155} & \num{0.132291 +- 0.127223} & \num{0.422257 +- 0.035412} & \num{0.501303 +- 0.023213} \\
 & \gls{gl:prestoxts} (D, L) & \bfseries \num{0.113227 +- 0.063296} & \bfseries \num{0.039501 +- 0.062546} & \bfseries \num{0.036024 +- 0.036161} & \num{0.022148 +- 0.035181} & \num{0.200349 +- 0.199174} & \num{0.220982 +- 0.219052} & \bfseries \num{0.509575 +- 0.012360} \\
 & \gls{gl:crossprestoxts} (D) & \num{0.045953 +- 0.038889} & \num{0.008545 +- 0.007606} & \num{0.015957 +- 0.012366} & \num{0.012360 +- 0.010038} & \num{0.109804 +- 0.089149} & \bfseries \num{0.437332 +- 0.030157} & \num{0.507624 +- 0.007572} \\
 & \gls{gl:crossprestoxts} (D, L) & \num{0.105183 +- 0.059667} & \num{0.033778 +- 0.059509} & \num{0.018257 +- 0.020615} & \num{0.030901 +- 0.065944} & \bfseries \num{0.215369 +- 0.176423} & \num{0.213352 +- 0.173032} & \num{0.492801 +- 0.020247} \\
\bottomrule
\end{tabular}

    \end{subtable}
\end{table*}

\begin{table*}[p]
    \caption{%
    Additional results for the \taskalldata\ benchmark.
    We report \emph{classification accuracy} on the test set for each algorithm and few-shot task.
    Metrics are shown as mean $\pm$ standard deviation over five runs, \cf \cref{sec:exp_finetuning}.
    The best result per task is marked in \textbf{bold}.
    }
    \label{tab:comparison_additional_S1S2ERA5}
    \begin{subtable}{\linewidth}
        \caption{Additional fine-tuning results for \taskalldata.
        \label{tab:results_additional_S1S2ERA5}}
        \centering\scriptsize

\robustify\bfseries
\sisetup{detect-all=true,uncertainty-mode=separate,table-align-uncertainty=true,round-mode=uncertainty,round-precision=2}
\begin{tabular}{llSSSSSSS}
\toprule
{} & {benchmark task ($k$-shot)} & {1} & {5} & {10} & {20} & {100} & {200} & {500} \\
{} & {algorithm} & {} & {} & {} & {} & {} & {} & {} \\

\cmidrule(r){1-2} \cmidrule(lr){3-3} \cmidrule(lr){4-4} \cmidrule(lr){5-5} \cmidrule(lr){6-6} \cmidrule(lr){7-7} \cmidrule(lr){8-8} \cmidrule(l){9-9}

\multirow[c]{4}{*}{\rotatebox[origin=c]{90}{accuracy}}
 & \gls{gl:prestoxts} (D) & \bfseries \num{0.330618 +- 0.085149} & \num{0.243755 +- 0.166418} & \num{0.145853 +- 0.105767} & \num{0.302280 +- 0.082548} & \num{0.415525 +- 0.035817} & \num{0.390899 +- 0.122825} & \num{0.539861 +- 0.113371} \\
 & \gls{gl:prestoxts} (D, L) & \num{0.253345 +- 0.170962} & \num{0.298545 +- 0.177430} & \num{0.202541 +- 0.118951} & \num{0.387459 +- 0.079811} & \num{0.425809 +- 0.040237} & \num{0.442408 +- 0.096116} & \num{0.562830 +- 0.063639} \\
 & \gls{gl:crossprestoxts} (D) & \num{0.265555 +- 0.045696} & \num{0.330361 +- 0.132793} & \num{0.278935 +- 0.086076} & \bfseries \num{0.443515 +- 0.025259} & \num{0.450642 +- 0.038166} & \num{0.458068 +- 0.031522} & \bfseries \num{0.593504 +- 0.039253} \\
 & \gls{gl:crossprestoxts} (D, L) & \num{0.276300 +- 0.176927} & \bfseries \num{0.333637 +- 0.102803} & \bfseries \num{0.303695 +- 0.173145} & \num{0.438366 +- 0.056369} & \bfseries \num{0.453664 +- 0.039403} & \bfseries \num{0.475090 +- 0.025688} & \num{0.592223 +- 0.024956} \\
\bottomrule
\end{tabular}

    \end{subtable}

    \medskip

    \begin{subtable}{\linewidth}
        \caption{Additional fine-tuning results for \taskalldata\ excluding the majority class \meadow. 
        \label{tab:results_additional_S1S2ERA5_nomeadow}}
        \centering\scriptsize

\robustify\bfseries
\sisetup{detect-all=true,uncertainty-mode=separate,table-align-uncertainty=true,round-mode=uncertainty,round-precision=2}
\begin{tabular}{llSSSSSSS}
\toprule
{} & {benchmark task ($k$-shot)} & {1} & {5} & {10} & {20} & {100} & {200} & {500} \\
{} & {algorithm} & {} & {} & {} & {} & {} & {} & {} \\

\cmidrule(r){1-2} \cmidrule(lr){3-3} \cmidrule(lr){4-4} \cmidrule(lr){5-5} \cmidrule(lr){6-6} \cmidrule(lr){7-7} \cmidrule(lr){8-8} \cmidrule(l){9-9}

\multirow[c]{4}{*}{\rotatebox[origin=c]{90}{accuracy}}
 & \gls{gl:prestoxts} (D) & \num{0.001907 +- 0.003718} & \num{0.013450 +- 0.025569} & \bfseries \num{0.072026 +- 0.036940} & \num{0.033986 +- 0.030578} & \num{0.114197 +- 0.091644} & \num{0.285443 +- 0.157406} & \num{0.418475 +- 0.181797} \\
 & \gls{gl:prestoxts} (D, L) & \bfseries \num{0.036100 +- 0.061134} & \bfseries \num{0.029811 +- 0.062543} & \num{0.057137 +- 0.058185} & \bfseries \num{0.038280 +- 0.057526} & \bfseries \num{0.193318 +- 0.133467} & \num{0.248842 +- 0.157117} & \num{0.389899 +- 0.260074} \\
 & \gls{gl:crossprestoxts} (D) & \num{0.002848 +- 0.003068} & \num{0.002684 +- 0.001745} & \num{0.020287 +- 0.014755} & \num{0.004265 +- 0.004408} & \num{0.013270 +- 0.009989} & \bfseries \num{0.385443 +- 0.061956} & \bfseries \num{0.463404 +- 0.095870} \\
 & \gls{gl:crossprestoxts} (D, L) & \num{0.027293 +- 0.060420} & \num{0.010398 +- 0.013820} & \num{0.009287 +- 0.013555} & \num{0.029963 +- 0.053245} & \num{0.058990 +- 0.078947} & \num{0.314960 +- 0.182034} & \num{0.458924 +- 0.035399} \\
\bottomrule
\end{tabular}

    \end{subtable}
\end{table*}

\begin{table*}[p]
    \caption{%
    Fine-tuning results for the \taskStwo\ benchmark.
    We report \emph{Cohen's kappa} on the test set for each algorithm and few-shot task.
    Metrics are shown as mean $\pm$ standard deviation over five runs, \cf \cref{sec:exp_finetuning}.
    The best result per few-shot scenario is marked in {\color{blue}\textbf{blue bold}}.
    \textbf{Black bold} values highlight the best result for supervised or \gls{gl:ssl} pre-training.
    Additionally, cases in which the \lvptee was an improvement over the corresponding \lvee result are highlighted in \textcolor{mediumseagreen}{green color}.
    }
    \vspace{7pt}
    \label{tab:comparison_results_eurocropsml_S2_CK}
    \centering\scriptsize

\robustify\bfseries
\sisetup{detect-all=true,uncertainty-mode=separate,table-align-uncertainty=true,round-mode=uncertainty,round-precision=2}
\begin{tabular}{llSSSSSSS}
\toprule
{} & {benchmark task ($k$-shot)} & {1} & {5} & {10} & {20} & {100} & {200} & {500} \\
{} & {algorithm} & {} & {} & {} & {} & {} & {} & {} \\

\cmidrule(r){1-2} \cmidrule(lr){3-3} \cmidrule(lr){4-4} \cmidrule(lr){5-5} \cmidrule(lr){6-6} \cmidrule(lr){7-7} \cmidrule(lr){8-8} \cmidrule(l){9-9}

\multirow[c]{21}{*}{\rotatebox[origin=c]{90}{Cohen's $\kappa$}}
 & no pre-training & \num{0.003943 +- 0.033466} & \num{0.094252 +- 0.034870} & \num{0.206467 +- 0.046576} & \num{0.247873 +- 0.126115} & \num{0.366522 +- 0.057806} & \num{0.465867 +- 0.018004} & \num{0.529218 +- 0.004147} \\
 \cmidrule(r){2-9}
 & transfer learning (\latvia) & \num{0.078961 +- 0.042835} & \num{0.123843 +- 0.059113} & \num{0.165136 +- 0.078976} & \num{0.265786 +- 0.072082} & \num{0.360756 +- 0.052804} & \num{0.399111 +- 0.037983} & \num{0.481769 +- 0.052511} \\
 & \gls{gl:anil} (\latvia) & \num{0.072956 +- 0.079063} & \num{0.155718 +- 0.020102} & \num{0.219838 +- 0.125464} & \num{0.357882 +- 0.039914} & \bfseries \color{blue} \num{0.420882 +- 0.039634} & \num{0.476311 +- 0.027588} & \bfseries \color{blue} \num{0.555626 +- 0.010196} \\
 & \gls{gl:maml} (\latvia) & \num{0.105690 +- 0.031705} & \num{0.119433 +- 0.029458} & \bfseries \color{blue} \num{0.272044 +- 0.021425} & \bfseries \color{blue} \num{0.362449 +- 0.049841} & \num{0.406335 +- 0.029844} & \bfseries \color{blue} \num{0.478939 +- 0.056396} & \num{0.534314 +- 0.033516} \\
 & \gls{gl:fomaml} (\latvia) & \num{0.068665 +- 0.067625} & \num{0.119755 +- 0.028356} & \num{0.224158 +- 0.041697} & \num{0.297610 +- 0.050845} & \num{0.382947 +- 0.038769} & \num{0.448730 +- 0.016480} & \num{0.510193 +- 0.007309} \\
 & \gls{gl:timl} (encoder) (\latvia) & \bfseries \color{blue} \num{0.112999 +- 0.071581} & \num{0.139438 +- 0.043090} & \num{0.188838 +- 0.038573} & \num{0.265663 +- 0.035984} & \num{0.357617 +- 0.028945} & \num{0.370211 +- 0.033033} & \num{0.476247 +- 0.019291} \\
 & \gls{gl:timl} (no encoder) (\latvia) & \num{0.106646 +- 0.051526} & \bfseries \color{blue} \num{0.180182 +- 0.043213} & \num{0.219423 +- 0.023912} & \num{0.314480 +- 0.045939} & \num{0.405761 +- 0.026271} & \num{0.425454 +- 0.017677} & \num{0.539305 +- 0.006409} \\
 \arrayrulecolor{gray} 
 \cmidrule(r){2-9}
 \arrayrulecolor{black}
 & transfer learning (\latvia + \portugal) & \num{0.065707 +- 0.007943} & \num{0.109282 +- 0.007445} & \color{mediumseagreen} \num{0.173478 +- 0.055036} & \num{0.208272 +- 0.061320} & \num{0.298326 +- 0.037112} & \num{0.379472 +- 0.030037} & \num{0.480642 +- 0.026814} \\
& \gls{gl:anil} (\latvia + \portugal) & \color{mediumseagreen} \num{0.108998 +- 0.008014} & \num{0.154693 +- 0.009150} & \num{0.214815 +- 0.068549} & \num{0.329799 +- 0.033568} & \num{0.395517 +- 0.026502} & \num{0.461207 +- 0.018519} & \num{0.518749 +- 0.024637} \\
& \gls{gl:maml} (\latvia + \portugal) & \num{0.084021 +- 0.044888} & \color{mediumseagreen} \num{0.140379 +- 0.011813} & \num{0.230856 +- 0.063911} & \num{0.320295 +- 0.064860} & \num{0.385673 +- 0.027734} & \num{0.458353 +- 0.021267} & \color{mediumseagreen} \num{0.536880 +- 0.014856} \\
& \gls{gl:fomaml} (\latvia + \portugal) & \color{mediumseagreen} \num{0.081529 +- 0.025749} & \color{mediumseagreen} \num{0.133469 +- 0.023330} & \num{0.197199 +- 0.048590} & \color{mediumseagreen} \num{0.316750 +- 0.041518} & \num{0.381449 +- 0.019691} & \color{mediumseagreen} \num{0.456667 +- 0.024708} & \color{mediumseagreen} \num{0.521612 +- 0.011850} \\
& \gls{gl:timl} (encoder) (\latvia + \portugal) & \num{0.068199 +- 0.026136} & \num{0.093660 +- 0.020811} & \num{0.147783 +- 0.086428} & \num{0.230075 +- 0.087681} & \num{0.332130 +- 0.015892} & \num{0.340179 +- 0.015534} & \num{0.458597 +- 0.039357} \\
& \gls{gl:timl} (no encoder) (\latvia + \portugal) & \num{0.101921 +- 0.023461} & \num{0.136948 +- 0.021891} & \num{0.205049 +- 0.016697} & \color{mediumseagreen} \num{0.315664 +- 0.046760} & \num{0.403138 +- 0.046969} & \color{mediumseagreen} \num{0.432441 +- 0.020132} & \num{0.525293 +- 0.011241} \\
\cmidrule(r){2-9}
 & \gls{gl:presto} & \bfseries \num{0.054569 +- 0.022766} & \bfseries \num{0.091412 +- 0.027760} & \num{0.109492 +- 0.029265} & \num{0.165872 +- 0.042706} & \num{0.322568 +- 0.025603} & \num{0.365870 +- 0.043121} & \num{0.429978 +- 0.010158} \\
 & \gls{gl:prestoxts} & \num{0.004639 +- 0.003694} & \num{0.053244 +- 0.038528} & \num{0.081511 +- 0.055111} & \num{0.055699 +- 0.061260} & \num{0.305045 +- 0.171922} & \num{0.379188 +- 0.022645} & \bfseries \num{0.518103 +- 0.024436} \\
 & \gls{gl:crossprestoxts} & \num{0.020338 +- 0.023742} & \num{0.039085 +- 0.040756} & \bfseries \num{0.122207 +- 0.066500} & \bfseries \num{0.248641 +- 0.009739} & \bfseries \num{0.344212 +- 0.037898} & \num{0.258626 +- 0.144724} & \num{0.495818 +- 0.018890} \\
  \arrayrulecolor{gray} 
 \cmidrule(r){2-9}
 \arrayrulecolor{black}
 & \gls{gl:prestoxts} (D) & \num{0.006708 +- 0.006674} & \num{0.035537 +- 0.033396} & \num{0.036813 +- 0.033721} & \num{0.079701 +- 0.080803} & \num{0.225447 +- 0.130339} & \num{0.360431 +- 0.023939} & \num{0.498123 +- 0.024313} \\
 & \gls{gl:prestoxts} (D, L) & \num{0.000039 +- 0.000088} & \num{0.024474 +- 0.021613} & \num{0.061425 +- 0.064032} & \num{0.078222 +- 0.111058} & \num{0.206183 +- 0.188930} & \num{0.233856 +- 0.214557} & \num{0.504012 +- 0.013701} \\
 & \gls{gl:crossprestoxts} (D) & \num{0.007909 +- 0.014817} & \num{0.037057 +- 0.039378} & \num{0.089006 +- 0.046923} & \num{0.064697 +- 0.033045} & \num{0.272807 +- 0.062110} & \bfseries \num{0.396030 +- 0.027137} & \num{0.498082 +- 0.008338} \\
 & \gls{gl:crossprestoxts} (D, L) & \num{0.000566 +- 0.000901} & \num{0.035993 +- 0.040391} & \num{0.057496 +- 0.058555} & \num{0.068372 +- 0.133702} & \num{0.262822 +- 0.156165} & \num{0.279918 +- 0.157220} & \num{0.485287 +- 0.020411} \\
\bottomrule
\end{tabular}

\end{table*}

\begin{table*}[p]
    \caption{%
    Fine-tuning results for the \taskalldata\ benchmark.
    We report \emph{Cohen's kappa} on the test set for each algorithm and few-shot task.
    Metrics are shown as mean $\pm$ standard deviation over five runs, \cf \cref{sec:exp_finetuning}.
    The best result per task is marked in \textbf{bold}.
    }
    \vspace{7pt}
    \label{tab:comparison_results_eurocropsml_S1S2ERA5_CK}
    \centering\scriptsize

\robustify\bfseries
\sisetup{detect-all=true,uncertainty-mode=separate,table-align-uncertainty=true,round-mode=uncertainty,round-precision=2}
\begin{tabular}{llSSSSSSS}
\toprule
{} & {benchmark task ($k$-shot)} & {1} & {5} & {10} & {20} & {100} & {200} & {500} \\
{} & {algorithm} & {} & {} & {} & {} & {} & {} & {} \\

\cmidrule(r){1-2} \cmidrule(lr){3-3} \cmidrule(lr){4-4} \cmidrule(lr){5-5} \cmidrule(lr){6-6} \cmidrule(lr){7-7} \cmidrule(lr){8-8} \cmidrule(l){9-9}

\multirow[c]{8}{*}{\rotatebox[origin=c]{90}{Cohen's $\kappa$}} 
 & no pre-training & \num{-0.005045 +- 0.012630} & \num{0.031545 +- 0.049923} & \num{0.072964 +- 0.034169} & \num{0.042928 +- 0.066192} & \num{0.223745 +- 0.129999} & \num{0.309342 +- 0.148613} & \num{0.486524 +- 0.033972} \\
\cmidrule(r){2-9}
 & \gls{gl:presto} & \bfseries \num{0.038433 +- 0.012606} & \bfseries \num{0.087200 +- 0.026785} & \bfseries \num{0.127639 +- 0.033429} & \bfseries \num{0.199295 +- 0.026882} & \bfseries \num{0.352414 +- 0.027591} & \bfseries \num{0.399062 +- 0.034501} & \num{0.495462 +- 0.018345} \\
 & \gls{gl:prestoxts} & \num{-0.003192 +- 0.020405} & \num{0.008315 +- 0.030214} & \num{0.065619 +- 0.046714} & \num{0.082042 +- 0.107911} & \num{0.278975 +- 0.158994} & \num{0.312009 +- 0.179021} & \bfseries \num{0.517587 +- 0.023104} \\
 & \gls{gl:crossprestoxts} & \num{0.000454 +- 0.002910} & \num{0.032813 +- 0.025397} & \num{0.091409 +- 0.010081} & \num{0.144865 +- 0.081365} & \num{0.240857 +- 0.134109} & \num{0.360997 +- 0.062310} & \num{0.491353 +- 0.021283} \\
 \arrayrulecolor{gray} 
 \cmidrule(r){2-9}
 \arrayrulecolor{black}
 & \gls{gl:prestoxts} (D) & \num{0.000036 +- 0.025388} & \num{0.004742 +- 0.014630} & \num{0.061718 +- 0.035345} & \num{0.077754 +- 0.070087} & \num{0.185659 +- 0.118015} & \num{0.270920 +- 0.132248} & \num{0.423223 +- 0.140480} \\
 & \gls{gl:prestoxts} (D, L) & \num{0.006172 +- 0.010131} & \num{0.007207 +- 0.010658} & \num{0.051830 +- 0.045800} & \num{0.109271 +- 0.088607} & \num{0.256189 +- 0.079989} & \num{0.284877 +- 0.103886} & \num{0.370499 +- 0.248284} \\
 & \gls{gl:crossprestoxts} (D) & \num{-0.010785 +- 0.032519} & \num{0.045937 +- 0.023751} & \num{0.083511 +- 0.026987} & \num{0.044973 +- 0.019932} & \num{0.051340 +- 0.025468} & \num{0.351844 +- 0.034321} & \num{0.478102 +- 0.058514} \\
 & \gls{gl:crossprestoxts} (D, L) & \num{0.005552 +- 0.028466} & \num{0.036326 +- 0.052350} & \num{0.023757 +- 0.042261} & \num{0.050494 +- 0.086189} & \num{0.198407 +- 0.133199} & \num{0.330008 +- 0.092102} & \num{0.483790 +- 0.022501} \\
\bottomrule
\end{tabular}

\end{table*}

\section{Run-time analysis}
\label{sec:runtimeanalysis}
\begin{table}
    \centering
     \caption{
     Run-times for the considered supervised algorithms during the pre-training and \num{500}-shot fine-tuning phases per run of the model training.
    For the meta-learning algorithms, we used the \num{4}-way \num{1}-shot \num{4}-step variant.
    For the pre-training and fine-tuning, we report the run-time of the training phase, excluding the tuning. 
    The fastest run-time per training phase is marked in \textbf{bold}.}
    \scriptsize

\robustify\bfseries
\sisetup{detect-all=true,uncertainty-mode=separate,table-align-uncertainty=true,round-mode=uncertainty,round-precision=2}

\begin{tabular}{@{}lSSSS@{}}
\toprule
{benchmark task} & \multicolumn{1}{X}{LV$\rightarrow$EE} & \multicolumn{1}{X}{LV$\rightarrow$EE} & 
\multicolumn{1}{X}{LV+PT$\rightarrow$EE} & \multicolumn{1}{X}{LV+PT$\rightarrow$EE} \\
{training phase} & \multicolumn{1}{X}{pre-training} & \multicolumn{1}{X}{fine-tuning} &\multicolumn{1}{X}{pre-training} & \multicolumn{1}{X}{fine-tuning}  \\
{algorithm / runtime} & \multicolumn{1}{X}{(in hours)} & \multicolumn{1}{X}{(in minutes)} &\multicolumn{1}{X}{(in hours)} & \multicolumn{1}{X}{(in minutes)}\\
\cmidrule(r){1-1} \cmidrule(lr){2-2} \cmidrule(lr){3-3}
\cmidrule(lr){4-4} \cmidrule(l){5-5}
no pre-training & n/a & \textbf{\num{3.82 +- 1.88}} &n/a & \num{11.66 +-2.86}\\
\cmidrule(r){1-5}
transfer learning & \textbf{\num{0.73}}  & \num{3.98 +- 1.19} & \textbf{\num{2.9}} & \num{10.06 +- 13.049}\\
ANIL & \num{3.8} & \num{5.84 +-2.3} &\num{3.9} & \num{7.4+-2.997} \\
MAML & \num{5.3} & \num{5.12 +- 2.78} &\num{5.3} & \textbf{\num{5.88 +- 2.3784}} \\
FOMAML & \num{4.1} & \num{7.44 +- 2.35} & \num{4.2} & \num{8.32+-7.524} \\
TIML (encoder) & \num{4.3} & \num{10 +- 3.47} & \num{4.3} & \num{19.94+-9.8839} \\
TIML (no encoder) & \num{5.4} & \num{4.52 +- 1.88} & \num{5.7} & \num{17.28+-12.491} \\
\bottomrule
\end{tabular}

    \label{tab:runtimes}
\end{table}

Besides performance measures such as prediction accuracy, computational costs and training times required to achieve performance are also of interest when benchmarking algorithms for their knowledge transfer capabilities.
A complete and fully comparable run-time evaluation of all experiments considered in our benchmark is not possible since the experimental runs were partly run in parallel on multiple machines with varying hardware specifications.
Hence, we cannot present a joint run-time analysis of supervised and \glsxtrlong{gl:ssl} algorithms.

Instead, for supervised methods, we present a run-time analysis of one pre-training variant per algorithm.
Moreover, since meta-learning promises faster task adaptation, we also evaluate one fine-tuning task per algorithm.
\Cref{tab:runtimes} shows the model training times per algorithm during supervised pre-training and \num{500}-shot fine-tuning.
For meta-learning algorithms, we report the times of the \num{4}-way \num{1}-shot \num{4}-step variant.
All of these were performed under comparable circumstances on the same machine with 16 cores of AMD® EPYC™ 9654 CPU, 64 GB RAM, and an NVIDIA® GeForce RTX™ 4090 GPU.

The randomly initialized benchmark model has the shortest overall training time, hence requiring no pre-training at all, as well as having the lowest fine-tuning times.
While meta-learning algorithms of the \gls{gl:maml} family fulfill the promise of quicker task adaptation and, thus, in general, good fine-tuning times, this comes at significantly increased computational costs during pre-training.
Altogether, the meta-learning approaches require longer total training times, with \gls{gl:timl} (without encoder) being the most compute-heavy of all compared algorithms. Interestingly, the run-times of the meta-learning algorithms are less affected by incorporating additional pre-training data from Portugal, while transfer learning takes significantly longer in this case.
After pre-training also on Portugal data, for the fine-tuning phase, the \gls{gl:maml} family performs best.

To assess the computational efficiency of \gls{gl:ssl} methods, we compare \gls{gl:prestoxts} with \gls{gl:crossprestoxts}, which is designed to reduce training costs during pre-training.
On an AMD® EPYC™ 7763 CPU (\num{64} cores), two NVIDIA® RTX™ \num{6000} Ada GPUs (\num{48} GB each), and \num{512} GB of system RAM, \gls{gl:crossprestoxts} achieves \num{13}\% higher throughput on average, processing each batch with \SI{0.31}{\second} instead of \SI{0.36}{\second}.
Moreover, it also converges faster, typically triggering early stopping sooner.
These improvements lead to an overall reduction in the average pre-training time of approximately \num{52}\%.

\end{document}